\documentclass[10pt,journal,compsoc]{IEEEtran}
\usepackage{amsmath,amsfonts}
\usepackage{array}
\usepackage[caption=false,font=normalsize,labelfont=sf,textfont=sf]{subfig}
\usepackage{textcomp}
\usepackage{stfloats}
\usepackage{url}
\usepackage{verbatim}
\usepackage{graphicx}
\usepackage{cite}
\usepackage[ruled]{algorithm2e}
\usepackage[pagebackref,breaklinks,colorlinks,linkcolor=blue,citecolor=blue, bookmarks=false]{hyperref}
\usepackage[capitalize]{cleveref}
\crefname{section}{Sec.}{Secs.}
\Crefname{section}{Section}{Sections}
\Crefname{table}{Table}{Tables}
\crefname{table}{Tab.}{Tabs.}
\crefname{figure}{Fig.}{Figs.}
\Crefname{figure}{Figure}{Figures}
\crefname{equation}{Eq.}{Eqs.}
\Crefname{equation}{Equation}{Equations}
\crefname{appendix}{Appendix.}{Appendices.}
\Crefname{appendix}{Appendix}{Appendices}

\usepackage{multirow}
\usepackage[normalem]{ulem}
\useunder{\uline}{\ul}{}
\usepackage{booktabs} 
\usepackage{pifont}
\usepackage{graphicx}
\usepackage{xcolor}
\definecolor{softgreen}{rgb}{60,179,113}
\usepackage[normalem]{ulem}

\newcommand{\exten}{\textcolor{black}}

\begin{document}

\title{
M3DM-NR: RGB-3D Noisy-Resistant Industrial Anomaly Detection via Multimodal Denoising
}


\author{Chengjie Wang,
        Haokun Zhu, 
        Jinlong Peng,
        Yue Wang, 
        Ran Yi\\
        Yunsheng Wu,
        Lizhuang Ma,
        Jiangning Zhang
\IEEEcompsocitemizethanks{
\IEEEcompsocthanksitem C.~Wang is with Shanghai Jiao Tong University and Youtu Lab, Shanghai, China.
\IEEEcompsocthanksitem H.~Zhu, Y.~Wang, R.~Yi, and L.~Ma are with the Shanghai Jiao Tong University, Shanghai, China.
\IEEEcompsocthanksitem J.~Peng, Y.~Wu, and J.~Zhang are with Youtu Lab, Tencent, China. 
}
\thanks{Corresponding author: Ran Yi}
}

\markboth{Journal of \LaTeX\ Class Files,~Vol.~14, No.~8, August~2021}%
{Shell \MakeLowercase{\textit{et al.}}: A Sample Article Using IEEEtran.cls for IEEE Journals}

\IEEEpubid{0000--0000/00\$00.00~\copyright~2021 IEEE}
\IEEEtitleabstractindextext{
\begin{abstract}
Existing industrial anomaly detection methods primarily concentrate on unsupervised learning with pristine RGB images. 
Yet, both RGB and 3D data are crucial for anomaly detection, and the datasets are seldom completely clean in practical scenarios. 
To address above challenges, this paper initially delves into the RGB-3D multi-modal noisy anomaly detection, proposing a novel noise-resistant M3DM-NR framework to leveraging strong multi-modal discriminative capabilities of CLIP.
M3DM-NR consists of three stages: 
\textbf{Stage-I} introduces the Suspected References Selection module to filter a few normal samples from the training dataset, using the multimodal features extracted by the Initial Feature Extraction, and a Suspected Anomaly Map Computation module to generate a suspected anomaly map to focus on abnormal regions as reference. 
\textbf{Stage-II} uses the suspected anomaly maps of the reference samples as reference, and inputs image, point cloud, and text information to achieve denoising of the training samples through intra-modal comparison and multi-scale aggregation operations. 
Finally, \textbf{Stage-III} proposes the Point Feature Alignment, Unsupervised Feature Fusion, Noise Discriminative Coreset Selection, and Decision Layer Fusion modules to learn the pattern of the training dataset, enabling anomaly detection and segmentation while filtering out noise.
Extensive experiments show that M3DM-NR outperforms state-of-the-art methods in 3D-RGB multi-modal noisy anomaly detection.

\end{abstract}
}

\maketitle

\section{Introduction}
\begin{figure}[t]
    \centering
    \includegraphics[width=0.9\linewidth]{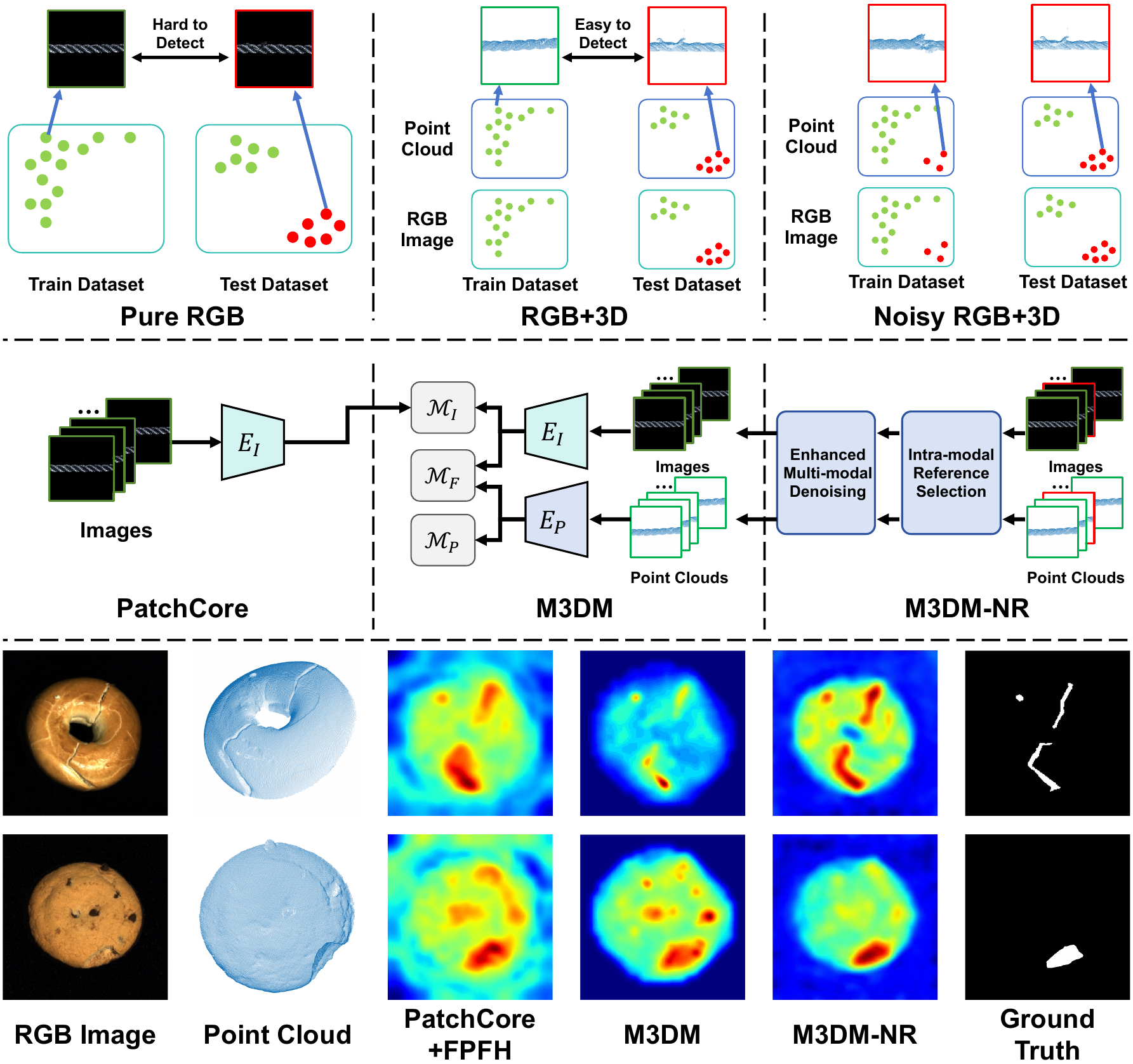}
    \caption{
    \textbf{Top}: Intuitive diagram of different task settings. 
    \textbf{Middle}: Representative PatchCore~\cite{patchcore} for solving RGB images, our M3DM~\cite{wang2023multimodal} (conference version) for solving multi-modal RGB+3D data, and new M3DM-NR to tackle more challenging and practial noisy setting. 
    \textbf{Bottom}: Quantitative visualization results on MVTec 3D-AD dataset~\cite{mvtec3dad}. Our M3DM-NR can predict more precise anomaly regions obviously compared to PatchCore+FPFH~\cite{horwitz2023back} and M3DM~\cite{wang2023multimodal}. 
    }
    \label{fig:vt_motivation}
\end{figure}

Industrial anomaly detection aims to find the abnormal region of products and plays an important role in industrial quality inspection. 
Most existing industrial anomaly detection methods~\cite{cao2024survey,liu2024deep} primarily focus on RGB images~\cite{mvtec,realiad} and use a vast number of normal examples for training. 
Consequently, current industrial anomaly detection methods predominantly rely on unsupervised approaches, meaning they train exclusively on normal RGB examples and only during inference are defect examples tested. These two factors contribute to two significant issues (\cref{fig:vt_motivation}-Top-Left).
First, during the quality inspection of industrial products, human inspectors rely on both 3D shape and color characteristics to assess product quality. 
The 3D shape information is crucial for accurate defect detection in particular, and identifying defects using only RGB images proves difficult. 
With advancements in 3D sensor technology, recent MVTec-3D AD dataset that includes both 2D images and 3D point cloud data is proposed to alleviate this problem and has bolstered research in multi-modal industrial anomaly detection (\cref{fig:vt_motivation}-Top-Middle). 
Second, the presence of noise in the normal dataset is an unavoidable issue in real-world applications, particularly in industrial manufacturing where products are mass-produced daily. Most existing unsupervised AD methods~\cite{patchcore, cflow, zheng2022focus} are prone to noisy data due to their exhaustive strategy to model the training set. However, noisy samples can easily mislead those overconfident AD algorithms, causing them to misclassify similar anomaly samples in the test set and generate incorrect locations. SoftPatch~\cite{jiang2022softpatch} is the first to introduce the setting for noisy industrial detection, but it explored only noisy industrial detection on RGB data. 

For the first issue, the core idea for existing unsupervised anomaly detection is to find out the difference between normal representations and anomalies. 
Current 2D industrial anomaly detection methods can be mainly categorized into two categories:
(1) Reconstruction-based methods.
Image reconstruction tasks are widely used in anomaly detection methods~\cite{mvtec, memorizing-ae, reconstruction, draem, ReverseDistillation, ocgan} to learn normal representation. 
Reconstruction-based methods are easy to implement for a single modal input (2D image or 3D point cloud).
But for multi-modal inputs, it is hard to find a reconstruction target.
(2) Pretrained feature extractor-based methods.
An intuitive way to utilize the feature extractor is to map the extracted feature to a normal distribution and find the out-of-distribution one as an anomaly.
Normalizing flow-based methods~\cite{cflow,fastflow,ast} use an invertible transformation to directly construct normal distribution, and memory bank-based methods\cite{padim, patchcore} store some representative features to implicitly construct the feature distribution.
Compared with reconstruction-based methods, directly using a pretrained feature extractor does not involve the design of a multi-modal reconstruction target and is a better choice for the multi-modal task.
Besides that, current multi-modal industrial anomaly detection methods~\cite{3d-ads, ast} directly concatenate the features of the two modalities together.
However, when the feature dimension is high, the disturbance between multi-modal features will be violent and cause performance reduction.

Regarding the second issue of noisy anomaly detection, existing methods in noisy industrial detection have primarily focused on single-modality noisy anomaly detection using RGB images, with a lack of research on RGB-3D multi-modal noisy data. However, in practical industrial detection, noise often contaminates 3D data, and RGB-3D multi-modal data serve as an important reference for determining whether a sample is anomalous. The absence of exploration in RGB-3D multi-modal noisy data means that current methods are vulnerable to the multi-modal noisy data in real-world production environments. Furthermore, existing approaches employ a simplistic and naive strategy of patch-level denoising and sample re-weighting based on outlier-detection weights, leading to unsatisfying denoising effects and the persistence of noise in the dataset.

\begin{figure*}[t]
    \centering
    \includegraphics[width=0.9\linewidth]{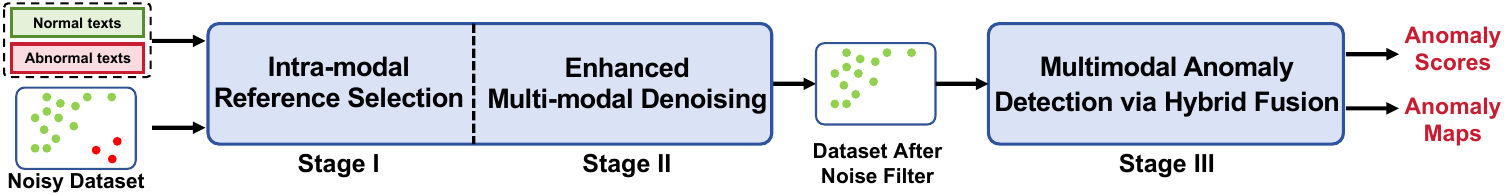}
    \caption{\textbf{Overall pipeline of our M3DM-NR} that comprises three stages: 1) selecting intra-modal reference samples, 2) denoising the dataset by comparing it with these samples, and 3) achieving multimodal anomaly detection through multimodal feature fusion.}
    \label{fig:vt_pipeline}
\end{figure*}

To solve the problems mentioned above, in this paper, we first delve into the RGB-3D multi-modal noisy industrial detection problem (\cref{fig:vt_motivation}-Top-Right). To address the challenges of RGB-3D multi-modal noisy data, we propose a novel \textit{three-stage multi-modal noise-resistant framework} termed M3DM-NR, which performs denoising at both sample-level and patch-level, as shown in \cref{fig:vt_pipeline}. 
This framework utilizes pretrained CLIP~\cite{radford2021learning} and Point-BIND~\cite{guo2023point} models to extract aligned text, RGB, and 3D point cloud features to denoise multi-modal data through both cross-modal comparison and intra-modality comparison. To the best of our knowledge, we are the first to employ a multi-modal learning approach based on pre-trained CLIP and Point-BIND to solve the RGB-3D multi-modal noisy industrial anomaly detection problem. 
In this framework, Stage I selects a few normal samples from the training dataset as intra-modal reference samples and compute the suspected anomaly map to focus on abnormal regions by the proposed \textit{Intra-Modal Reference Selection}. 
In Stage II, recognizing the fact that in industrial anomaly detection, anomalies often constitute only a small fraction of the entire sample, we thus propose a novel \textit{Enhanced Multi-modal Denoising} module to rank the anomalies of each training sample by performing multi-scale feature comparison and weighting with a suspected reference, enabling the filtering of anomalous samples.
In Stage III, to address the above problems concerning multi-modal anomaly detection, we propose a novel \textit{Multimodal Anomaly Detection via Hybrid Fusion} scheme to Learn the pattern of the training dataset to conduct anomaly detection and segmentation while filtering out noise at the patch level. 
Different from the existing methods that directly concatenate the features of the two modalities, we propose a hybrid fusion scheme to reduce the disturbance between multi-modal features and encourage feature interaction.
We propose \textit{Unsupervised Feature Fusion (UFF)} to fuse multi-modal features, which is trained using a patch-wise contrastive loss to learn the inherent relation between multi-modal feature patches at the same position. 
To encourage the anomaly detection model to keep the single domain inference ability, we construct three memory banks separately for RGB, 3D and fused features.
For the final decision, we construct \textit{Decision Layer Fusion (DLF)} to consider all memory banks for anomaly detection and segmentation. Besides, we further propose a \textit{Point Feature Alignment (PFA)} operation to better align 3D and 2D features and \textit{Noise Discriminative Coreset Selection} to filter out noise at patch-level.

To evaluate our method, we conduct extensive experiments on the MVTec 3D-AD~\cite{mvtec3dad} and Eyecandies~\cite{eyecandies} datasets, comparing our method with existing RGB, 3D, and RGB-3D based industrial detection methods. Moreover, to further highlight the robustness of our method, we follow the experiment setting in SoftPatch~\cite{jiang2022softpatch} and conduct experiments under Non-Overlap and, more challenging, Overlap settings. The extensive experimental results and metrics (I-AUROC, P-AUROC, AUPRO) demonstrate that our method surpasses existing state-of-the-art approaches. Additionally, we performed a comprehensive ablation study, thoroughly validating the effectiveness of all novel modules proposed.

This is an extension of the previous conference version (M3DM~\cite{wang2023multimodal} in CVPR'23). 
In the conference papar, we mainly proposed M3DM, a novel multi-modal industrial anomaly detection method with hybrid feature fusion, which outperforms the state-of-the-art detection and segmentation precision on MVTec 3D-AD~\cite{mvtec3dad}. 
In this extended journal version, we make the following four contributions: 
\begin{itemize}
    \item  We study a new RGB-3D multi-modal noisy industrial anomaly detection task and have substantially broadened our research to this practical setting, proposing a novel \textit{three-stage multi-modal noise-resistant framework} termed M3DM-NR. It addresses reference selection, denoising, and final anomaly detection and segmentation, ensuring systematic and hierarchical processing. 
    \item We design three novel \textit{Initial Feature Extraction}, \textit{Suspected References Selection}, and \textit{Suspected Anomaly Map Computation} modules in Stage I to select a few normal samples from the training dataset as intra-modal reference samples, and it generates suspected anomaly maps to focus on abnormal regions as the reference for the next stage. 
    \item To obtain cleaner training data, we propose an extra Stage II termed \textit{Enhanced Multi-modal Denoising} to introduce multi-scale feature comparison and weighting methods to finely rank and denoise training samples. 
    \item We employ M3DM as Stage III to achieve final anomaly detection and segmentation. Extensive quantitative experiments across various settings demonstrate the performance of our approach over existing state-of-the-art methods in 3D-RGB multi-modal noisy anomaly detection. We also conduct massive ablation study to illustrate the effectiveness of each designed component. 
\end{itemize}
    
\section{Related Work}

\subsection{2D Industrial Anomaly Detection}
Current anomaly detection can be mainly categorized into following three parts: 
\textbf{\textit{1)} Data augmentation based} methods~\cite{cutpaste,defectgan,draem,simplenet,memseg,rrd} propose to introduce pseudo anomalies to normal samples with the aim of improving the system's ability to identify such anomalies during training. 
\textbf{\textit{2)} Reconstruction based} methods~\cite{utrad,memorizing-ae,reconstruction,ocgan,ReverseDistillation,ocrgan,vitad,mambaad,invad,he2023diad} leverage auto-encoders and generative adversarial networks. Although these reconstruction methods may not accurately recover anomalous regions, comparing the reconstructed image with the original can pinpoint anomalies and facilitate decision-making.
\textbf{\textit{3)} Feature embedding based} methods~\cite{wan2022unsupervised,cdo,cflow,fastflow,patchcore,pyramidflow,salehi2021multiresolution,cao2022informative} depend on pre-trained feature extractors, with additional detection modules that learn to identify abnormal areas using the extracted features or representations. 
Drawing parallels between 2D and 3D anomaly detection, our work expands the application of the memory bank approach to 3D and multi-modal contexts, yielding impressive outcomes.

\subsection{3D Industrial Anomaly Detection} 
The first public 3D industrial anomaly detection dataset is the MVTec 3D-AD dataset~\cite{mvtec3dad}, which includes both RGB information and point position data for each instance. 
Current 3D anomaly detection can be mainly categorized into following four parts: 
\textbf{\textit{1)} Data augmentation-based} methods~\cite{chen2023easynet, zavrtanik2024keep} draw inspirations from 2D anomaly detection strategies to generate pseudo RGB and 3D anomaly samples, enhancing the model's capacity to identify anomalies.
\textbf{\textit{2)} Reconstruction-based} methods~\cite{li2023towards, chen2023easynet} utilize auto-encoders and generative adversarial networks trained to generate normal samples for both RGB and 3D data, irrespective of whether the input is normal or anomalous. This approach fails to reconstruct regions with anomalies effectively. By comparing these reconstructed samples with the originals, anomalies can be identified, thus aiding in decision-making.
\textbf{\textit{3)} Feature embedding-based} methods~\cite{horwitz2023back, wang2023multimodal, cao2023complementary, chu2023shape, tu2024self, zhao2024pointcore} rely on pre-trained feature extractors, supplemented with additional fusion modules that align and integrate multi-modal information. Detection modules then utilize these fused features or representations to identify abnormal areas, enhancing the system’s ability to detect anomalies.
\textbf{\textit{4)} Knowledge distillation-based} methods~\cite{3d-st, ast, gu2024rethinking} train a student network to reconstruct samples or extract features, where the disparity between the teacher and student networks serves as an indicator of anomalies.
In our research, we adopt the feature embedding-based approach but diverge with a novel pipeline.

\subsection{Learning with Noisy Data} 
Recognizing noisy labels is increasingly gaining attention in the realm of supervised learning. Yet, this concept has scarcely been ventured into within unsupervised anomaly detection, largely due to the absence of clear labels. In classification tasks, certain studies have suggested filtering pseudo-labeled data that carry a high confidence threshold to mitigate noise~\cite{hu2021simple, sohn2020fixmatch}. Li et al.~\cite{li2020dividemix} employ a mixture model to identify noisy-labeled data, adopting a semi-supervised approach for training. 
In the field of object detection, strategies such as multi-augmentation~\cite{xu2021end}, a teacher-student model~\cite{liu2021unbiased}, or contrastive learning~\cite{yang2022class} have been leveraged, drawing on the expertise of expert models to reduce noise.
However, the prevailing methods for recognizing noisy labels depend heavily on labeled data for correcting inaccuracies. Our research diverges by aiming to enhance a model's resistance to noise in an unsupervised manner, thereby eliminating the need for manual annotations. 
A recent review~\cite{han2022adbench} examines the robustness of 30 AD algorithms, yet overlooks unsupervised approaches in the context of annotation errors. Pang et al.~\cite{pang2020self} address anomalies in video without relying on manually labeled data, exploiting information across consecutive frames, contrasting our focus on detecting anomalies in single images. Other studies~\cite{liu2021rca, zhou2017anomaly, wu2022understanding} tackle the elimination of noisy and corrupted data in semantic anomaly detection.
SoftPatch~\cite{jiang2022softpatch} proposed to filter out noise at patch-level using outlier detection, but the employed outlier detection method is rather naive and doesn't produce very good results. 
In this paper, we introduce a method that utilizes a pretrained CLIP-based model to extract and align multi-modal information, enabling the effective filtration of noise at sample-level.

\subsection{Multi-modal Learning}
Among the recent successes of large pre-trained vision-language models (VLMs)~\cite{alayrac2022flamingo, jia2021scaling, radford2021learning}, CLIP~\cite{radford2021learning} stands out as the first to employ pre-training on web-scale image-text data, demonstrating unprecedented generality. Notable features include its language-driven zero-shot inference capabilities, which have significantly enhanced both effective robustness~\cite{taori2020measuring} and perceptual alignment~\cite{goh2021multimodal}. 
Other studies~\cite{rao2022denseclip, zhong2022regionclip, zhou2022extract} have also utilized the pre-trained CLIP model for downstream tasks, such as language-guided detection and segmentation, achieving promising results.
Beyond aligning vision and language, Point-Bind~\cite{guo2023point} extends this alignment to include 3D modality. 
Recently, some recent works have attempted to apply the multimodal CLIP model to the AD domain~\cite{jeong2023winclip,aprilgan,saa,clipad,gpt-4v-ad}. Specific WinCLIP~\cite{jeong2023winclip} leverages the robust multi-modal capabilities of the pre-trained CLIP model for effective zero-shot 2D anomaly detection.

In this paper, we utilize the Point-BIND's aligned embedding space of image, language, and 3D modalities to effectively filter out noise at sample-level in the training set.

\begin{figure*}[t]
    \centering
    \includegraphics[width=1.0\linewidth]{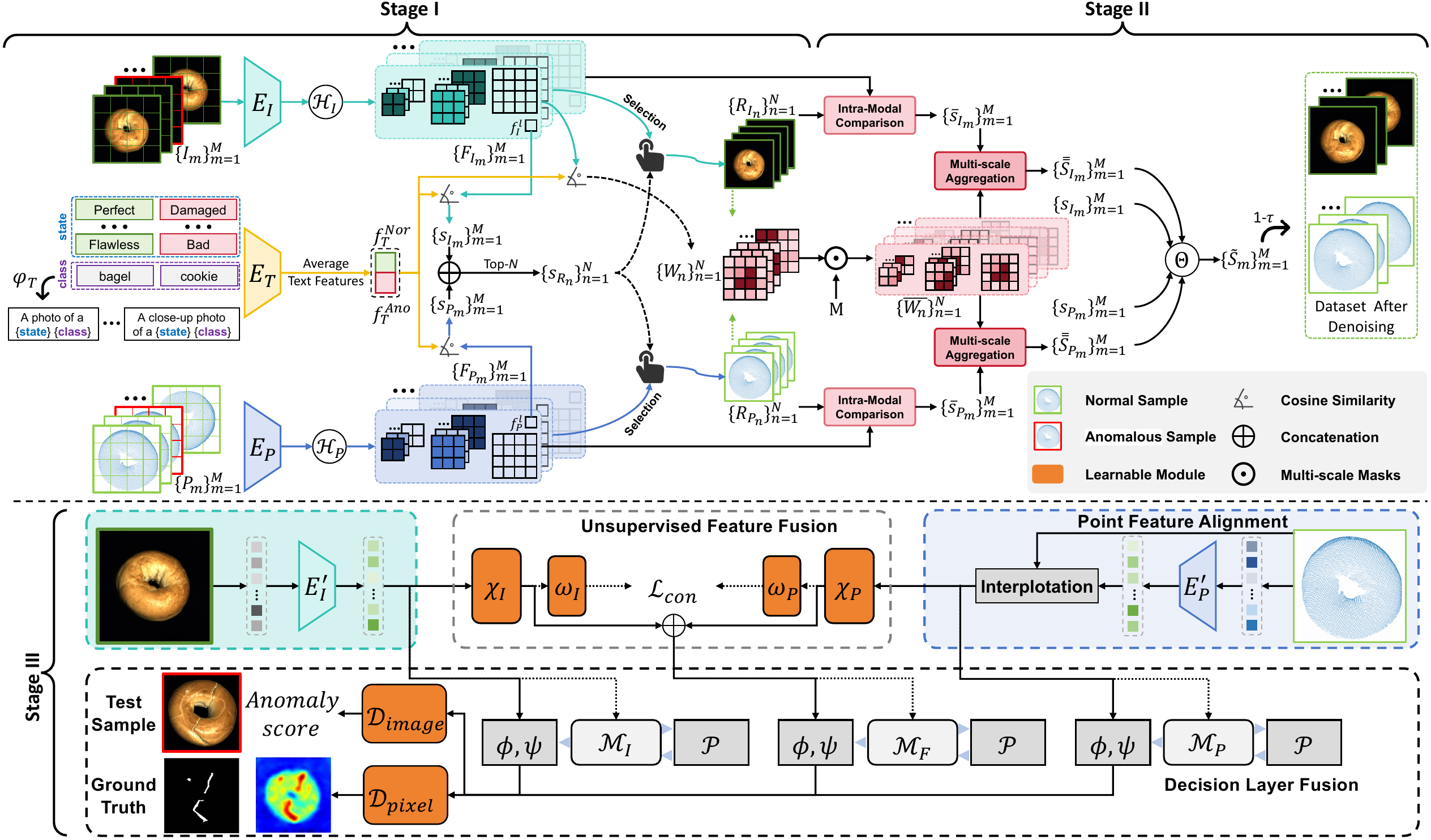}
    \caption{\textbf{Overview framework of our M3DM-NR}, which contains three stages to tackle challenging noisy anomaly detection task: 
    \textbf{Stage I} introduces a text prompt ensemble strategy $\varphi_T$, utilizing pre-trained image encoder $E_{I}$, point cloud encoder $E_{P}$, and text encoder $E_{T}$ to extract initial features $\left\{F_{I_m}\right\}_{m=1}^M$, $\left\{F_{P_m}\right\}_{m=1}^M$, $f_{T}^{Nor}$, and $f_{T}^{Ano}$. These features are then used to select suspected reference samples $\left\{s_{R_n}\right\}_{n=1}^N$ through similarity measurement and to compute corresponding anomaly maps $\left\{W_n\right\}_{n=1}^N$. 
    Based on the suspected samples, \textbf{Stage II} calculates the anomaly scores $\left\{\tilde{S}_m\right\}_{m=1}^M$ for each training sample using multi-scale and feature weighting methods, ultimately filtering out the top-$\tau$ samples to obtain a denoised training set. 
    \textbf{Stage III} comprises four modules to achieve final anomaly detection and segmentation.
 }
    \label{fig:vt_method}
\end{figure*}
\section{Methodology}
\label{sec:method}

As shown in \cref{fig:vt_method}, our proposed M3DM-NR framework takes RGB images and 3D point clouds as input to perform RGB-3D based multi-modal noisy anomaly detection and segmentation. Specifically, M3DM-NR consists of three stages to achieves this goal: 
1) Intra-modal Reference Selection (\textbf{Stage I} in \cref{sec:stage1}) selects a few normal samples from the training dataset as intra-modal reference samples, and the suspected anomaly map is computed to focus on abnormal regions. 
2) Enhanced Multi-modal Denoising (\textbf{Stage II} in \cref{sec:stage2}) ranks the anomalies of each training sample by performing multi-scale feature comparison and weighting with a suspected reference, enabling the filtering of anomalous samples. 
3) Multimodal Anomaly Detection via Hybrid Fusion (\textbf{Stage III} in \cref{sec:stage3}) learns the pattern of the training dataset to conduct anomaly detection and segmentation while filtering out noise at patch-level. 

\subsection{Stage I: Intra-modal Reference Selection} \label{sec:stage1}

\subsubsection{Initial Feature Extraction}
Given $M$ image and point cloud pairs $\left\{I_m\right\}_{m=1}^M$ and $\left\{P_m\right\}_{m=1}^M$, RGB-3D anomaly detection requires three modes of information input, so it contains three parts of feature pre-extraction algorithm: 

\noindent
\textbf{Text prompt ensemble.}
The effectiveness of text descriptions is crucial for multimodal anomaly detection. Following APRIL-GAN~\cite{aprilgan}, we employ a text prompt ensemble strategy $\varphi_T$ to fully explore the textual representation of defects. Specifically, the proposed strategy $\varphi_T$ includes several templates, each in the format ``A photo of a {state} {class}", where `state' denotes predefined normal and abnormal state descriptions, and `class' represents the class name. The output features are averaged using pooling to obtain the final descriptive features $f_{T}^{Nor} \in \mathbb{R}^{d} $ and $f_{T}^{Ano} \in \mathbb{R}^{d}$.
  
\noindent
\textbf{Multi-scale image feature representation.} 
For each image $I_{m}$ in the training dataset, we first use pretrained image encoder $E_{I}$ in CLIP model to extract corresponding feature $F_{I_{m}}$:
\begin{equation}
\label{eq:rgb}
    \begin{aligned}
        F_{I_{m}} = E_{I}(I_{m}). 
    \end{aligned}
\end{equation}
Then, a multi-scale segmentation operation $\mathcal{H}_I$ is used to segment $F_{I_{m}}$ into 3 different scales $F_{I_{m}}^{\sigma}, \sigma \in\{l, m, s\}$, denoted as: 
\begin{equation}
\label{eq:rgb}
    \begin{aligned}
        f_{I_{m}}^{l}, F_{I_{m}}^{l}, F_{I_{m}}^{m}, F_{I_{m}}^{s} &= \mathcal{H}_I\left(F_{I_{m}}\right). 
    \end{aligned}
\end{equation}
where $f_{I_{m}}^{l}$ is the class token and $F_{I_{m}}^{\sigma}$ is obtained by the following equation: 
\begin{equation}
\label{eq:rgb}
    \begin{aligned}
        F_{I_{m}}^{\sigma} &= \left\{f_{u v}^{\sigma}\right\}_{I m} \\
        &= F_{I_{m}} \odot \left\{M_{u v}^\sigma\right\} \\
        & \textit{s.t.} ~\sigma \in\{l, m, s\}. 
    \end{aligned}
\end{equation}
$M = \left\{M_{u v}^\sigma\right\}$ is the multi-scale mask, where each $M_{uv}^\sigma \in \{0,1\}^{h \times w}$ is a binary mask that selects $k \times k$ kernel size centered at $(u,v)$, with ${M_{uv}^{l}}$ specifically selects the entire point cloud. $F_{I_{m}}^{\sigma}$ is the set of image patches at big, middle, or small scale, $uv$ indicates the coordinate of patches in the original image, and $\odot$ denotes the element-wise multiplication. 

\begin{figure}[t]
    \centering
    \includegraphics[width=1.0\linewidth]{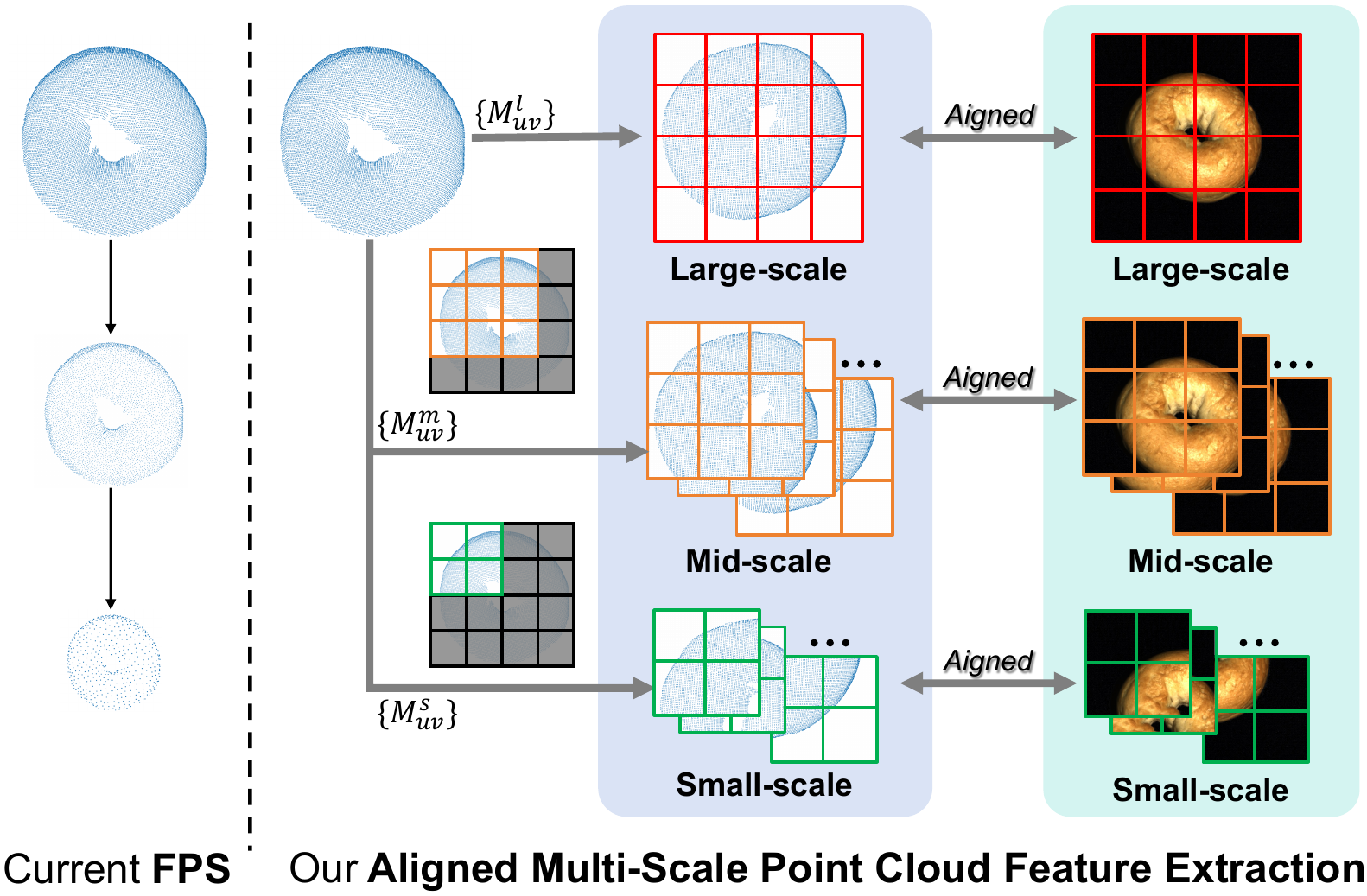}
    \caption{\textbf{Visualization of Aligned Multi-Scale Point Cloud Feature Extraction (AMPCFE),} which extracts local point cloud features aligned with the granularity of image patching, focusing more on local details and improving the efficacy of multi-modal anomaly detection.}
    \label{fig:vt_multi_scale}
\end{figure}

\noindent
\textbf{Aligned multi-scale point cloud feature extraction.} 
As previous work~\cite{wang2023multimodal} shown, in the MVTec 3D-AD~\cite{mvtec3dad} dataset, many anomalies cannot be detected through RGB images alone. For example, in the `potato' category, an anomaly type named `cut' can only be identified using 3D point cloud data. Thus, incorporating 3D point cloud data in the noise-filtering process is crucial. Therefore, we proposed to use 3D point cloud modality in noise detection.

However, we find that relying solely on the whole point cloud was insufficient during the experiments. In the MVTec 3D-AD dataset, defects often occupy only a small portion of the entire sample's point cloud data, meaning that most areas of a sample are normal. Furthermore, existing works~\cite{zhang2022pointclip, zhu2023pointclip, guo2023point, xue2023ulip} aligning point cloud encoders with CLIP focus on object classification tasks, which prioritize the global information of the object's 3D point cloud data and overlook local details. Traditional multi-scale point cloud data segmentation based on FPS sampling (Fig.~\ref{fig:vt_multi_scale}-Left) presents a full point cloud perspective with varying levels of sparsity but fails to specifically highlight local details. Yet, focusing on these details is crucial for detecting noise samples.

To address this problem, we propose a novel Aligned Multi-Scale Point Cloud Feature Extraction module, as shown in the right part of Fig.~\ref{fig:vt_multi_scale}. 
This approach enhances the ability of localized noise detection by extracting local point cloud features aligned with the granularity of image patching. 
Specifically, for each point cloud $P_{m} \in \mathbb{R}^{h \times w \times 3}$ in the training dataset, we segment $P_{m}$ into three scales, mirroring the approach used for image segmentation. 
Also, we generate 3 sets of masks $\{M_{uv}^{l}\}$, $\{M_{uv}^{m}\}$, and $\{M_{uv}^{s}\}$ as aforementioned operation of image. By applying these three sets of masks to the entire point cloud, we obtain three distinct sets of point clouds at different scales: 

\begin{equation}
     \{P_{uv}^{\sigma}\}_{m} = P_{m} \odot \{M_{uv}^{\sigma}\}, \; \sigma \in \{l, m, s\},
\end{equation}
Unlike images, in point cloud modality, only the points that do not fall on the backplane are meaningful. Consequently, some smaller patches of the point cloud may contain only a few meaningful points or none at all, making them insignificant or even obstructive for anomaly detection. To enhance efficiency, we identify and discard these non-contributory patches during the segmentation. This process results in filtered sets of point clouds: 

\begin{equation}
    \begin{aligned}
        \{\hat{P}_{uv}^{\sigma}\}_{m} = \{P_{uv}^{\sigma}|Num(P_{uv}^{\sigma})>\theta\}_{m}, \; \sigma \in \{l, m, s\},
    \end{aligned}
\end{equation}
where $\theta$ is a hyper-parameter representing the thresholds for the minimum number of points required in a point cloud patch to be considered meaningful.

These sets of point clouds constitute three distinct scales of point cloud representation. 
The granularity of these patches is aligned with that of image patches, enhancing the efficacy of subsequent multi-modal anomaly detection. We extract features from these multi-scale point cloud patches: 
\begin{equation}
\label{eq:rgb}
    \begin{aligned}
        f_{P_{m}}^{l}, F_{P_{m}}^{l} &= \mathcal{H}_P\left(E_P(\{\hat{P}_{uv}^{l}\}_{m})\right) \\
        F_{P_{m}}^{m} &= \mathcal{H}_P\left(E_P(\{\hat{P}_{uv}^{m}\}_{m})\right) \\
        F_{P_{m}}^{s} &= \mathcal{H}_P\left(E_P(\{\hat{P}_{uv}^{s}\}_{m})\right) \\
    \end{aligned}
\end{equation}
where $f_{P_{m}}^{l}$ is the class token and $F_{P_{m}}^{\sigma}$ is the feature map of $\sigma$-scale point cloud.

\subsubsection{Suspected References Selection}
We first try to identify noise samples in the training dataset solely by comparing the class tokens of text and RGB images.
However, we observed that certain samples in the MVTec 3D-AD~\cite{mvtec3dad} dataset cannot be straightforwardly classified using only cross-modal comparison, \textit{i.e.}, text and image class tokens. For example, the `Foam' category in MVTec 3D-AD includes a defect type labeled `color', which defies classification with our text templates and necessitates comparison with an RGB reference image of a normal sample. Consequently, to achieve comprehensive anomaly classification, a language-guided zero-shot approach falls short, as some defects are only identifiable through intra-modal references, not merely by cross-modal comparison.
Given that noise data constitutes a relatively small fraction of the entire training set, the majority of data are normal samples, we propose to select $N$ samples that are most representative of normality from the training set in Stage I. These samples will then serve as intra-modal references in Stage II to compensate for the shortcomings of cross-modal comparison.
Specifically, $f_{I_{m}}^{l}$ is used to get suspected anomaly score by computing similarity with $f_{T}^{Nor}$ and $f_{T}^{Ano}$ as follows:
\begin{equation}
\label{eq:zero_image}
    s_{I_{m}} = \frac{<f_{I_{m}}^{l}, f_{T}^{Ano}>}{{<f_{I_{m}}^{l}, f_{T}^{Ano}>} + {<f_{I_{m}}^{l}, f_{T}^{Nor}>}},
\end{equation}
where $<\cdot,\cdot>$ denotes the cosine similarity. $s_{P_{m}}$ is calculated with $f_{P_{m}}^{l}$, $f_{T}^{Nor}$, and $f_{T}^{Ano}$ in the same way. 
\begin{equation}
\label{eq:zero_pc}
    s_{P_{m}} = \frac{<f_{P_{m}}^{l}, f_{T}^{Ano}>}{{<f_{P_{m}}^{l}, f_{T}^{Ano}>} + {<f_{P_{m}}^{l}, f_{T}^{Nor}>}}.
\end{equation}
Final suspected score $s_{ref}$ combines $s_{I_{m}}$ and $s_{P_{m}}$ together:
\begin{equation}
    \label{eq:s_ref}
    s_{ref} = s_{I_{m}} + s_{P_{m}}.
\end{equation}
We select $N$ normal samples with the smallest $s_{ref}$ as intra-modal references for the next Stage II that is identified as $\left\{R_{I_n}\right\}_{n=1}^N$ and $\left\{R_{P_n}\right\}_{n=1}^N$ in Fig.~\ref{fig:vt_method}.

\subsubsection{Suspected Anomaly Map Computation} \label{sec:suspected}
Furthermore, we have observed that in industrial anomaly detection tasks, anomalies typically constitute only a small fraction of the entire sample. This means that focusing on all small local patch with uniform attention will not effectively facilitate optimal noise sample detection. Consequently, we propose using the preliminary suspected anomaly map obtained from Stage I as the attention map in Noise-Focused Aggregation within Stage II. This strategy allows for differentiated attention across all local patches, enabling our model to more precisely focus on specific local patches that may contain noise.
To generate the preliminary suspected anomaly map, we follow WinCLIP~\cite{jeong2023winclip}, using Harmonic aggregation of windows and multi-scale aggregation to get the suspected anomaly map $W_n \in \mathbb{R}^{h \times w}$ ($n=1,\cdots,N$). 
This suspected anomaly maps $\left\{W_n\right\}_{n=1}^N$ serve as the attention map to enhance the denoising process in Stage II. 

\subsection{Stage II: Enhanced Multi-modal Denoising}
\label{sec:stage2}

In industrial anomaly detection tasks, anomalies often occupy only a small portion of the entire sample. Therefore, after segmenting the sample into multi-scale patches, some patches will contain anomalies while others will not. Naturally, we aim to focus more on those patches containing anomalies and less on those without when computing the suspected anomaly score through intra-modality comparison, to enhance the accuracy of anomaly detection. This is achieved by assigning a weight to each patch based on the suspected anomaly map computed in \cref{sec:suspected}, thereby allowing differential attention to patches based on their likelihood of containing anomalies. Specifically, this process is divided into four steps: 

\noindent
\textbf{Intra-modal comparison.} 
With $N$ intra-modal references selected during Stage I, we employ these image features $\left\{R_{I_n}\right\}_{n=1}^N$ and point cloud features $\left\{R_{P_n}\right\}_{n=1}^N$ for reference:

\begin{equation}
\label{eq:rgb}
    \begin{aligned}
        r_{I_{n}}^{l}, R_{I_{n}}^{l}, R_{I_{n}}^{m}, F_{I_{n}}^{s} &= R_{I_n} \\
        r_{P_{n}}^{l}, R_{P_{n}}^{l}, R_{P_{n}}^{m}, F_{P_{n}}^{s} &= R_{P_n}, 
    \end{aligned}
\end{equation}
where $r_{I_{n}}^{l}$ and $r_{P_{n}}^{l}$ are class tokens, while $R_{I_{n}}^{\sigma} = \left\{r_{u v}^\sigma\right\}_{I_{n}}$ and $R_{P_{n}}^{\sigma} = \left\{r_{u v}^\sigma\right\}_{P_{n}}$ are $\sigma$-scale feature maps. The intra-modality suspected anomaly score is determined by the cosine similarity between the feature vectors of the original query samples and those of intra-modal references: 

\begin{equation}
    \begin{aligned}
    \{\bar{s}_{uv}^{\sigma}\}_{I_{m}} = \{1 - \max < f_{uv}^{\sigma}|I_{m} , r_{u v}^{\sigma}|I_{[1, N]} > \}_{m} \\
    \{\bar{s}_{uv}^{\sigma}\}_{P_{m}} = \{1 - \max < f_{uv}^{\sigma}|P_{m} , r_{u v}^{\sigma}|P_{[1, N]} > \}_{m}, 
    \end{aligned}
\end{equation}
where $\bar{s}_{I_{m}} = \{\bar{s}_{uv}^{\sigma}\}_{I_{m}}$, $\bar{s}_{P_{m}} = \{\bar{s}_{uv}^{\sigma}\}_{P_{m}}$, and $\sigma \in \{l, m, s\}$. 

\noindent
\textbf{Compute weights for local patches.} 
We first compute weight for every local patch. Given the suspected anomaly map $W \in \mathbb{R}^{h \times w}$, we initially procure individual suspected anomaly maps for distinct patches by applying the masks generated in \cref{sec:stage1} to the whole suspected anomaly map.

\begin{equation}
    \begin{aligned}
    \{W_{uv}^{\sigma}\}_{n} &= \{W_{n} \odot M_{uv}^{\sigma}\}, \; \sigma \in \{l, m, s\}. 
    \end{aligned}
\end{equation}
In this way, we can determine the weight for each local patch at both middle and small scales. For large scale, the entire suspected anomaly map can be directly used as the weight.

\begin{figure}
    \centering
    \includegraphics[width=1.0\linewidth]{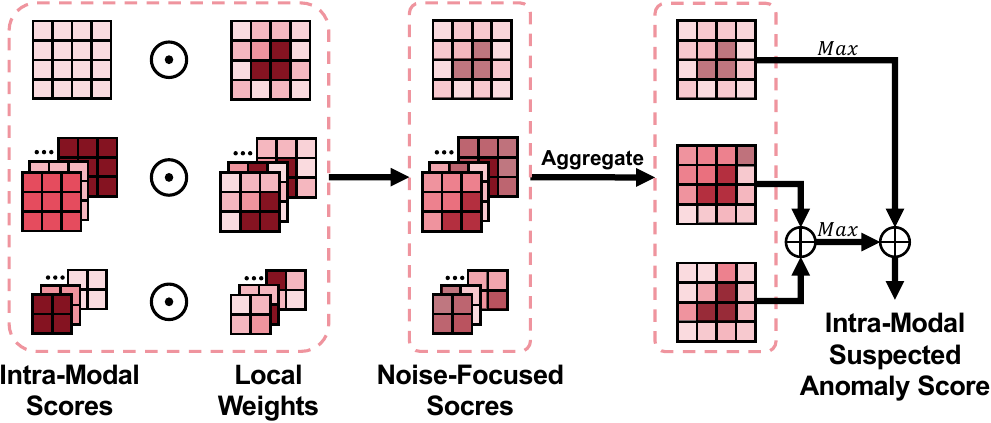}
    \caption{\textbf{Detailed Explanation of multi-scale suspected anomaly score computation,} which focuses more on the patches containing anomalies and less on those without when computing the intra-modal suspected anomaly scores to enhance the accuracy of anomaly detection.}
    \label{fig:vt_aggregation}
\end{figure}

\noindent
\textbf{Multi-scale anomaly score aggregation.} 
For each local patch, the suspected anomaly score $\bar{s}^{\sigma}_{uv}$ is first distributed to every pixel of the local patch. Then at each pixel in the whole point cloud, we aggregate multiple scores from all overlapping local patches to improve anomaly classification. 
In order to focus more on those patches which contain anomalies, we re-weight the score $\bar{s}^{\sigma}_{uv}$ using $W^{\sigma}_{uv}$ while aggregating multi-scale information. In this way, regions will be paid attention based on  their likelihood of containing anomalies (Fig.~\ref{fig:vt_aggregation}-Left): 
\begin{equation}
\label{eq:pc_aggregation}
    \begin{aligned}
    \{\bar{\bar{s}}_{uv}^{\sigma }\}_{I_{m}} = \{ \frac{ \sum_{p,q}  (W_{pq}^{\sigma } \odot \bar{s}_{pq}^{\sigma })_{uv}}{{\sum_{p,q} (M_{pq}^{\sigma }})_{uv} }\}_{I_{m}} \\
    \{\bar{\bar{s}}_{uv}^{\sigma }\}_{P_{m}} = \{ \frac{ \sum_{p,q}  (W_{pq}^{\sigma } \odot \bar{s}_{pq}^{\sigma })_{uv}}{{\sum_{p,q} (M_{pq}^{\sigma }})_{uv} }\}_{P_{m}}.
    \end{aligned}
\end{equation}

\noindent
\textbf{Final suspected anomaly score computation.} 
The final suspected image anomaly score $\tilde{s}_{I_{m}}$ is computed using both cross-modal score $s_{P}$ calculated in \cref{eq:zero_image} and intra-modality score $\{\bar{\bar{s}}_{uv}^{\sigma }\}_{I_{m}} = \{ \{\bar{\bar{s}}_{uv}^{l}\}_{I_{m}}, \{\bar{\bar{s}}_{uv}^{m}\}_{I_{m}}, \{\bar{\bar{s}}_{uv}^{s}\}_{I_{m}} \}$ calculated in \cref{eq:pc_aggregation}:

\begin{equation}
    \tilde{s}_{I_{m}} = \frac{1}{3}(s_{I_{m}} + \max_{uv}\{ \{\bar{\bar{s}}_{uv}^{m}\}_{I_{m}} + \{\bar{\bar{s}}_{uv}^{s}\}_{I_{m}} \} + \max_{uv} \{\bar{\bar{s}}_{uv}^{\l }\}_{I_{m}}).
\end{equation}
Detailed explaination can be viewed in the right part of Fig.~\ref{fig:vt_aggregation}-Left. The final suspected point cloud anomaly score $\tilde{s}_{P_{m}}$ is computed using the same way: 
\begin{equation}
    \tilde{s}_{P_{m}} = \frac{1}{3}(s_{P_{m}} + \max_{uv}\{ \{\bar{\bar{s}}_{uv}^{m}\}_{IP_{m}} + \{\bar{\bar{s}}_{uv}^{s}\}_{P_{m}} \} + \max_{uv} \{\bar{\bar{s}}_{uv}^{\l }\}_{P_{m}}).
\end{equation}
Analogously, the final suspected anomaly score $\tilde{s_{I}}$ is calculated as a weighted combination of $\tilde{s}_{I_{m}}$ and $\tilde{s}_{I_{m}}$, given by the equation: 
\begin{equation}
\tilde{s_{I}} = \lambda_{I} \tilde{s}_{I_{m}} + \lambda_{P} \tilde{s}_{P_{m}}, 
\end{equation}
where $\lambda_{I}$ and $\lambda_{P}$ are hyper-parameters controlling the extent to which RGB and point cloud modalities are integrated. Finally, we remove the samples with top $\tau$ percent scores.

\subsection{Fused Anomaly Detection}
\label{sec:stage3}
\begin{figure}[t]
    \centering
    \includegraphics[width=1.0\linewidth]{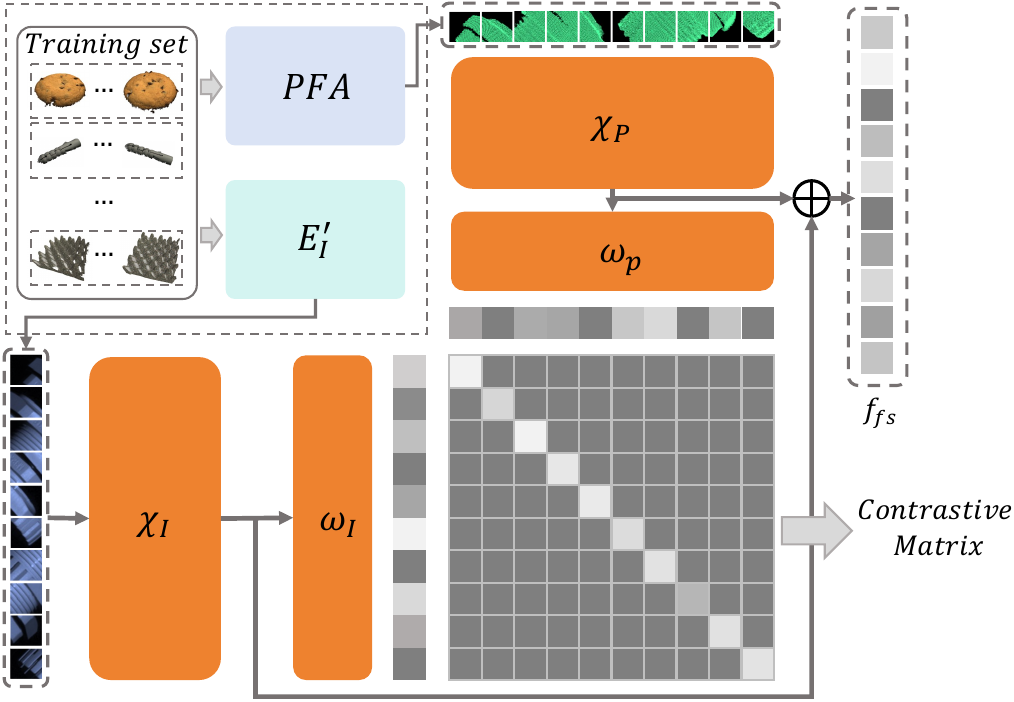}
    \caption{\textbf{Details of Unsupervised Feature Fusion (UFF),} which is a unified module trained with all training data of MVTec 3D-AD. The patch-wise contrastive loss $\mathcal{L}_{con}$ encourages the multimodal patch features in the same position to have the most mutual information, i.e., the diagonal elements of the contrastive matrix have the biggest values.}
    \label{fig:UFF}
\end{figure}
As shown in \cref{fig:vt_method}, Stage III takes in the dataset filtered by Stage I\&II as input and learns its pattern to conduct anomaly detection and segmentation. Besides, Stage III also filters out noise at patch-level in case some hard noise samples still exist in the training dataset.
\subsubsection{Point Feature Alignment}
\label{sec:pfa}

\noindent\textbf{Point Feature Interpolation.}
Post-FPS conducted within the Point Transformer ($E^{\prime}_{P}$), the center points of the point cloud are unevenly distributed, leading to an imbalance in the density of point features. To address this, we interpolate the features back to the original point cloud. With $K$ point features ${g_i}$ corresponding to $K$ center points $c_i$, we employ inverse distance weighting to interpolate the feature for each point $p_j$ in the input point cloud. The interpolation is mathematically represented as:
\begin{equation}
\begin{aligned}
    p^{\prime}_{j} = \sum_{i=1}^{K} \alpha_i g_i, \quad
    \alpha_i = \frac{ \frac{1}{\Vert c_i - p_j \Vert_2+\epsilon}} {\sum_{k=1}^K \sum_{t=1}^{T} \frac{1}{\Vert c_k - p_t \Vert_2+\epsilon}},
    \label{eq:interpolation}
\end{aligned}
\end{equation}
where $\epsilon$ is a small constant to prevent division by zero. 

\noindent\textbf{Point Feature Projection.}
After interpolation, we project the interpolated point features $p^{\prime}_{j}$ onto a 2D plane as $\hat{p}$ using the point coordinates and camera parameters. Noting the sparsity of point clouds, we assign a value of 0 to any 2D plane position lacking a corresponding point. The resulting projected feature map 
matches the size of the RGB image.

\subsubsection{Unsupervised Feature Fusion}
\label{sec:uff}
The interaction between multi-modal features can yield new information beneficial for industrial anomaly detection. For instance, as shown in Fig.~\ref{fig:vt_motivation}, detecting a hole in a cookie necessitates the integration of both its black color and the shape depression. To decipher the intrinsic relationship between these modalities in the training data, we developed the Unsupervised Feature Fusion (UFF) module.

We introduce a patch-wise contrastive loss to train this module. Given RGB features $f_{I}$ and point cloud features $f_{P}$, our goal is to promote a higher correlation of information between features from different modalities at identical spatial positions while minimizing this correlation for features at distinct positions. 

The features of a sample are represented as $\{ \{f_{uv}\}_{I_{i}},\{f_{uv}\}_{P_{i}}\}$, where $i$ denotes the index of the training sample, and $u,v$ represents the patch position. We employ MLP $\{\chi_{I}, \chi_{P} \}$ to derive interaction information between the two modalities and utilize fully connected layers $\{\sigma_{I}, \sigma_{P} \}$ to transform the processed features into query or key vectors, denoted as $\{ \{h_{uv}\}_{I_{i}},\{h_{uv}\}_{P_{i}}\}$. 
For contrastive learning, we apply the InfoNCE loss:
\begin{equation}
\mathcal{L}_{con} = \frac{\{h_{uv}\}_{I_{i}} \cdot \{h_{uv}\}_{P_{i}}}{\sum_{t=1}^{N_b} \sum_{uv} \{h_{uv}\}^{t}_{I} \cdot \{h_{uv}\}^{t}_{P}},
\end{equation}
where $N_b$ is the batch size. 
The UFF module, trained with collective training data from all categories in MVTec 3D-AD, is depicted in \cref{fig:UFF}.

During inference, outputs of the MLP layers are concatenated to form a fused patch feature, denoted as $\{f_{uv}\}_{F_{i}}$.

\subsubsection{Noise Discriminative Coreset Selection}
\label{sec:lof}
In our experimental process, we found that, despite pre-processing the training data to remove noise at the sample level, some noise samples that closely resembled normal samples could not be eliminated. To address this, we conducted a second round of denoising at the patch level. Following Softpatch~\cite{jiang2022softpatch}, we discard noise patches in coreset selection process.
Initially, we calculated outlier scores for all patches. These scores were then aggregated to identify the noise patches, after which we just remove the patches with top $\tau$ percent scores.
We implemented it using the Local Outlier Factor (LOF) method.

LOF is a local-density-based outlier detector. Inspired by Softpatch, we propose to use LOF in M3DM in two ways. Firstly, we will use LOF to rule out noise patches with the aim of making the training datset contain only normal samples. Secondly, we will use the LOF as the soft weight for patches to achieve more accurate anomaly detection.

The k-distance-based absolute local reachability density ${lrd}_{{uv}_{i}}$ is first calculated as:
\begin{equation}
    \begin{gathered}
    {lrd}_{{uv}_{i}}=(\frac{\sum_{b\in\mathcal{N}_k(f_{{uv}_{i}})}dist_k^{reach}(f_{{uv}_{i}},f^b_{uv})}{|\mathcal{N}_k(f_{{uv}_{i}})|})^{-1},\\
    {dist}_k^{reach}(f_{{uv}_{i}},f^b_{uv})=\max (dist_k(f^b_{uv}),d(f_{{uv}_{i}},f^b_{uv})),
    \end{gathered}
\end{equation}
where $d(f_{{uv}_{i}},f^b_{uv})$ is L2-norm, $dist_k(f_{{uv}_{i}})$ is the distance of kth-neighbor, $\mathcal{N}_k(f_{{uv}_{i}})$ is the set of k-nearest neighbors of $f_{{uv}_{i}}$ and $|\mathcal{N}_k(f_{{uv}_{i}})|$ is the number of the set which usually equal k when without repeated neighbors. With the local reachability density of each patch, the overwhelming effect of large clusters is largely reduced. To normalize local density to relative density for treating all clusters equally, the relative density $\eta^i$ of image $i$ is defined below:
\begin{equation}
    \eta_{{uv}_{i}}=\frac{\sum_{b\in\mathcal{N}_k(f_{{uv}_{i}})}{lrd}^b_{uv}}{|\mathcal{N}_k(f_{{uv}_{i}})|\cdot {lrd}_{{uv}_{i}}}.
\end{equation}

$\eta_{{uv}_{i}}$ is the relative density of the neighbors over patch's own, and represents as a patch's confidence of inlier. Patches with top $\tau$ scores are removed before coreset selection.
  
\subsubsection{Decision Layer Fusion}
\label{sec:dlf}
As depicted in Fig.~\ref{fig:vt_motivation}, certain industrial anomalies, such as the protruding part of a potato, manifest exclusively in a single domain, making the correlation between multi-modal features less evident. Additionally, despite the advantages of Feature Fusion in enhancing multi-modal feature interaction, we observed some loss of information during the fusion process. Furthermore, we observed that, despite undergoing denoising at both the image and patch levels, some hard noise patches remain within the dataset. These hard noise elements can adversely affect the precision of anomaly scores during the final inference stage.

To address these issues, we propose utilizing multiple memory banks to preserve the original color feature ($f_{I}$), point cloud feature ($f_{P}$), and fusion feature ($f_{F}$). These are denoted as $\mathcal{M}_{I}$, $\mathcal{M}_{P}$, and $\mathcal{M}_{F}$ respectively. Besides, we propose to use $\eta_{{uv}_{i}}$ obtained in \cref{sec:lof} to re-weight the anomaly score during inference, which can down-weight noisy samples according to outlier scores.
During inference, each bank contributes to predicting an anomaly score and a segmentation map. Two learnable One-Class Support Vector Machines (OCSVMs), $\mathcal{D}_{image}$ and $\mathcal{D}_{pixel}$, are employed to finalize the anomaly score $S_{image}$ and the segmentation map $S_{pixel}$. This procedure is referred to as Decision Layer Fusion (DLF) and can be mathematically represented as follows:
\begin{equation}
    \begin{gathered}
    S_{image} = \mathcal D_{image} (\phi(\mathcal{M}_{I},f_{I}), \phi(\mathcal{M}_{P},f_{P}), \phi(\mathcal{M}_{F},f_{F})), \\
    S_{pixel} = \mathcal D_{pixel} (\psi (\mathcal{M}_{I},f_{I}), \psi(\mathcal{M}_{P},f_{P}), \psi(\mathcal{M}_{F},f_{F})),
    \end{gathered}
\label{eq:dlf_a}
\end{equation}
where $\phi$ and $\psi$ are scoring functions, defined as follows:
\begin{equation}
    \begin{gathered}
        \phi(\mathcal{M}, f) = \eta_{{uv}_{i}}\Vert f^{*}_{{uv}_{i}} - m^*\Vert_2 \\
        \psi(\mathcal{M}, f) = \{\min _{m\in \mathcal{M}} \Vert f_{{uv}_{i}} - m\Vert_2 \Big{|} f_{{uv}_{i}}\in f\} \\
        f^{i, *}_{uv}, m^* = \arg \max _{f_{{uv}_{i}}\in f} \arg \min _{m\in \mathcal{M}}\Vert  f_{{uv}_{i}} - m\Vert_2,
    \end{gathered}
\end{equation}
where $\mathcal{M} \in \{ \mathcal{M}_{I}, \mathcal{M}_{P}, \mathcal{M}_{F} \}$, $f \in \{f_{I}, f_{P}, f_{F}\}$ and $\eta_{{uv}_{i}}$ is the weight parameter obtained in \cref{sec:lof}. 

\section{Experiment}
\begin{table*}[t]
  \centering
  \caption{\textbf{I-AUROC score for regular anomaly detection of all categories of MVTec-3D AD.} Our method maintains the regular anomaly detection ability. The results of baselines are from the ~\cite{mvtec3dad, 3d-ads, ast, benchmarking}. Optimal and sub-optimal results are in \textbf{bold} and {\ul underlined}, respectively.}
  \label{tab:iaucroc_ori}
  \setlength{\tabcolsep}{3.7pt}
\begin{tabular}{cc|cccccccccc|c}
    \toprule
    & Method & Bagel & \begin{tabular}[c]{@{}c@{}}Cable\\ Gland\end{tabular} & Carrot & Cookie & Dowel & Foam & Peach & Potato & Rope & Tire & Mean\\
    \midrule
     \multirow{5}*{\rotatebox{90}{3D}}
    ~&3D-ST\cite{3d-st} & 86.2& 48.4 & 83.2 & 89.4 & 84.8 & 66.3 & 76.3 & 68.7 & \bf{95.8} & 48.6 & 74.8 \\
    ~&FPFH\cite{3d-ads} & 82.5 & 55.1 & {95.2} & 79.7 & {88.3} & 58.2 & 75.8 & 88.9 & 92.9 & 65.3 & 78.2\\
    ~&AST\cite{ast} & {88.1} & 57.6 & {\bf 96.5} & {95.7} & 67.9 & {\bf 79.7} & {\bf 99.0} & {91.5} & \underline{95.6} & 61.1 & {83.3}\\
    &M3DM\cite{wang2023multimodal} & {\ul 94.1} & 65.1 & {\bf 96.5} & {\ul 96.9} & {\bf 90.5} & {76.0} & {88.0} & {\bf 97.4}& 92.6 & {\ul 76.5} & {\bf 87.4}\\
    ~&Ours & {\bf 94.2} & 66.1 & {\ul 95.5} & {\bf 97.2} & {\ul 90.4} & {\ul 77.2} & \underline{88.1} & {\ul 96.4}& 91.6 & {\bf 78.5} & {\bf 87.4}\\
    \midrule
      \multirow{6}*{\rotatebox{90}{RGB}}
    &PADiM\cite{padim} & {\bf 97.5} & 77.5 & 69.8 & 58.2 & 95.9 & 66.3 & 85.8 & 53.5 & 83.2 & 76.0 & 76.4\\
    &PatchCore\cite{patchcore} & 87.6 & 88.0 & 79.1 & 68.2 & 91.2 & 70.1 & 69.5 & 61.8 & 84.1 & 70.2 & 77.0\\
    &STFPM\cite{STFPM} & 93.0 & 84.7 & {89.0} & 57.5 & 94.7 & 76.6 & 71.0 & 59.8 & 96.5 & 70.1 & 79.3\\
    &CS-Flow\cite{cflow} & 94.1 & {\bf 93.0} & 82.7 & \underline{79.5} & {\bf 99.0} & \underline{88.6} & 73.1 & 47.1 & \underline{98.6} & 74.5 & 83.0\\
    &AST\cite{ast} & \underline{ 94.7} & \underline{92.8} & 85.1 & {\bf 82.5} & \underline{98.1} & {\bf 95.1} & {89.5} & 61.3 & {\bf 99.2}& {\bf 82.1} & {\bf 88.0}\\
    &M3DM\cite{wang2023multimodal} & 94.4 & 91.8 & {\bf 89.6} & 74.9 & 95.9 & 76.7 & {\ul 91.9} & {\ul 64.8} & 93.8 & {76.7} & {85.0}\\
    &Ours & 94.2 & 91.7 & {\ul 89.4} & 73.9 & 96.1 & 77.8 & {\bf 93.3} & {\bf 64.9} & 92.8 & \underline{77.7} & \underline{85.1}\\
    \midrule
     \multirow{5}*{\rotatebox{90}{RGB + 3D}}
     &Voxel GAN\cite{mvtec3dad} & 68.0& 32.4& 56.5 & 39.9& 49.7& 48.2& 56.6& 57.9& 60.1& 48.2& 51.7\\
    &PatchCore + FPFH\cite{3d-ads} & 91.8 & 74.8 & 96.7 & 88.3 & \underline{93.2} & 58.2 & 89.6 & \underline{91.2} & 92.1 & {\bf 88.6} & 86.5\\
    &AST\cite{ast} & {98.3} & {87.3} & {\ul 97.6} & {97.1} & {93.2} & {88.5} & {\bf 97.4} & \bf{98.1} & {\bf 100.0} & 79.7 & {93.7} \\
    &M3DM~\cite{wang2023multimodal} & {\bf 99.4} & {\ul 90.9} & {97.2} & {\bf 97.6} & {\bf 96.0} & {\bf 94.2} & \underline{97.3} & 89.9& \underline{97.2} & {85.0} & {\bf 94.5}\\
    &Ours & {\ul 99.3} & {\bf 91.1} & {\bf 97.7} & {\bf 97.6} & {\bf 96.0} & {\ul 92.2} & \underline{97.3} & 89.9& {95.5} & \underline{88.2} & {\bf 94.5}\\
    \bottomrule
  \end{tabular}
\end{table*}
\begin{table*}[t]
  \centering
    \caption{\textbf{AUPRO score for regular anomaly segmentation of all categories of MVTec-3D.} Our method maintains the regular anomaly segmentation ability.  The results of baselines are from the ~\cite{mvtec3dad, 3d-ads, benchmarking}. Optimal and sub-optimal results are in \textbf{bold} and {\ul underlined}, respectively.}
  \label{tab:aupro_ori}
    \setlength{\tabcolsep}{3.7pt}
\begin{tabular}{cc|cccccccccc|c}
    \toprule
    & Method & Bagel & \begin{tabular}[c]{@{}c@{}}Cable\\ Gland\end{tabular} & Carrot & Cookie & Dowel & Foam & Peach & Potato & Rope & Tire & Mean\\
    \midrule
     \multirow{4}*{\rotatebox{90}{3D}}
     ~&3D-ST\cite{3d-st} & \underline{95.0} & 48.3 & {\bf 98.6} & {\bf 92.1} & {\bf 90.5} & 63.2 & 94.5 & {\bf 98.8} & {\bf 97.6} & 54.2 & 83.3\\
     ~&FPFH\cite{3d-ads} & {\bf 97.3} & {\bf 87.9} & {\underline{98.2}} & {\underline{ 90.6}} & {\underline{89.2}} & \underline{73.5} & {\bf 97.7} & {\underline{98.2}} & {\underline{95.6}} & {\bf 96.1} & {\bf 92.4}\\
     ~&M3DM~\cite{wang2023multimodal} & 94.3 & \underline{81.8} & 97.7 & 88.2 & 88.1 & {\bf 74.3} & \underline{95.8} & 97.4 & 95.0 & \underline{92.9} & \underline{90.6} \\
     ~&Ours & 94.2 & \underline{81.8} & 97.8 & 88.3 & 88.0 & {\bf 74.3} & \underline{95.8} & 97.4 & 95.0 & \underline{92.9} & \underline{90.6} \\
    \midrule
     \multirow{5}*{\rotatebox{90}{RGB}}& CFlow\cite{cflow} & 85.5 & 91.9 & \underline{95.8} & 86.7 & \bf 96.9 & 50.0 & 88.9 & 93.5 & 90.4 & 91.9 & 87.1 \\
     & PatchCore\cite{patchcore} & 90.1 & {94.9} & 92.8 & 87.7 & 89.2 & 56.3 & 90.4 & 93.2 & 90.8 & 90.6 & 87.6\\
     &  PADiM\cite{padim} & \bf 98.0 &  94.4 & 94.5 & \underline{92.5} & \underline{96.1} & \underline{79.2} & \underline{96.6} & \underline{94.0} & \underline{93.7} & \underline{91.2} & \underline{93.0}  \\
     & M3DM~\cite{wang2023multimodal} & {95.2}& {\bf 97.2} & {\bf 97.3} & {\bf 89.1} & 93.2 & {\bf 84.3} & {\bf 97.0} & {\bf 95.6} & {\bf 96.8} & {\bf 96.6} & {\bf 94.2} \\
      & Ours & \underline{95.4}& {\ul 97.0} & {\bf 97.3} & {\bf 89.1} & 93.4 & {\bf 84.3} & {\bf 97.0} & {\bf 95.6} & {\bf 96.8} & {\bf 96.6} & {\bf 94.2} \\
      \midrule
     \multirow{4}*{\rotatebox{90}{RGB+3D}}
     &Voxel GAN\cite{mvtec3dad} & 66.4 & 62.0 & 76.6 & 74.0 & 78.3 & 33.2 & 58.2 & 79.0 & 63.3 & 48.3 & 63.9 \\
     &PatchCore + FPFH\cite{3d-ads} & {\bf 97.6} & {96.9} & \bf{97.9} & {\bf 97.3} & {93.3} & {88.8} & {97.5} & \bf{98.1} & {95.0} & {97.1} & {95.9} \\
     &M3DM~\cite{wang2023multimodal} & {97.0} & {\bf 97.1} & \bf{97.9} & \underline{95.0} & {\bf 94.1} & {\ul 93.2} & {\ul 97.7} & \underline{97.1}  & {\ul 97.1} & {\bf 97.5} & {\ul 96.4}\\
      &Ours & \underline{97.4} & {\bf 97.1} & {97.8} & {94.5} & {\ul 93.8} & {\bf 94.7} & {\bf 97.8} & \underline{97.1}  & \bf{97.2} & {\ul 97.4} & {\bf 96.5}\\
    \bottomrule
  \end{tabular}
\end{table*}

\subsection{Experimental Setup}
\label{sec:experiments_detail}

\noindent\textbf{Dataset.}
3D industrial anomaly detection is in the beginning stage. 
The MVTec-3D AD dataset is the first 3D industrial anomaly detection dataset.
Our experiments were performed on the MVTec-3D dataset.
MVTec-3D AD\cite{mvtec3dad} dataset consists of 10 categories, a total of 2,656 training samples, and 1,137 testing samples.
The 3D scans were acquired by an industrial sensor using structured light, and position information was stored in 3 channel tensors representing $x$, $y$ and $z$ coordinates.
Those 3 channel tensors can be single-mapped to the corresponding point clouds.
Additionally, the RGB information is recorded for each point.
Because all samples in the dataset are viewed from the same angle, the RGB information of each sample can be stored in a single image.
Totally, each sample of the MVTec-3D AD dataset contains a colored point cloud.

We conduct both regular anomaly detection in \cref{sec:regular_anomaly} and noisy anomaly detection in \cref{sec:noisy_anomaly}. For noisy anomaly detection, in odrder to generate a noisy training set, we randomly select 10\% anomalous samples from the test set and integrate them into the existing training samples. Additionally, we establish two distinct settings, \textit{Overlap} and \textit{Non-Overlap}, to assess the robustness of our model. In the \textit{Overlap} setting, the anomalous samples added to the training dataset will also be included in the test dataset to demonstrate the risk that defects with similar appearance will severely exacerbate the performance of an anomaly detector trained with noisy data. Conversely, in the \textit{Non-Overlap} setting, these samples will not be retested.

\noindent\textbf{Data Pre-processing.}
Different from 2D data, 3D ones are easier to remove the background information.
Following~\cite{3d-ads}, we estimate the background plane with RANSAC\cite{RANSAC} and any point within 0.005 distance is removed.
At the same time, we set the corresponding pixel of removed points in the RGB image as 0.
This operation not only accelerates the 3D feature processing during training and inference but also reduces the background disturbance for anomaly detection.
Finally, we resize both the position tensor and the RGB image to $224 \times 224$ size, which is matched with the feature extractor input size.

\noindent\textbf{Feature Extractors.}
In Stage I\&II, we use text and image encoder from LAION-2B based CLIP with ViT-H/14 and point cloud encoder from Point-BIND.
In Stage III, we use the ViT-B/8 pretrained on ImageNet\cite{imagenet} with DINO\cite{DINO} as the RGB image encoder and a Point Transformer\cite{point_transformer, pointmae}, which is pretrained on ShapeNet\cite{shapenet} dataset as the 3D point cloud encoder, use the $\{3, 7, 11\}$ layer output as our 3D point cloud feature.

\noindent\textbf{Learnable Module Details.}
Stage I\&II are traing-free and Stage III has 2 learnable modules: the Unsupervised Feature Fusion module and the Decision Layer Fusion module. 
1) For UFF, $\chi_{I}$ and $\chi_{P}$ are 2 two-layer MLPs with $4\times$ hidden dimension as input feature. We use AdamW~\cite{} optimizer with the learning rate as 0.003 and cosine warm-up in 250 steps. Batch size as 16 and we report the best anomaly detection results under 750 UFF training steps. 
2) For DLF, we use two linear OCSVMs~\cite{OCSVM} with SGD~\cite{paszke2019pytorch} optimizers, and the learning rate is set as $1\times10^{-4}$ and each class is trained for 1000 steps.

\noindent\textbf{Evaluation Metrics.}
All evaluation metrics are exactly the same as in~\cite{mvtec3dad}.
We evaluate the image-level anomaly detection performance with the area under the receiver operator curve (I-AUROC), and higher I-AUROC means better image-level anomaly detection performance. 
For segmentation evaluation, we use the per-region Overlap (AUPRO) metric, 
which is defined as the average relative Overlap of the binary prediction with each connected component of the ground truth.
Similar to I-AUROC, the receiver operator curve of pixel level predictions can be used to calculate P-AUROC for evaluating the segmentation performance.
\begin{table*}[t]
\begin{center}
\caption{\textbf{I-AUROC score for anomaly detection under \textit{Overlap} setting of all categories in MVTec 3D-AD.} Our method clearly outperforms other methods in 3D, RGB, and 3D + RGB settings, indicating the superior anomaly detection ability of our method. We report the mean and standard deviation over 3 random seeds for each measurement. Optimal and sub-optimal results are in \textbf{bold} and {\ul underlined}, respectively.}
\vspace{-10pt}
\label{tab:i-auroc overlap}
\setlength{\tabcolsep}{4.7pt}
\begin{tabular}{cc|cccccccccc|c}
\toprule
\textbf{}               &{Method}         & Bagel                & \begin{tabular}[c]{@{}c@{}}Cable\\ Gland\end{tabular}          & Carrot               & Cookie               & Dowel                & Foam                 & Peach                & Potato               & Rope                 & Tire                 & Mean                \\ \midrule
\multirow{6}{*}{\rotatebox{90}{3D}}     &{SIFT}           & 50.0{\fontsize{6.5pt}{7pt}\selectfont$\pm$0.8}          & 48.5{\fontsize{6.5pt}{7pt}\selectfont$\pm$1.9}          & 67.8{\fontsize{6.5pt}{7pt}\selectfont$\pm$0.2}          & 58.1{\fontsize{6.5pt}{7pt}\selectfont$\pm$0.4}          & 58.2{\fontsize{6.5pt}{7pt}\selectfont$\pm$3.8}          & 49.2{\fontsize{6.5pt}{7pt}\selectfont$\pm$2.8}          & 40.5{\fontsize{6.5pt}{7pt}\selectfont$\pm$0.6}          & 47.0{\fontsize{6.5pt}{7pt}\selectfont$\pm$1.3}          & 43.3{\fontsize{6.5pt}{7pt}\selectfont$\pm$1.1}          & 45.0{\fontsize{6.5pt}{7pt}\selectfont$\pm$2.7}          & 50.8{\fontsize{6.5pt}{7pt}\selectfont$\pm$0.5}          \\
                        &{FPFH}           & 53.4{\fontsize{6.5pt}{7pt}\selectfont$\pm$2.8}          & 40.9{\fontsize{6.5pt}{7pt}\selectfont$\pm$3.2}          & 71.4{\fontsize{6.5pt}{7pt}\selectfont$\pm$1.2}          & 62.7{\fontsize{6.5pt}{7pt}\selectfont$\pm$0.8}          & 64.5{\fontsize{6.5pt}{7pt}\selectfont$\pm$2.4}          & 38.5{\fontsize{6.5pt}{7pt}\selectfont$\pm$0.3}          & 46.8{\fontsize{6.5pt}{7pt}\selectfont$\pm$2.6}          & 45.3{\fontsize{6.5pt}{7pt}\selectfont$\pm$1.5}          & 52.2{\fontsize{6.5pt}{7pt}\selectfont$\pm$1.5}          & 51.5{\fontsize{6.5pt}{7pt}\selectfont$\pm$4.2}          & 52.7{\fontsize{6.5pt}{7pt}\selectfont$\pm$0.3}          \\
                        &{AST}            & 61.0{\fontsize{6.5pt}{7pt}\selectfont$\pm$0.6}          & 38.4{\fontsize{6.5pt}{7pt}\selectfont$\pm$0.6}          & {\ul 72.9{\fontsize{6.5pt}{7pt}\selectfont$\pm$0.6}}    & 75.2{\fontsize{6.5pt}{7pt}\selectfont$\pm$0.6}          & 47.8{\fontsize{6.5pt}{7pt}\selectfont$\pm$0.6}          & 55.7{\fontsize{6.5pt}{7pt}\selectfont$\pm$0.6}          & {\ul 66.9{\fontsize{6.5pt}{7pt}\selectfont$\pm$0.6}}    & 60.6{\fontsize{6.5pt}{7pt}\selectfont$\pm$0.6}          & 55.5{\fontsize{6.5pt}{7pt}\selectfont$\pm$1.0}          & 49.2{\fontsize{6.5pt}{7pt}\selectfont$\pm$0.6}          & 58.3{\fontsize{6.5pt}{7pt}\selectfont$\pm$0.2}          \\
                        &{Shape-Guided}   & 66.1{\fontsize{6.5pt}{7pt}\selectfont$\pm$5.1}          & {\ul 58.7{\fontsize{6.5pt}{7pt}\selectfont$\pm$10.4}}   & 71.4{\fontsize{6.5pt}{7pt}\selectfont$\pm$6.0}          & {\ul 76.4{\fontsize{6.5pt}{7pt}\selectfont$\pm$1.4}}    & 71.6{\fontsize{6.5pt}{7pt}\selectfont$\pm$0.7}          & 54.1{\fontsize{6.5pt}{7pt}\selectfont$\pm$3.1}          & 61.0{\fontsize{6.5pt}{7pt}\selectfont$\pm$4.5}          & 59.3{\fontsize{6.5pt}{7pt}\selectfont$\pm$5.7}          & 60.7{\fontsize{6.5pt}{7pt}\selectfont$\pm$4.5}          & 64.3{\fontsize{6.5pt}{7pt}\selectfont$\pm$7.4}          & 64.4{\fontsize{6.5pt}{7pt}\selectfont$\pm$1.9}          \\
                        &{M3DM}           & {\ul 74.0{\fontsize{6.5pt}{7pt}\selectfont$\pm$0.7}}    & 56.7{\fontsize{6.5pt}{7pt}\selectfont$\pm$1.8}          & 72.2{\fontsize{6.5pt}{7pt}\selectfont$\pm$1.7}          & 74.5{\fontsize{6.5pt}{7pt}\selectfont$\pm$0.6}          & {\ul 77.4{\fontsize{6.5pt}{7pt}\selectfont$\pm$0.7}}    & {\ul 62.3{\fontsize{6.5pt}{7pt}\selectfont$\pm$0.6}}    & 56.2{\fontsize{6.5pt}{7pt}\selectfont$\pm$1.9}          & {\ul 64.1{\fontsize{6.5pt}{7pt}\selectfont$\pm$0.5}}    & {\ul 72.5{\fontsize{6.5pt}{7pt}\selectfont$\pm$0.5}}    & {\ul 74.3{\fontsize{6.5pt}{7pt}\selectfont$\pm$1.8}}    & {\ul 68.4{\fontsize{6.5pt}{7pt}\selectfont$\pm$0.7}}    \\
                        &{Ours}           & \textbf{93.5{\fontsize{6.5pt}{7pt}\selectfont$\pm$1.6}} & \textbf{71.8{\fontsize{6.5pt}{7pt}\selectfont$\pm$1.3}} & \textbf{93.8{\fontsize{6.5pt}{7pt}\selectfont$\pm$0.7}} & \textbf{91.1{\fontsize{6.5pt}{7pt}\selectfont$\pm$2.3}} & \textbf{78.0{\fontsize{6.5pt}{7pt}\selectfont$\pm$2.7}} & \textbf{67.2{\fontsize{6.5pt}{7pt}\selectfont$\pm$3.2}} & \textbf{79.9{\fontsize{6.5pt}{7pt}\selectfont$\pm$1.4}} & \textbf{79.9{\fontsize{6.5pt}{7pt}\selectfont$\pm$2.2}} & \textbf{87.9{\fontsize{6.5pt}{7pt}\selectfont$\pm$0.4}} & \textbf{79.8{\fontsize{6.5pt}{7pt}\selectfont$\pm$3.5}} & \textbf{82.3{\fontsize{6.5pt}{7pt}\selectfont$\pm$0.4}} \\ \midrule
\multirow{7}{*}{\rotatebox{90}{RGB}}    &{PaDim}          & 70.8{\fontsize{6.5pt}{7pt}\selectfont$\pm$0.7}          & 57.3{\fontsize{6.5pt}{7pt}\selectfont$\pm$2.6}          & 54.7{\fontsize{6.5pt}{7pt}\selectfont$\pm$0.5}          & 43.2{\fontsize{6.5pt}{7pt}\selectfont$\pm$1.6}          & 72.1{\fontsize{6.5pt}{7pt}\selectfont$\pm$0.3}          & 55.4{\fontsize{6.5pt}{7pt}\selectfont$\pm$2.2}          & 61.7{\fontsize{6.5pt}{7pt}\selectfont$\pm$0.3}          & 36.8{\fontsize{6.5pt}{7pt}\selectfont$\pm$1.3}          & 74.8{\fontsize{6.5pt}{7pt}\selectfont$\pm$2.5}          & 55.2{\fontsize{6.5pt}{7pt}\selectfont$\pm$1.5}          & 58.2{\fontsize{6.5pt}{7pt}\selectfont$\pm$0.4}          \\
                        &{PatchCore}      & 64.9{\fontsize{6.5pt}{7pt}\selectfont$\pm$0.7}          & 71.4{\fontsize{6.5pt}{7pt}\selectfont$\pm$0.9}          & 71.5{\fontsize{6.5pt}{7pt}\selectfont$\pm$1.5}          & 52.5{\fontsize{6.5pt}{7pt}\selectfont$\pm$2.2}          & 73.3{\fontsize{6.5pt}{7pt}\selectfont$\pm$1.2}          & 56.5{\fontsize{6.5pt}{7pt}\selectfont$\pm$2.9}          & 46.6{\fontsize{6.5pt}{7pt}\selectfont$\pm$1.1}          & 36.8{\fontsize{6.5pt}{7pt}\selectfont$\pm$0.4}          & 54.2{\fontsize{6.5pt}{7pt}\selectfont$\pm$1.3}          & 57.2{\fontsize{6.5pt}{7pt}\selectfont$\pm$1.3}          & 58.5{\fontsize{6.5pt}{7pt}\selectfont$\pm$0.4}          \\
                        &{AST}            & 57.6{\fontsize{6.5pt}{7pt}\selectfont$\pm$0.6}          & 62.2{\fontsize{6.5pt}{7pt}\selectfont$\pm$0.0}          & 50.7{\fontsize{6.5pt}{7pt}\selectfont$\pm$0.0}          & 47.5{\fontsize{6.5pt}{7pt}\selectfont$\pm$0.6}          & 58.8{\fontsize{6.5pt}{7pt}\selectfont$\pm$0.0}          & 56.0{\fontsize{6.5pt}{7pt}\selectfont$\pm$0.0}          & 54.6{\fontsize{6.5pt}{7pt}\selectfont$\pm$0.0}          & 43.7{\fontsize{6.5pt}{7pt}\selectfont$\pm$0.6}          & 42.8{\fontsize{6.5pt}{7pt}\selectfont$\pm$0.0}          & 44.6{\fontsize{6.5pt}{7pt}\selectfont$\pm$0.6}          & 51.8{\fontsize{6.5pt}{7pt}\selectfont$\pm$0.2}          \\
                        &{Shape-Guided}   & 62.7{\fontsize{6.5pt}{7pt}\selectfont$\pm$4.4}          & 64.3{\fontsize{6.5pt}{7pt}\selectfont$\pm$9.3}          & 66.9{\fontsize{6.5pt}{7pt}\selectfont$\pm$7.3}          & 57.3{\fontsize{6.5pt}{7pt}\selectfont$\pm$16.4}         & 72.1{\fontsize{6.5pt}{7pt}\selectfont$\pm$0.9}          & 51.5{\fontsize{6.5pt}{7pt}\selectfont$\pm$3.2}          & 52.9{\fontsize{6.5pt}{7pt}\selectfont$\pm$10.0}         & {\ul 50.3{\fontsize{6.5pt}{7pt}\selectfont$\pm$11.1}}   & 50.5{\fontsize{6.5pt}{7pt}\selectfont$\pm$9.4}          & 58.2{\fontsize{6.5pt}{7pt}\selectfont$\pm$9.3}          & 58.7{\fontsize{6.5pt}{7pt}\selectfont$\pm$5.8}          \\
                        &{SoftPatch}      & {\ul 88.8{\fontsize{6.5pt}{7pt}\selectfont$\pm$1.1}}    & {\ul 87.3{\fontsize{6.5pt}{7pt}\selectfont$\pm$2.2}}    & {\ul 84.9{\fontsize{6.5pt}{7pt}\selectfont$\pm$1.3}}    & {\ul 63.3{\fontsize{6.5pt}{7pt}\selectfont$\pm$1.2}}    & \textbf{96.5{\fontsize{6.5pt}{7pt}\selectfont$\pm$0.8}} & {\ul 75.0{\fontsize{6.5pt}{7pt}\selectfont$\pm$1.6}}    & {\ul 62.3{\fontsize{6.5pt}{7pt}\selectfont$\pm$0.7}}    & 43.6{\fontsize{6.5pt}{7pt}\selectfont$\pm$2.1}          & {\ul 89.3{\fontsize{6.5pt}{7pt}\selectfont$\pm$1.4}}    & {\ul 71.0{\fontsize{6.5pt}{7pt}\selectfont$\pm$0.9}}    & {\ul 76.2{\fontsize{6.5pt}{7pt}\selectfont$\pm$0.3}}    \\
                        &{M3DM}           & 64.1{\fontsize{6.5pt}{7pt}\selectfont$\pm$1.4}          & 62.1{\fontsize{6.5pt}{7pt}\selectfont$\pm$2.1}          & 65.5{\fontsize{6.5pt}{7pt}\selectfont$\pm$0.9}          & 53.6{\fontsize{6.5pt}{7pt}\selectfont$\pm$2.1}          & 70.7{\fontsize{6.5pt}{7pt}\selectfont$\pm$0.9}          & 57.0{\fontsize{6.5pt}{7pt}\selectfont$\pm$1.2}          & 54.7{\fontsize{6.5pt}{7pt}\selectfont$\pm$2.0}          & 42.1{\fontsize{6.5pt}{7pt}\selectfont$\pm$2.3}          & 53.8{\fontsize{6.5pt}{7pt}\selectfont$\pm$1.1}          & 58.3{\fontsize{6.5pt}{7pt}\selectfont$\pm$0.9}          & 58.2{\fontsize{6.5pt}{7pt}\selectfont$\pm$0.5}          \\
                        &{Ours}           & \textbf{90.3{\fontsize{6.5pt}{7pt}\selectfont$\pm$0.4}} & \textbf{87.5{\fontsize{6.5pt}{7pt}\selectfont$\pm$3.4}} & \textbf{86.5{\fontsize{6.5pt}{7pt}\selectfont$\pm$1.8}} & \textbf{67.1{\fontsize{6.5pt}{7pt}\selectfont$\pm$4.6}} & {\ul 86.1{\fontsize{6.5pt}{7pt}\selectfont$\pm$0.6}}    & \textbf{79.2{\fontsize{6.5pt}{7pt}\selectfont$\pm$2.8}} & \textbf{84.4{\fontsize{6.5pt}{7pt}\selectfont$\pm$2.3}} & \textbf{54.6{\fontsize{6.5pt}{7pt}\selectfont$\pm$6.2}} & \textbf{90.0{\fontsize{6.5pt}{7pt}\selectfont$\pm$2.2}} & \textbf{73.1{\fontsize{6.5pt}{7pt}\selectfont$\pm$1.1}} & \textbf{79.9{\fontsize{6.5pt}{7pt}\selectfont$\pm$0.4}} \\ \midrule
\multirow{5}{*}{\rotatebox{90}{3D+RGB}} &{PatchCore+FPFH} & 61.3{\fontsize{6.5pt}{7pt}\selectfont$\pm$2.7}          & 58.3{\fontsize{6.5pt}{7pt}\selectfont$\pm$0.9}          & 72.3{\fontsize{6.5pt}{7pt}\selectfont$\pm$0.4}          & 69.0{\fontsize{6.5pt}{7pt}\selectfont$\pm$1.1}          & 67.2{\fontsize{6.5pt}{7pt}\selectfont$\pm$1.0}          & 47.1{\fontsize{6.5pt}{7pt}\selectfont$\pm$1.9}          & 53.0{\fontsize{6.5pt}{7pt}\selectfont$\pm$2.0}          & 52.1{\fontsize{6.5pt}{7pt}\selectfont$\pm$1.3}          & 52.7{\fontsize{6.5pt}{7pt}\selectfont$\pm$1.0}          & 68.2{\fontsize{6.5pt}{7pt}\selectfont$\pm$0.8}          & 60.1{\fontsize{6.5pt}{7pt}\selectfont$\pm$0.4}          \\
                        &{AST}            & 65.3{\fontsize{6.5pt}{7pt}\selectfont$\pm$0.6}          & {\ul 69.5{\fontsize{6.5pt}{7pt}\selectfont$\pm$0.6}}    & 73.8{\fontsize{6.5pt}{7pt}\selectfont$\pm$0.6}          & {\ul 83.1{\fontsize{6.5pt}{7pt}\selectfont$\pm$0.0}}    & 68.1{\fontsize{6.5pt}{7pt}\selectfont$\pm$0.6}          & {\ul 64.4{\fontsize{6.5pt}{7pt}\selectfont$\pm$0.6}}    & {\ul 64.7{\fontsize{6.5pt}{7pt}\selectfont$\pm$0.6}}    & {\ul 64.1{\fontsize{6.5pt}{7pt}\selectfont$\pm$0.6}}    & 49.7{\fontsize{6.5pt}{7pt}\selectfont$\pm$0.6}          & 55.8{\fontsize{6.5pt}{7pt}\selectfont$\pm$0.0}          & 65.8{\fontsize{6.5pt}{7pt}\selectfont$\pm$0.0}          \\
                        &{Shape-Guided}   & 69.1{\fontsize{6.5pt}{7pt}\selectfont$\pm$0.7}          & 67.2{\fontsize{6.5pt}{7pt}\selectfont$\pm$1.4}          & {\ul 76.3{\fontsize{6.5pt}{7pt}\selectfont$\pm$0.5}}    & 71.3{\fontsize{6.5pt}{7pt}\selectfont$\pm$0.8}          & 71.8{\fontsize{6.5pt}{7pt}\selectfont$\pm$0.3}          & 58.0{\fontsize{6.5pt}{7pt}\selectfont$\pm$0.3}          & 62.0{\fontsize{6.5pt}{7pt}\selectfont$\pm$0.3}          & 60.4{\fontsize{6.5pt}{7pt}\selectfont$\pm$0.7}          & 55.3{\fontsize{6.5pt}{7pt}\selectfont$\pm$0.3}          & 67.8{\fontsize{6.5pt}{7pt}\selectfont$\pm$0.6}          & 65.9{\fontsize{6.5pt}{7pt}\selectfont$\pm$0.2}          \\
                        &{M3DM}           & {\ul 72.5{\fontsize{6.5pt}{7pt}\selectfont$\pm$2.2}}    & 62.4{\fontsize{6.5pt}{7pt}\selectfont$\pm$0.8}          & 69.6{\fontsize{6.5pt}{7pt}\selectfont$\pm$1.4}          & 72.4{\fontsize{6.5pt}{7pt}\selectfont$\pm$2.1}          & {\ul 73.9{\fontsize{6.5pt}{7pt}\selectfont$\pm$0.9}}    & 64.3{\fontsize{6.5pt}{7pt}\selectfont$\pm$2.0}          & 60.1{\fontsize{6.5pt}{7pt}\selectfont$\pm$0.3}          & 54.0{\fontsize{6.5pt}{7pt}\selectfont$\pm$2.0}          & {\ul 62.1{\fontsize{6.5pt}{7pt}\selectfont$\pm$1.8}}    & {\ul 71.4{\fontsize{6.5pt}{7pt}\selectfont$\pm$2.1}}    & {\ul 66.3{\fontsize{6.5pt}{7pt}\selectfont$\pm$0.5}}    \\
                        &{Ours}           & \textbf{96.7{\fontsize{6.5pt}{7pt}\selectfont$\pm$2.1}} & \textbf{86.2{\fontsize{6.5pt}{7pt}\selectfont$\pm$3.0}} & \textbf{95.5{\fontsize{6.5pt}{7pt}\selectfont$\pm$1.3}} & \textbf{90.3{\fontsize{6.5pt}{7pt}\selectfont$\pm$3.4}} & \textbf{86.0{\fontsize{6.5pt}{7pt}\selectfont$\pm$3.0}} & \textbf{79.1{\fontsize{6.5pt}{7pt}\selectfont$\pm$3.7}} & \textbf{86.6{\fontsize{6.5pt}{7pt}\selectfont$\pm$3.7}} & \textbf{72.2{\fontsize{6.5pt}{7pt}\selectfont$\pm$3.3}} & \textbf{92.0{\fontsize{6.5pt}{7pt}\selectfont$\pm$0.5}} & \textbf{81.3{\fontsize{6.5pt}{7pt}\selectfont$\pm$1.6}} & \textbf{86.6{\fontsize{6.5pt}{7pt}\selectfont$\pm$1.3}}                        \\ \bottomrule
\end{tabular}
\end{center}
\vspace{-15pt}
\end{table*}

\begin{table*}[t]
\begin{center}
\caption{\textbf{AUPRO score for anomaly segmentation under \textit{Overlap} setting of all categories in MVTec 3D-AD.} Our method clearly outperforms other methods in 3D, RGB, and 3D + RGB settings, indicating the superior anomaly segmentation ability of our method. We report the mean and standard deviation over 3 random seeds for each measurement. Optimal and sub-optimal results are in \textbf{bold} and {\ul underlined}, respectively.}
\label{tab:aupro overlap}
\setlength{\tabcolsep}{4.7pt}
\begin{tabular}{cc|cccccccccc|c}
\toprule
\textbf{}               &{Method}         & Bagel                & \begin{tabular}[c]{@{}c@{}}Cable\\ Gland\end{tabular}          & Carrot               & Cookie               & Dowel                & Foam                 & Peach                & Potato               & Rope                 & Tire                 & Mean                \\ \midrule
\multirow{5}{*}{\rotatebox{90}{3D}}     &{SIFT}           & 69.1{\fontsize{6.5pt}{7pt}\selectfont$\pm$1.6}          & 68.2{\fontsize{6.5pt}{7pt}\selectfont$\pm$0.8}          & 85.3{\fontsize{6.5pt}{7pt}\selectfont$\pm$0.4}          & 72.3{\fontsize{6.5pt}{7pt}\selectfont$\pm$0.8}          & 67.1{\fontsize{6.5pt}{7pt}\selectfont$\pm$1.4}          & 55.7{\fontsize{6.5pt}{7pt}\selectfont$\pm$1.5}          & 64.3{\fontsize{6.5pt}{7pt}\selectfont$\pm$1.4}          & 66.6{\fontsize{6.5pt}{7pt}\selectfont$\pm$1.7}          & 69.9{\fontsize{6.5pt}{7pt}\selectfont$\pm$0.8}          & 72.6{\fontsize{6.5pt}{7pt}\selectfont$\pm$1.2}          & 69.1{\fontsize{6.5pt}{7pt}\selectfont$\pm$0.4}          \\
                        &{FPFH}           & 70.5{\fontsize{6.5pt}{7pt}\selectfont$\pm$1.6}          & 73.7{\fontsize{6.5pt}{7pt}\selectfont$\pm$0.6}          & 88.5{\fontsize{6.5pt}{7pt}\selectfont$\pm$0.2}          & 72.6{\fontsize{6.5pt}{7pt}\selectfont$\pm$0.8}          & 72.6{\fontsize{6.5pt}{7pt}\selectfont$\pm$2.7}          & 56.7{\fontsize{6.5pt}{7pt}\selectfont$\pm$2.4}          & 66.7{\fontsize{6.5pt}{7pt}\selectfont$\pm$1.6}          & 75.0{\fontsize{6.5pt}{7pt}\selectfont$\pm$2.2}          & 65.5{\fontsize{6.5pt}{7pt}\selectfont$\pm$1.8}          & 77.2{\fontsize{6.5pt}{7pt}\selectfont$\pm$1.3}          & 71.9{\fontsize{6.5pt}{7pt}\selectfont$\pm$0.4}          \\
                        &{Shape-Guided}   & 74.6{\fontsize{6.5pt}{7pt}\selectfont$\pm$0.6}          & \textbf{83.7{\fontsize{6.5pt}{7pt}\selectfont$\pm$2.2}} & \textbf{98.1{\fontsize{6.5pt}{7pt}\selectfont$\pm$0.1}} & {\ul 81.9{\fontsize{6.5pt}{7pt}\selectfont$\pm$5.4}}    & \textbf{88.6{\fontsize{6.5pt}{7pt}\selectfont$\pm$0.1}} & \textbf{80.4{\fontsize{6.5pt}{7pt}\selectfont$\pm$6.7}} & {\ul 88.9{\fontsize{6.5pt}{7pt}\selectfont$\pm$7.3}}    & 88.2{\fontsize{6.5pt}{7pt}\selectfont$\pm$0.0}          & 88.7{\fontsize{6.5pt}{7pt}\selectfont$\pm$3.6}          & \textbf{93.7{\fontsize{6.5pt}{7pt}\selectfont$\pm$5.5}} & {\ul 86.7{\fontsize{6.5pt}{7pt}\selectfont$\pm$1.7}}    \\
                        &{M3DM}           & {\ul 84.0{\fontsize{6.5pt}{7pt}\selectfont$\pm$1.0}}    & {\ul 79.7{\fontsize{6.5pt}{7pt}\selectfont$\pm$1.1}}    & 95.8{\fontsize{6.5pt}{7pt}\selectfont$\pm$0.4}          & 79.6{\fontsize{6.5pt}{7pt}\selectfont$\pm$1.3}          & {\ul 85.5{\fontsize{6.5pt}{7pt}\selectfont$\pm$0.6}}    & {\ul 68.3{\fontsize{6.5pt}{7pt}\selectfont$\pm$1.6}}    & 86.4{\fontsize{6.5pt}{7pt}\selectfont$\pm$0.9}          & \textbf{91.3{\fontsize{6.5pt}{7pt}\selectfont$\pm$0.8}} & {\ul 90.3{\fontsize{6.5pt}{7pt}\selectfont$\pm$1.5}}    & 88.7{\fontsize{6.5pt}{7pt}\selectfont$\pm$0.4}          & 85.0{\fontsize{6.5pt}{7pt}\selectfont$\pm$0.4}          \\
                        &{Ours}           & \textbf{95.0{\fontsize{6.5pt}{7pt}\selectfont$\pm$1.3}} & 78.8{\fontsize{6.5pt}{7pt}\selectfont$\pm$0.8}          & {\ul 97.2{\fontsize{6.5pt}{7pt}\selectfont$\pm$0.1}}    & \textbf{84.5{\fontsize{6.5pt}{7pt}\selectfont$\pm$1.4}} & 83.9{\fontsize{6.5pt}{7pt}\selectfont$\pm$3.0}          & 66.6{\fontsize{6.5pt}{7pt}\selectfont$\pm$2.4}          & \textbf{91.2{\fontsize{6.5pt}{7pt}\selectfont$\pm$1.6}} & {\ul 89.9{\fontsize{6.5pt}{7pt}\selectfont$\pm$0.6}}    & \textbf{92.7{\fontsize{6.5pt}{7pt}\selectfont$\pm$0.5}} & {\ul 89.9{\fontsize{6.5pt}{7pt}\selectfont$\pm$0.7}}    & \textbf{87.0{\fontsize{6.5pt}{7pt}\selectfont$\pm$0.2}} \\ \midrule
\multirow{6}{*}{\rotatebox{90}{RGB}}    &{PaDim}          & 77.9{\fontsize{6.5pt}{7pt}\selectfont$\pm$2.7}          & 79.9{\fontsize{6.5pt}{7pt}\selectfont$\pm$3.8}          & {\ul 91.8{\fontsize{6.5pt}{7pt}\selectfont$\pm$0.2}}    & 72.2{\fontsize{6.5pt}{7pt}\selectfont$\pm$1.3}          & {\ul 90.0{\fontsize{6.5pt}{7pt}\selectfont$\pm$0.7}}    & \textbf{92.4{\fontsize{6.5pt}{7pt}\selectfont$\pm$1.9}} & \textbf{91.4{\fontsize{6.5pt}{7pt}\selectfont$\pm$1.2}} & \textbf{92.6{\fontsize{6.5pt}{7pt}\selectfont$\pm$1.2}} & {\ul 91.3{\fontsize{6.5pt}{7pt}\selectfont$\pm$1.3}}    & \textbf{92.2{\fontsize{6.5pt}{7pt}\selectfont$\pm$0.8}} & {\ul 87.2{\fontsize{6.5pt}{7pt}\selectfont$\pm$0.7}}    \\
                        &{PatchCore}      & 67.1{\fontsize{6.5pt}{7pt}\selectfont$\pm$1.7}          & 73.3{\fontsize{6.5pt}{7pt}\selectfont$\pm$0.0}          & 77.0{\fontsize{6.5pt}{7pt}\selectfont$\pm$0.3}          & 72.1{\fontsize{6.5pt}{7pt}\selectfont$\pm$0.8}          & 69.9{\fontsize{6.5pt}{7pt}\selectfont$\pm$1.2}          & 59.1{\fontsize{6.5pt}{7pt}\selectfont$\pm$2.4}          & 61.7{\fontsize{6.5pt}{7pt}\selectfont$\pm$1.2}          & 64.3{\fontsize{6.5pt}{7pt}\selectfont$\pm$1.1}          & 56.1{\fontsize{6.5pt}{7pt}\selectfont$\pm$1.6}          & 73.1{\fontsize{6.5pt}{7pt}\selectfont$\pm$1.2}          & 67.4{\fontsize{6.5pt}{7pt}\selectfont$\pm$0.8}          \\
                        &{Shape-Guided}   & 67.5{\fontsize{6.5pt}{7pt}\selectfont$\pm$0.6}          & 73.9{\fontsize{6.5pt}{7pt}\selectfont$\pm$0.7}          & 81.2{\fontsize{6.5pt}{7pt}\selectfont$\pm$0.1}          & 72.1{\fontsize{6.5pt}{7pt}\selectfont$\pm$0.1}          & 76.1{\fontsize{6.5pt}{7pt}\selectfont$\pm$0.6}          & 56.0{\fontsize{6.5pt}{7pt}\selectfont$\pm$0.0}          & 62.5{\fontsize{6.5pt}{7pt}\selectfont$\pm$0.2}          & 71.6{\fontsize{6.5pt}{7pt}\selectfont$\pm$1.0}          & 64.7{\fontsize{6.5pt}{7pt}\selectfont$\pm$0.5}          & 73.8{\fontsize{6.5pt}{7pt}\selectfont$\pm$0.1}          & 69.9{\fontsize{6.5pt}{7pt}\selectfont$\pm$0.1}          \\
                        &{SoftPatch}      & {\ul 83.9{\fontsize{6.5pt}{7pt}\selectfont$\pm$2.0}}    & {\ul 89.3{\fontsize{6.5pt}{7pt}\selectfont$\pm$2.7}}    & 91.4{\fontsize{6.5pt}{7pt}\selectfont$\pm$0.5}          & {\ul 79.2{\fontsize{6.5pt}{7pt}\selectfont$\pm$0.7}}    & \textbf{91.8{\fontsize{6.5pt}{7pt}\selectfont$\pm$1.8}} & 72.4{\fontsize{6.5pt}{7pt}\selectfont$\pm$2.8}          & 76.5{\fontsize{6.5pt}{7pt}\selectfont$\pm$2.4}          & 72.9{\fontsize{6.5pt}{7pt}\selectfont$\pm$2.7}          & 89.8{\fontsize{6.5pt}{7pt}\selectfont$\pm$2.6}          & 90.1{\fontsize{6.5pt}{7pt}\selectfont$\pm$1.7}          & 83.7{\fontsize{6.5pt}{7pt}\selectfont$\pm$0.3}          \\
                        &{M3DM}           & 68.6{\fontsize{6.5pt}{7pt}\selectfont$\pm$1.7}          & 72.7{\fontsize{6.5pt}{7pt}\selectfont$\pm$0.8}          & 77.4{\fontsize{6.5pt}{7pt}\selectfont$\pm$0.3}          & 70.5{\fontsize{6.5pt}{7pt}\selectfont$\pm$0.6}          & 68.6{\fontsize{6.5pt}{7pt}\selectfont$\pm$1.3}          & 59.8{\fontsize{6.5pt}{7pt}\selectfont$\pm$1.4}          & 64.9{\fontsize{6.5pt}{7pt}\selectfont$\pm$1.4}          & 65.0{\fontsize{6.5pt}{7pt}\selectfont$\pm$1.4}          & 57.0{\fontsize{6.5pt}{7pt}\selectfont$\pm$0.8}          & 75.1{\fontsize{6.5pt}{7pt}\selectfont$\pm$1.2}          & 68.0{\fontsize{6.5pt}{7pt}\selectfont$\pm$0.7}          \\
                        &{Ours}           & \textbf{93.1{\fontsize{6.5pt}{7pt}\selectfont$\pm$1.6}} & \textbf{91.9{\fontsize{6.5pt}{7pt}\selectfont$\pm$1.3}} & \textbf{96.1{\fontsize{6.5pt}{7pt}\selectfont$\pm$0.4}} & \textbf{82.1{\fontsize{6.5pt}{7pt}\selectfont$\pm$1.8}} & 81.5{\fontsize{6.5pt}{7pt}\selectfont$\pm$5.6}          & {\ul 73.9{\fontsize{6.5pt}{7pt}\selectfont$\pm$1.0}}    & {\ul 90.4{\fontsize{6.5pt}{7pt}\selectfont$\pm$2.1}}    & {\ul 84.3{\fontsize{6.5pt}{7pt}\selectfont$\pm$1.4}}    & \textbf{94.2{\fontsize{6.5pt}{7pt}\selectfont$\pm$1.0}} & {\ul 90.2{\fontsize{6.5pt}{7pt}\selectfont$\pm$0.6}}    & \textbf{87.8{\fontsize{6.5pt}{7pt}\selectfont$\pm$0.5}} \\ \midrule
\multirow{4}{*}{\rotatebox{90}{3D+RGB}} &{PatchCore+FPFH} & 70.4{\fontsize{6.5pt}{7pt}\selectfont$\pm$1.5}          & 72.8{\fontsize{6.5pt}{7pt}\selectfont$\pm$0.6}          & 77.9{\fontsize{6.5pt}{7pt}\selectfont$\pm$0.3}          & 77.5{\fontsize{6.5pt}{7pt}\selectfont$\pm$1.0}          & 68.8{\fontsize{6.5pt}{7pt}\selectfont$\pm$1.5}          & 64.9{\fontsize{6.5pt}{7pt}\selectfont$\pm$1.0}          & 65.0{\fontsize{6.5pt}{7pt}\selectfont$\pm$1.7}          & 65.9{\fontsize{6.5pt}{7pt}\selectfont$\pm$1.3}          & 56.4{\fontsize{6.5pt}{7pt}\selectfont$\pm$0.8}          & 75.3{\fontsize{6.5pt}{7pt}\selectfont$\pm$1.3}          & 69.5{\fontsize{6.5pt}{7pt}\selectfont$\pm$0.6}          \\
                        &{Shape-Guided}   & {\ul 74.6{\fontsize{6.5pt}{7pt}\selectfont$\pm$0.6}}    & {\ul 80.9{\fontsize{6.5pt}{7pt}\selectfont$\pm$0.5}}    & {\ul 93.6{\fontsize{6.5pt}{7pt}\selectfont$\pm$0.3}}    & {\ul 79.3{\fontsize{6.5pt}{7pt}\selectfont$\pm$0.9}}    & \textbf{89.3{\fontsize{6.5pt}{7pt}\selectfont$\pm$0.9}} & {\ul 76.6{\fontsize{6.5pt}{7pt}\selectfont$\pm$0.2}}    & {\ul 82.4{\fontsize{6.5pt}{7pt}\selectfont$\pm$0.2}}    & \textbf{94.0{\fontsize{6.5pt}{7pt}\selectfont$\pm$0.3}} & {\ul 86.6{\fontsize{6.5pt}{7pt}\selectfont$\pm$0.1}}    & \textbf{93.7{\fontsize{6.5pt}{7pt}\selectfont$\pm$0.8}} & {\ul 85.1{\fontsize{6.5pt}{7pt}\selectfont$\pm$0.0}}    \\
                        &{M3DM}           & 69.0{\fontsize{6.5pt}{7pt}\selectfont$\pm$1.4}          & 72.5{\fontsize{6.5pt}{7pt}\selectfont$\pm$0.8}          & 77.8{\fontsize{6.5pt}{7pt}\selectfont$\pm$0.4}          & 72.8{\fontsize{6.5pt}{7pt}\selectfont$\pm$1.0}          & 68.0{\fontsize{6.5pt}{7pt}\selectfont$\pm$1.5}          & 61.3{\fontsize{6.5pt}{7pt}\selectfont$\pm$0.7}          & 65.2{\fontsize{6.5pt}{7pt}\selectfont$\pm$1.5}          & 65.3{\fontsize{6.5pt}{7pt}\selectfont$\pm$1.4}          & 57.2{\fontsize{6.5pt}{7pt}\selectfont$\pm$0.8}          & 75.3{\fontsize{6.5pt}{7pt}\selectfont$\pm$1.2}          & 68.4{\fontsize{6.5pt}{7pt}\selectfont$\pm$0.6}          \\
                        &{Ours}           & \textbf{95.9{\fontsize{6.5pt}{7pt}\selectfont$\pm$1.3}} & \textbf{92.0{\fontsize{6.5pt}{7pt}\selectfont$\pm$1.2}} & \textbf{96.7{\fontsize{6.5pt}{7pt}\selectfont$\pm$0.4}} & \textbf{90.4{\fontsize{6.5pt}{7pt}\selectfont$\pm$1.1}} & {\ul 84.6{\fontsize{6.5pt}{7pt}\selectfont$\pm$2.3}}    & \textbf{83.4{\fontsize{6.5pt}{7pt}\selectfont$\pm$1.7}} & \textbf{91.9{\fontsize{6.5pt}{7pt}\selectfont$\pm$2.7}} & {\ul 85.8{\fontsize{6.5pt}{7pt}\selectfont$\pm$1.7}}    & \textbf{94.5{\fontsize{6.5pt}{7pt}\selectfont$\pm$0.3}} & {\ul 91.4{\fontsize{6.5pt}{7pt}\selectfont$\pm$0.5}}    & \textbf{90.7{\fontsize{6.5pt}{7pt}\selectfont$\pm$0.2}}                        \\ \bottomrule
\end{tabular}
\end{center}
\end{table*}
\subsection{Regular Anomaly Detection on MVTec 3D-AD}
\label{sec:regular_anomaly}
In the regular anomaly detection setting, we compare our method with several 3D-based, RGB-based, and hybrid multi-modal 3D/RGB methods on MVTec-3D. 
\cref{tab:iaucroc_ori,tab:aupro_ori} show the anomaly detection results record with I-AUROC and the segmentation results record with AUPRO respectively. 
We report the P-AUROC in \cref{app:paurco_regular}. From \cref{tab:iaucroc_ori,tab:iaucroc_ori}, we can conclude that our M3DM-NR also maintains the regular anomaly detection ability.

\subsection{Noisy Anomaly Detection on MVTec 3D-AD}
\label{sec:noisy_anomaly}
\begin{table*}[t]
\begin{center}
\caption{\textbf{I-AUROC score for anomaly detection under \textit{Non-Overlap} setting of all categories in MVTec 3D-AD.} Our method clearly outperforms other methods in 3D, RGB, and 3D + RGB settings, indicating the superior anomaly detection ability of our method. We report the mean and standard deviation over 3 random seeds for each measurement. Optimal and sub-optimal results are in \textbf{bold} and {\ul underlined}, respectively.}
\label{tab:i-auroc}
\setlength{\tabcolsep}{4.7pt}
\begin{tabular}{cc|cccccccccc|c}
\toprule
\textbf{}               &{Method}         & Bagel                & \begin{tabular}[c]{@{}c@{}}Cable\\ Gland\end{tabular}          & Carrot               & Cookie               & Dowel                & Foam                 & Peach                & Potato               & Rope                 & Tire                 & Mean                \\ \midrule
\multirow{6}{*}{\rotatebox{90}{3D}}     &{SIFT}           & 68.8{\fontsize{6.5pt}{7pt}\selectfont$\pm$1.1}          & 65.0{\fontsize{6.5pt}{7pt}\selectfont$\pm$2.6}          & 86.1{\fontsize{6.5pt}{7pt}\selectfont$\pm$0.3}          & 72.9{\fontsize{6.5pt}{7pt}\selectfont$\pm$0.6}          & 79.7{\fontsize{6.5pt}{7pt}\selectfont$\pm$5.2}          & 69.1{\fontsize{6.5pt}{7pt}\selectfont$\pm$3.9}          & 61.3{\fontsize{6.5pt}{7pt}\selectfont$\pm$1.0}          & 69.7{\fontsize{6.5pt}{7pt}\selectfont$\pm$1.9}          & 74.6{\fontsize{6.5pt}{7pt}\selectfont$\pm$1.9}          & 59.3{\fontsize{6.5pt}{7pt}\selectfont$\pm$3.6}          & 70.7{\fontsize{6.5pt}{7pt}\selectfont$\pm$0.6}          \\
                        &{FPFH}           & 73.4{\fontsize{6.5pt}{7pt}\selectfont$\pm$3.8}          & 54.8{\fontsize{6.5pt}{7pt}\selectfont$\pm$4.3}          & 90.7{\fontsize{6.5pt}{7pt}\selectfont$\pm$1.5}          & 78.5{\fontsize{6.5pt}{7pt}\selectfont$\pm$0.9}          & \textbf{88.3{\fontsize{6.5pt}{7pt}\selectfont$\pm$3.3}} & 54.0{\fontsize{6.5pt}{7pt}\selectfont$\pm$0.4}          & 70.9{\fontsize{6.5pt}{7pt}\selectfont$\pm$3.8}          & 67.2{\fontsize{6.5pt}{7pt}\selectfont$\pm$2.3}          & 90.0{\fontsize{6.5pt}{7pt}\selectfont$\pm$2.6}          & 67.9{\fontsize{6.5pt}{7pt}\selectfont$\pm$5.6}          & 73.6{\fontsize{6.5pt}{7pt}\selectfont$\pm$0.4}          \\
                        &{AST}            & 82.8{\fontsize{6.5pt}{7pt}\selectfont$\pm$0.6}          & 51.9{\fontsize{6.5pt}{7pt}\selectfont$\pm$0.6}          & 91.3{\fontsize{6.5pt}{7pt}\selectfont$\pm$0.6}          & {\ul 92.3{\fontsize{6.5pt}{7pt}\selectfont$\pm$1.2}}    & 64.3{\fontsize{6.5pt}{7pt}\selectfont$\pm$1.2}          & \textbf{78.5{\fontsize{6.5pt}{7pt}\selectfont$\pm$0.2}} & \textbf{98.3{\fontsize{6.5pt}{7pt}\selectfont$\pm$2.9}} & \textbf{90.3{\fontsize{6.5pt}{7pt}\selectfont$\pm$0.3}} & \textbf{94.7{\fontsize{6.5pt}{7pt}\selectfont$\pm$1.7}} & 63.3{\fontsize{6.5pt}{7pt}\selectfont$\pm$1.2}          & 80.8{\fontsize{6.5pt}{7pt}\selectfont$\pm$0.9}          \\
                        &{Shape-Guided}   & {\ul 90.2{\fontsize{6.5pt}{7pt}\selectfont$\pm$0.9}}    & 67.5{\fontsize{6.5pt}{7pt}\selectfont$\pm$0.1}          & {\ul 91.4{\fontsize{6.5pt}{7pt}\selectfont$\pm$0.3}}    & 92.1{\fontsize{6.5pt}{7pt}\selectfont$\pm$1.2}          & 80.8{\fontsize{6.5pt}{7pt}\selectfont$\pm$10.1}         & 67.7{\fontsize{6.5pt}{7pt}\selectfont$\pm$4.1}          & {\ul 86.5{\fontsize{6.5pt}{7pt}\selectfont$\pm$7.4}}    & 87.1{\fontsize{6.5pt}{7pt}\selectfont$\pm$1.0}          & 89.6{\fontsize{6.5pt}{7pt}\selectfont$\pm$1.3}          & \textbf{83.3{\fontsize{6.5pt}{7pt}\selectfont$\pm$6.7}} & {\ul 83.6{\fontsize{6.5pt}{7pt}\selectfont$\pm$2.0}}    \\
                        &{M3DM}           & 87.1{\fontsize{6.5pt}{7pt}\selectfont$\pm$0.8}          & {\ul 68.2{\fontsize{6.5pt}{7pt}\selectfont$\pm$1.2}}    & 79.4{\fontsize{6.5pt}{7pt}\selectfont$\pm$3.1}          & 87.8{\fontsize{6.5pt}{7pt}\selectfont$\pm$1.3}          & 83.8{\fontsize{6.5pt}{7pt}\selectfont$\pm$2.8}          & {\ul 73.0{\fontsize{6.5pt}{7pt}\selectfont$\pm$2.5}}    & 76.6{\fontsize{6.5pt}{7pt}\selectfont$\pm$2.6}          & 82.6{\fontsize{6.5pt}{7pt}\selectfont$\pm$0.7}          & {\ul 92.9{\fontsize{6.5pt}{7pt}\selectfont$\pm$2.0}}    & 80.0{\fontsize{6.5pt}{7pt}\selectfont$\pm$1.6}          & 81.1{\fontsize{6.5pt}{7pt}\selectfont$\pm$0.8}          \\
                        &{Ours}           & \textbf{94.5{\fontsize{6.5pt}{7pt}\selectfont$\pm$0.6}} & \textbf{74.4{\fontsize{6.5pt}{7pt}\selectfont$\pm$2.4}} & \textbf{94.8{\fontsize{6.5pt}{7pt}\selectfont$\pm$0.9}} & \textbf{93.7{\fontsize{6.5pt}{7pt}\selectfont$\pm$0.8}} & {\ul 83.8{\fontsize{6.5pt}{7pt}\selectfont$\pm$1.1}}    & 72.8{\fontsize{6.5pt}{7pt}\selectfont$\pm$3.5}          & 84.0{\fontsize{6.5pt}{7pt}\selectfont$\pm$0.2}          & {\ul 87.3{\fontsize{6.5pt}{7pt}\selectfont$\pm$0.4}}    & 89.8{\fontsize{6.5pt}{7pt}\selectfont$\pm$1.3}          & {\ul 82.2{\fontsize{6.5pt}{7pt}\selectfont$\pm$1.2}}    & \textbf{85.7{\fontsize{6.5pt}{7pt}\selectfont$\pm$0.7}} \\ \midrule
\multirow{7}{*}{\rotatebox{90}{RGB}}    &{PaDim}          & \textbf{93.0{\fontsize{6.5pt}{7pt}\selectfont$\pm$1.0}} & 73.3{\fontsize{6.5pt}{7pt}\selectfont$\pm$3.3}          & 66.3{\fontsize{6.5pt}{7pt}\selectfont$\pm$0.7}          & 52.4{\fontsize{6.5pt}{7pt}\selectfont$\pm$2.0}          & 88.3{\fontsize{6.5pt}{7pt}\selectfont$\pm$1.0}          & 72.2{\fontsize{6.5pt}{7pt}\selectfont$\pm$3.2}          & {\ul 84.3{\fontsize{6.5pt}{7pt}\selectfont$\pm$1.3}}    & 50.7{\fontsize{6.5pt}{7pt}\selectfont$\pm$2.2}          & 91.9{\fontsize{6.5pt}{7pt}\selectfont$\pm$2.7}          & 68.6{\fontsize{6.5pt}{7pt}\selectfont$\pm$2.2}          & 74.1{\fontsize{6.5pt}{7pt}\selectfont$\pm$0.6}          \\
                        &{PatchCore}      & 89.2{\fontsize{6.5pt}{7pt}\selectfont$\pm$0.9}          & 95.2{\fontsize{6.5pt}{7pt}\selectfont$\pm$1.4}          & 90.8{\fontsize{6.5pt}{7pt}\selectfont$\pm$1.9}          & 65.9{\fontsize{6.5pt}{7pt}\selectfont$\pm$2.8}          & 97.5{\fontsize{6.5pt}{7pt}\selectfont$\pm$1.0}          & 77.4{\fontsize{6.5pt}{7pt}\selectfont$\pm$4.7}          & 70.6{\fontsize{6.5pt}{7pt}\selectfont$\pm$1.7}          & 54.6{\fontsize{6.5pt}{7pt}\selectfont$\pm$0.6}          & 93.5{\fontsize{6.5pt}{7pt}\selectfont$\pm$2.2}          & 75.4{\fontsize{6.5pt}{7pt}\selectfont$\pm$1.7}          & 81.0{\fontsize{6.5pt}{7pt}\selectfont$\pm$0.7}          \\
                        &{AST}            & 79.5{\fontsize{6.5pt}{7pt}\selectfont$\pm$0.1}          & 83.1{\fontsize{6.5pt}{7pt}\selectfont$\pm$0.1}          & 63.2{\fontsize{6.5pt}{7pt}\selectfont$\pm$0.8}          & 60.2{\fontsize{6.5pt}{7pt}\selectfont$\pm$0.1}          & 80.7{\fontsize{6.5pt}{7pt}\selectfont$\pm$0.6}          & 77.5{\fontsize{6.5pt}{7pt}\selectfont$\pm$1.8}          & 81.1{\fontsize{6.5pt}{7pt}\selectfont$\pm$1.0}          & 63.4{\fontsize{6.5pt}{7pt}\selectfont$\pm$0.1}          & 74.3{\fontsize{6.5pt}{7pt}\selectfont$\pm$0.8}          & 59.2{\fontsize{6.5pt}{7pt}\selectfont$\pm$0.0}          & 72.2{\fontsize{6.5pt}{7pt}\selectfont$\pm$0.1}          \\
                        &{Shape-Guided}   & 79.3{\fontsize{6.5pt}{7pt}\selectfont$\pm$1.0}          & 89.6{\fontsize{6.5pt}{7pt}\selectfont$\pm$2.4}          & 77.4{\fontsize{6.5pt}{7pt}\selectfont$\pm$0.3}          & 58.6{\fontsize{6.5pt}{7pt}\selectfont$\pm$2.0}          & 94.3{\fontsize{6.5pt}{7pt}\selectfont$\pm$0.2}          & 71.4{\fontsize{6.5pt}{7pt}\selectfont$\pm$3.6}          & 67.7{\fontsize{6.5pt}{7pt}\selectfont$\pm$0.7}          & {\ul 62.1{\fontsize{6.5pt}{7pt}\selectfont$\pm$0.0}}    & 72.0{\fontsize{6.5pt}{7pt}\selectfont$\pm$1.6}          & 66.5{\fontsize{6.5pt}{7pt}\selectfont$\pm$0.3}          & 73.9{\fontsize{6.5pt}{7pt}\selectfont$\pm$0.8}          \\
                        &{SoftPatch}      & 90.6{\fontsize{6.5pt}{7pt}\selectfont$\pm$0.2}          & \textbf{91.8{\fontsize{6.5pt}{7pt}\selectfont$\pm$1.7}} & \textbf{87.6{\fontsize{6.5pt}{7pt}\selectfont$\pm$0.4}} & {\ul 67.8{\fontsize{6.5pt}{7pt}\selectfont$\pm$0.8}}    & \textbf{98.0{\fontsize{6.5pt}{7pt}\selectfont$\pm$0.6}} & {\ul 78.0{\fontsize{6.5pt}{7pt}\selectfont$\pm$4.8}}    & 70.6{\fontsize{6.5pt}{7pt}\selectfont$\pm$0.7}          & 55.3{\fontsize{6.5pt}{7pt}\selectfont$\pm$1.5}          & \textbf{93.4{\fontsize{6.5pt}{7pt}\selectfont$\pm$2.7}} & 75.6{\fontsize{6.5pt}{7pt}\selectfont$\pm$1.2}          & 80.9{\fontsize{6.5pt}{7pt}\selectfont$\pm$0.4}          \\
                        &{M3DM}           & 87.7{\fontsize{6.5pt}{7pt}\selectfont$\pm$2.3}          & 83.0{\fontsize{6.5pt}{7pt}\selectfont$\pm$2.7}          & 83.1{\fontsize{6.5pt}{7pt}\selectfont$\pm$1.1}          & 66.4{\fontsize{6.5pt}{7pt}\selectfont$\pm$1.7}          & {\ul 96.7{\fontsize{6.5pt}{7pt}\selectfont$\pm$1.4}}    & 77.7{\fontsize{6.5pt}{7pt}\selectfont$\pm$1.7}          & 82.7{\fontsize{6.5pt}{7pt}\selectfont$\pm$3.1}          & \textbf{62.5{\fontsize{6.5pt}{7pt}\selectfont$\pm$3.4}} & 92.9{\fontsize{6.5pt}{7pt}\selectfont$\pm$1.8}          & \textbf{76.7{\fontsize{6.5pt}{7pt}\selectfont$\pm$1.2}} & {\ul 80.9{\fontsize{6.5pt}{7pt}\selectfont$\pm$0.8}}    \\
                        &{Ours}           & {\ul 90.8{\fontsize{6.5pt}{7pt}\selectfont$\pm$1.3}}    & {\ul 90.2{\fontsize{6.5pt}{7pt}\selectfont$\pm$4.0}}    & {\ul 86.9{\fontsize{6.5pt}{7pt}\selectfont$\pm$1.8}}    & \textbf{68.0{\fontsize{6.5pt}{7pt}\selectfont$\pm$3.6}} & 91.0{\fontsize{6.5pt}{7pt}\selectfont$\pm$3.6}          & \textbf{83.2{\fontsize{6.5pt}{7pt}\selectfont$\pm$1.8}} & \textbf{88.7{\fontsize{6.5pt}{7pt}\selectfont$\pm$2.1}} & 57.7{\fontsize{6.5pt}{7pt}\selectfont$\pm$6.7}          & {\ul 93.3{\fontsize{6.5pt}{7pt}\selectfont$\pm$1.1}}    & {\ul 75.9{\fontsize{6.5pt}{7pt}\selectfont$\pm$1.6}}    & \textbf{82.6{\fontsize{6.5pt}{7pt}\selectfont$\pm$0.5}} \\ \midrule
\multirow{5}{*}{\rotatebox{90}{3D+RGB}} &{PatchCore+FPFH} & 81.1{\fontsize{6.5pt}{7pt}\selectfont$\pm$4.0}          & 77.8{\fontsize{6.5pt}{7pt}\selectfont$\pm$1.4}          & 91.7{\fontsize{6.5pt}{7pt}\selectfont$\pm$0.5}          & 84.5{\fontsize{6.5pt}{7pt}\selectfont$\pm$1.6}          & 91.8{\fontsize{6.5pt}{7pt}\selectfont$\pm$1.3}          & 64.8{\fontsize{6.5pt}{7pt}\selectfont$\pm$2.6}          & 79.5{\fontsize{6.5pt}{7pt}\selectfont$\pm$3.1}          & 77.3{\fontsize{6.5pt}{7pt}\selectfont$\pm$1.9}          & 90.9{\fontsize{6.5pt}{7pt}\selectfont$\pm$1.6}          & \textbf{89.8{\fontsize{6.5pt}{7pt}\selectfont$\pm$1.1}} & 82.9{\fontsize{6.5pt}{7pt}\selectfont$\pm$0.8}          \\
                        &{AST}            & 85.4{\fontsize{6.5pt}{7pt}\selectfont$\pm$0.6}          & {\ul 88.9{\fontsize{6.5pt}{7pt}\selectfont$\pm$0.6}}    & 91.3{\fontsize{6.5pt}{7pt}\selectfont$\pm$0.6}          & \textbf{95.6{\fontsize{6.5pt}{7pt}\selectfont$\pm$0.6}} & 89.2{\fontsize{6.5pt}{7pt}\selectfont$\pm$1.0}          & 85.9{\fontsize{6.5pt}{7pt}\selectfont$\pm$0.6}          & {\ul 92.8{\fontsize{6.5pt}{7pt}\selectfont$\pm$0.6}}    & \textbf{91.6{\fontsize{6.5pt}{7pt}\selectfont$\pm$0.6}} & 79.6{\fontsize{6.5pt}{7pt}\selectfont$\pm$0.6}          & 70.0{\fontsize{6.5pt}{7pt}\selectfont$\pm$0.6}          & 87.0{\fontsize{6.5pt}{7pt}\selectfont$\pm$0.3}          \\
                        &{Shape-Guided}   & 91.0{\fontsize{6.5pt}{7pt}\selectfont$\pm$0.5}          & 86.3{\fontsize{6.5pt}{7pt}\selectfont$\pm$2.0}          & {\ul 94.2{\fontsize{6.5pt}{7pt}\selectfont$\pm$0.5}}    & 86.4{\fontsize{6.5pt}{7pt}\selectfont$\pm$1.0}          & {\ul 94.2{\fontsize{6.5pt}{7pt}\selectfont$\pm$0.1}}    & 77.1{\fontsize{6.5pt}{7pt}\selectfont$\pm$0.5}          & 88.6{\fontsize{6.5pt}{7pt}\selectfont$\pm$0.1}          & {\ul 85.8{\fontsize{6.5pt}{7pt}\selectfont$\pm$1.0}}    & 88.3{\fontsize{6.5pt}{7pt}\selectfont$\pm$0.1}          & {\ul 85.1{\fontsize{6.5pt}{7pt}\selectfont$\pm$0.2}}    & {\ul 87.7{\fontsize{6.5pt}{7pt}\selectfont$\pm$0.3}}    \\
                        &{M3DM}           & {\ul 96.6{\fontsize{6.5pt}{7pt}\selectfont$\pm$2.2}}    & 85.7{\fontsize{6.5pt}{7pt}\selectfont$\pm$1.9}          & 88.4{\fontsize{6.5pt}{7pt}\selectfont$\pm$2.5}          & 86.4{\fontsize{6.5pt}{7pt}\selectfont$\pm$3.1}          & \textbf{96.1{\fontsize{6.5pt}{7pt}\selectfont$\pm$1.3}} & {\ul 86.3{\fontsize{6.5pt}{7pt}\selectfont$\pm$5.4}}    & 85.1{\fontsize{6.5pt}{7pt}\selectfont$\pm$0.6}          & 76.5{\fontsize{6.5pt}{7pt}\selectfont$\pm$2.3}          & {\ul 94.8{\fontsize{6.5pt}{7pt}\selectfont$\pm$1.3}}    & 79.3{\fontsize{6.5pt}{7pt}\selectfont$\pm$2.4}          & 87.5{\fontsize{6.5pt}{7pt}\selectfont$\pm$0.5}          \\
                        &{Ours}           & \textbf{98.1{\fontsize{6.5pt}{7pt}\selectfont$\pm$0.8}} & \textbf{91.0{\fontsize{6.5pt}{7pt}\selectfont$\pm$2.6}} & \textbf{96.8{\fontsize{6.5pt}{7pt}\selectfont$\pm$0.8}} & {\ul 94.2{\fontsize{6.5pt}{7pt}\selectfont$\pm$2.0}}    & 93.7{\fontsize{6.5pt}{7pt}\selectfont$\pm$0.8}          & \textbf{90.6{\fontsize{6.5pt}{7pt}\selectfont$\pm$2.0}} & \textbf{92.9{\fontsize{6.5pt}{7pt}\selectfont$\pm$1.6}} & 81.9{\fontsize{6.5pt}{7pt}\selectfont$\pm$2.0}          & \textbf{95.3{\fontsize{6.5pt}{7pt}\selectfont$\pm$1.4}} & 84.7{\fontsize{6.5pt}{7pt}\selectfont$\pm$2.4}          & \textbf{91.9{\fontsize{6.5pt}{7pt}\selectfont$\pm$1.0}}                        \\ \bottomrule
\end{tabular}
\end{center}
\end{table*}
\begin{table*}[t]
\begin{center}
\caption{\textbf{AUPRO score for anomaly segmentation under \textit{Non-Overlap} setting of all categories in MVTec 3D-AD.} Our method clearly outperforms other methods in 3D and 3D + RGB settings, indicating the superior anomaly segmentation ability of our method. We report the mean and standard deviation over 3 random seeds for each measurement. Optimal and sub-optimal results are in \textbf{bold} and {\ul underlined}, respectively.}
\label{tab:aupro}
\setlength{\tabcolsep}{4.7pt}
\begin{tabular}{cc|cccccccccc|c}
\toprule
\textbf{}               &{Method}         & Bagel                & \begin{tabular}[c]{@{}c@{}}Cable\\ Gland\end{tabular}          & Carrot               & Cookie               & Dowel                & Foam                 & Peach                & Potato               & Rope                 & Tire                 & Mean                \\ \midrule
\multirow{5}{*}{\rotatebox{90}{3D}}     &{SIFT}           & 86.4{\fontsize{6.5pt}{7pt}\selectfont$\pm$0.0}          & 70.2{\fontsize{6.5pt}{7pt}\selectfont$\pm$0.0}          & 90.3{\fontsize{6.5pt}{7pt}\selectfont$\pm$0.0}          & 86.1{\fontsize{6.5pt}{7pt}\selectfont$\pm$0.0}          & \textbf{90.6{\fontsize{6.5pt}{7pt}\selectfont$\pm$0.0}} & 60.3{\fontsize{6.5pt}{7pt}\selectfont$\pm$0.0}          & 85.0{\fontsize{6.5pt}{7pt}\selectfont$\pm$0.0}          & 95.3{\fontsize{6.5pt}{7pt}\selectfont$\pm$0.0}          & 93.8{\fontsize{6.5pt}{7pt}\selectfont$\pm$0.0}          & 86.3{\fontsize{6.5pt}{7pt}\selectfont$\pm$0.0}          & 84.4{\fontsize{6.5pt}{7pt}\selectfont$\pm$0.0}          \\
                        &{FPFH}           & 92.6{\fontsize{6.5pt}{7pt}\selectfont$\pm$0.0}          & 78.3{\fontsize{6.5pt}{7pt}\selectfont$\pm$0.0}          & 92.1{\fontsize{6.5pt}{7pt}\selectfont$\pm$0.0}          & 85.5{\fontsize{6.5pt}{7pt}\selectfont$\pm$0.0}          & {\ul 88.2{\fontsize{6.5pt}{7pt}\selectfont$\pm$0.0}}    & 68.3{\fontsize{6.5pt}{7pt}\selectfont$\pm$0.0}          & 90.5{\fontsize{6.5pt}{7pt}\selectfont$\pm$0.0}          & 94.3{\fontsize{6.5pt}{7pt}\selectfont$\pm$0.0}          & 92.1{\fontsize{6.5pt}{7pt}\selectfont$\pm$0.0}          & 90.3{\fontsize{6.5pt}{7pt}\selectfont$\pm$0.0}          & 87.2{\fontsize{6.5pt}{7pt}\selectfont$\pm$0.0}          \\
                        &{Shape-Guided}   & {\ul 95.6{\fontsize{6.5pt}{7pt}\selectfont$\pm$0.0}}    & 80.3{\fontsize{6.5pt}{7pt}\selectfont$\pm$0.0}          & \textbf{98.1{\fontsize{6.5pt}{7pt}\selectfont$\pm$0.0}} & \textbf{89.5{\fontsize{6.5pt}{7pt}\selectfont$\pm$0.0}} & {\ul 88.2{\fontsize{6.5pt}{7pt}\selectfont$\pm$0.0}}    & 70.3{\fontsize{6.5pt}{7pt}\selectfont$\pm$0.0}          & 95.2{\fontsize{6.5pt}{7pt}\selectfont$\pm$0.6}          & 96.3{\fontsize{6.5pt}{7pt}\selectfont$\pm$0.0}          & 93.1{\fontsize{6.5pt}{7pt}\selectfont$\pm$0.0}          & \textbf{93.7{\fontsize{6.5pt}{7pt}\selectfont$\pm$0.0}} & 90.0{\fontsize{6.5pt}{7pt}\selectfont$\pm$0.1}          \\
                        &{M3DM}           & 93.7{\fontsize{6.5pt}{7pt}\selectfont$\pm$0.5}          & {\ul 81.1{\fontsize{6.5pt}{7pt}\selectfont$\pm$0.3}}    & {\ul 97.6{\fontsize{6.5pt}{7pt}\selectfont$\pm$0.2}}    & 86.3{\fontsize{6.5pt}{7pt}\selectfont$\pm$0.4}          & 87.9{\fontsize{6.5pt}{7pt}\selectfont$\pm$1.3}          & \textbf{75.3{\fontsize{6.5pt}{7pt}\selectfont$\pm$4.6}} & {\ul 95.4{\fontsize{6.5pt}{7pt}\selectfont$\pm$0.2}}    & \textbf{96.9{\fontsize{6.5pt}{7pt}\selectfont$\pm$0.4}} & \textbf{94.6{\fontsize{6.5pt}{7pt}\selectfont$\pm$0.4}} & 92.7{\fontsize{6.5pt}{7pt}\selectfont$\pm$0.3}          & {\ul 90.1{\fontsize{6.5pt}{7pt}\selectfont$\pm$0.6}}    \\
                        &{Ours}           & \textbf{95.8{\fontsize{6.5pt}{7pt}\selectfont$\pm$0.3}} & \textbf{81.2{\fontsize{6.5pt}{7pt}\selectfont$\pm$0.4}} & {\ul 97.6{\fontsize{6.5pt}{7pt}\selectfont$\pm$0.1}}    & {\ul 86.6{\fontsize{6.5pt}{7pt}\selectfont$\pm$0.7}}    & 88.0{\fontsize{6.5pt}{7pt}\selectfont$\pm$1.1}          & {\ul 73.0{\fontsize{6.5pt}{7pt}\selectfont$\pm$4.0}}    & \textbf{95.5{\fontsize{6.5pt}{7pt}\selectfont$\pm$0.4}} & {\ul 96.5{\fontsize{6.5pt}{7pt}\selectfont$\pm$0.1}}    & {\ul 94.2{\fontsize{6.5pt}{7pt}\selectfont$\pm$0.6}}    & {\ul 93.5{\fontsize{6.5pt}{7pt}\selectfont$\pm$0.8}}    & \textbf{90.2{\fontsize{6.5pt}{7pt}\selectfont$\pm$0.5}} \\ \midrule
\multirow{6}{*}{\rotatebox{90}{RGB}}    &{PaDim}          & 93.0{\fontsize{6.5pt}{7pt}\selectfont$\pm$2.4}          & 87.5{\fontsize{6.5pt}{7pt}\selectfont$\pm$2.6}          & 93.7{\fontsize{6.5pt}{7pt}\selectfont$\pm$0.4}          & 86.8{\fontsize{6.5pt}{7pt}\selectfont$\pm$0.9}          & 92.7{\fontsize{6.5pt}{7pt}\selectfont$\pm$1.3}          & \textbf{93.3{\fontsize{6.5pt}{7pt}\selectfont$\pm$7.0}} & 94.9{\fontsize{6.5pt}{7pt}\selectfont$\pm$0.5}          & \textbf{95.0{\fontsize{6.5pt}{7pt}\selectfont$\pm$1.0}} & 92.4{\fontsize{6.5pt}{7pt}\selectfont$\pm$0.6}          & 94.9{\fontsize{6.5pt}{7pt}\selectfont$\pm$0.6}          & 92.4{\fontsize{6.5pt}{7pt}\selectfont$\pm$0.5}          \\
                        &{PatchCore}      & 90.9{\fontsize{6.5pt}{7pt}\selectfont$\pm$0.6}          & \textbf{97.0{\fontsize{6.5pt}{7pt}\selectfont$\pm$0.1}} & 96.2{\fontsize{6.5pt}{7pt}\selectfont$\pm$0.5}          & {\ul 88.4{\fontsize{6.5pt}{7pt}\selectfont$\pm$0.5}}    & \textbf{95.7{\fontsize{6.5pt}{7pt}\selectfont$\pm$0.4}} & 79.1{\fontsize{6.5pt}{7pt}\selectfont$\pm$2.5}          & 89.2{\fontsize{6.5pt}{7pt}\selectfont$\pm$0.5}          & 93.4{\fontsize{6.5pt}{7pt}\selectfont$\pm$0.9}          & 96.5{\fontsize{6.5pt}{7pt}\selectfont$\pm$0.7}          & 95.1{\fontsize{6.5pt}{7pt}\selectfont$\pm$0.2}          & 92.2{\fontsize{6.5pt}{7pt}\selectfont$\pm$0.2}          \\
                        &{Shape-Guided}   & 90.2{\fontsize{6.5pt}{7pt}\selectfont$\pm$1.9}          & 94.5{\fontsize{6.5pt}{7pt}\selectfont$\pm$2.2}          & 94.9{\fontsize{6.5pt}{7pt}\selectfont$\pm$1.3}          & 86.5{\fontsize{6.5pt}{7pt}\selectfont$\pm$1.2}          & 93.6{\fontsize{6.5pt}{7pt}\selectfont$\pm$0.5}          & 74.8{\fontsize{6.5pt}{7pt}\selectfont$\pm$6.5}          & 90.7{\fontsize{6.5pt}{7pt}\selectfont$\pm$4.0}          & 92.4{\fontsize{6.5pt}{7pt}\selectfont$\pm$1.7}          & 91.8{\fontsize{6.5pt}{7pt}\selectfont$\pm$4.3}          & 93.3{\fontsize{6.5pt}{7pt}\selectfont$\pm$2.2}          & 90.3{\fontsize{6.5pt}{7pt}\selectfont$\pm$2.2}          \\
                        &{SoftPatch}      & 93.2{\fontsize{6.5pt}{7pt}\selectfont$\pm$0.3}          & 96.1{\fontsize{6.5pt}{7pt}\selectfont$\pm$0.1}          & 96.4{\fontsize{6.5pt}{7pt}\selectfont$\pm$0.1}          & \textbf{89.7{\fontsize{6.5pt}{7pt}\selectfont$\pm$0.7}} & {\ul 95.3{\fontsize{6.5pt}{7pt}\selectfont$\pm$0.5}}    & 78.4{\fontsize{6.5pt}{7pt}\selectfont$\pm$1.7}          & 90.0{\fontsize{6.5pt}{7pt}\selectfont$\pm$0.3}          & 93.5{\fontsize{6.5pt}{7pt}\selectfont$\pm$0.7}          & 96.2{\fontsize{6.5pt}{7pt}\selectfont$\pm$0.7}          & 94.7{\fontsize{6.5pt}{7pt}\selectfont$\pm$0.5}          & 92.3{\fontsize{6.5pt}{7pt}\selectfont$\pm$0.2}          \\
                        &{M3DM}           & {\ul 93.5{\fontsize{6.5pt}{7pt}\selectfont$\pm$0.3}}    & {\ul 96.8{\fontsize{6.5pt}{7pt}\selectfont$\pm$0.3}}    & \textbf{96.9{\fontsize{6.5pt}{7pt}\selectfont$\pm$0.5}} & 86.0{\fontsize{6.5pt}{7pt}\selectfont$\pm$0.6}          & 93.8{\fontsize{6.5pt}{7pt}\selectfont$\pm$0.8}          & 79.2{\fontsize{6.5pt}{7pt}\selectfont$\pm$1.6}          & \textbf{96.2{\fontsize{6.5pt}{7pt}\selectfont$\pm$0.4}} & 94.8{\fontsize{6.5pt}{7pt}\selectfont$\pm$0.6}          & \textbf{96.8{\fontsize{6.5pt}{7pt}\selectfont$\pm$0.4}} & \textbf{96.9{\fontsize{6.5pt}{7pt}\selectfont$\pm$0.1}} & \textbf{93.1{\fontsize{6.5pt}{7pt}\selectfont$\pm$0.1}} \\
                        &{Ours}           & \textbf{93.7{\fontsize{6.5pt}{7pt}\selectfont$\pm$0.9}} & 96.0{\fontsize{6.5pt}{7pt}\selectfont$\pm$0.6}          & {\ul 96.8{\fontsize{6.5pt}{7pt}\selectfont$\pm$0.3}}    & 84.0{\fontsize{6.5pt}{7pt}\selectfont$\pm$1.5}          & 92.4{\fontsize{6.5pt}{7pt}\selectfont$\pm$1.0}          & {\ul 79.5{\fontsize{6.5pt}{7pt}\selectfont$\pm$2.4}}    & {\ul 95.6{\fontsize{6.5pt}{7pt}\selectfont$\pm$0.1}}    & {\ul 94.8{\fontsize{6.5pt}{7pt}\selectfont$\pm$0.6}}    & {\ul 96.8{\fontsize{6.5pt}{7pt}\selectfont$\pm$0.6}}    & {\ul 95.3{\fontsize{6.5pt}{7pt}\selectfont$\pm$0.3}}    & {\ul 92.5{\fontsize{6.5pt}{7pt}\selectfont$\pm$0.2}}    \\ \midrule
\multirow{4}{*}{\rotatebox{90}{3D+RGB}} &{PatchCore+FPFH} & {\ul 96.6{\fontsize{6.5pt}{7pt}\selectfont$\pm$0.4}}    & 96.1{\fontsize{6.5pt}{7pt}\selectfont$\pm$1.2}          & \textbf{97.7{\fontsize{6.5pt}{7pt}\selectfont$\pm$0.5}} & 92.6{\fontsize{6.5pt}{7pt}\selectfont$\pm$3.2}          & 92.5{\fontsize{6.5pt}{7pt}\selectfont$\pm$1.4}          & 89.1{\fontsize{6.5pt}{7pt}\selectfont$\pm$0.5}          & {\ul 96.5{\fontsize{6.5pt}{7pt}\selectfont$\pm$0.2}}    & {\ul 96.7{\fontsize{6.5pt}{7pt}\selectfont$\pm$0.2}}    & 95.3{\fontsize{6.5pt}{7pt}\selectfont$\pm$1.1}          & \textbf{97.2{\fontsize{6.5pt}{7pt}\selectfont$\pm$0.1}} & 95.0{\fontsize{6.5pt}{7pt}\selectfont$\pm$0.4}          \\
                        &{Shape-Guided}   & 93.5{\fontsize{6.5pt}{7pt}\selectfont$\pm$0.1}          & 94.0{\fontsize{6.5pt}{7pt}\selectfont$\pm$0.2}          & 97.5{\fontsize{6.5pt}{7pt}\selectfont$\pm$0.3}          & \textbf{93.0{\fontsize{6.5pt}{7pt}\selectfont$\pm$0.3}} & \textbf{95.5{\fontsize{6.5pt}{7pt}\selectfont$\pm$0.1}} & \textbf{93.1{\fontsize{6.5pt}{7pt}\selectfont$\pm$0.8}} & 95.3{\fontsize{6.5pt}{7pt}\selectfont$\pm$0.1}          & \textbf{97.9{\fontsize{6.5pt}{7pt}\selectfont$\pm$0.1}} & 95.6{\fontsize{6.5pt}{7pt}\selectfont$\pm$0.1}          & {\ul 97.2{\fontsize{6.5pt}{7pt}\selectfont$\pm$0.2}}    & {\ul 95.2{\fontsize{6.5pt}{7pt}\selectfont$\pm$0.1}}    \\
                        &{M3DM}           & 94.3{\fontsize{6.5pt}{7pt}\selectfont$\pm$0.8}          & \textbf{96.5{\fontsize{6.5pt}{7pt}\selectfont$\pm$0.3}} & 97.4{\fontsize{6.5pt}{7pt}\selectfont$\pm$0.5}          & 89.2{\fontsize{6.5pt}{7pt}\selectfont$\pm$0.2}          & 92.7{\fontsize{6.5pt}{7pt}\selectfont$\pm$0.9}          & 82.8{\fontsize{6.5pt}{7pt}\selectfont$\pm$1.0}          & 96.4{\fontsize{6.5pt}{7pt}\selectfont$\pm$0.3}          & 95.4{\fontsize{6.5pt}{7pt}\selectfont$\pm$0.6}          & \textbf{97.2{\fontsize{6.5pt}{7pt}\selectfont$\pm$0.4}} & 96.7{\fontsize{6.5pt}{7pt}\selectfont$\pm$0.3}          & 93.9{\fontsize{6.5pt}{7pt}\selectfont$\pm$0.2}          \\
                        &{Ours}           & \textbf{96.9{\fontsize{6.5pt}{7pt}\selectfont$\pm$0.3}} & {\ul 96.3{\fontsize{6.5pt}{7pt}\selectfont$\pm$0.2}}    & {\ul 97.6{\fontsize{6.5pt}{7pt}\selectfont$\pm$0.0}}    & {\ul 92.7{\fontsize{6.5pt}{7pt}\selectfont$\pm$0.5}}    & {\ul 93.9{\fontsize{6.5pt}{7pt}\selectfont$\pm$0.4}}    & {\ul 91.8{\fontsize{6.5pt}{7pt}\selectfont$\pm$1.3}}    & \textbf{97.0{\fontsize{6.5pt}{7pt}\selectfont$\pm$0.5}} & 96.4{\fontsize{6.5pt}{7pt}\selectfont$\pm$0.1}          & {\ul 97.0{\fontsize{6.5pt}{7pt}\selectfont$\pm$0.2}}    & 96.5{\fontsize{6.5pt}{7pt}\selectfont$\pm$0.1}          & \textbf{95.6{\fontsize{6.5pt}{7pt}\selectfont$\pm$0.1}}                        \\ \bottomrule
\end{tabular}
\end{center}
\end{table*}
In the noisy anomaly detection setting, we compare our method with several 3D-based, RGB-based, and hybrid multi-modal 3D/RGB methods on MVTec-3D. 
\cref{tab:i-auroc overlap,tab:i-auroc} show the anomaly detection results record with I-AUROC under \textit{Overlap} and \textit{Non-Overlap} settings respectively. 
\cref{tab:aupro overlap,tab:aupro} show the segmentation results record with AUPRO under \textit{Overlap} and \textit{Non-Overlap} settings respectively. We report the P-AUROC in \cref{app.pauroc_noisy}.

\noindent \textbf{Overlap and Non-Overlap Analysis.}
Compared to the \textit{Non-Overlap} setting, our method significantly outperformed all baseline methods in the \textit{Overlap} setting, especially in anomaly detection (I-AUROC). Specifically, our approach exceeded the second-best by 13.9\%, 3.7\%, and 20.3\% in I-AUROC for the 3D, RGB, and 3D+RGB settings, respectively. This indicates the effectiveness of sample-level denoising in Stage I \& II of our method, as most baseline methods struggled with anomalies existing in both the training and test datasets. This includes approaches like SoftPatch~\cite{jiang2022softpatch}, which only perform denoising at the patch-level, whereas our method remained largely unaffected. This demonstrates the enhanced robustness of our proposed Stage I \& II, especially in situations where defects with similar appearances existing in both the training and test datasets, \textit{i.e.}, a common scenario in real-world industrial settings.

\noindent \textbf{3D-Based.} On pure 3D anomaly detection, we get the highest I-AUROC and outperform M3DM~\cite{wang2023multimodal} 13.9\% in \textit{Overlap} and Shape-Guided~\cite{chu2023shape} 2.1\% in \textit{Non-Overlap}. For segmentation, we get the best result with AUPRO and outperform Shape-Guided 0.3\% in \textit{Overlap} and M3DM 0.1\% in \textit{Non-Overlap}. This shows our method has much better detection and segementation performance than the previous method, and with our PFA, the Point Transformer is the better 3D feature extractor for this task.

\noindent \textbf{RGB-Based.}  Our I-AUROC in RGB domain is 3.7\% higher than SoftPatch in \textit{Overlap} and 1.7\% higher than Softpatch and M3DM in \textit{Non-Overlap}. For segmentation, we get the highest AUPRO score, 0.6\% higher than PaDim in \textit{Overlap} and second best score in \textit{Non-Overlap}.

\noindent \textbf{Hybrid 3D/RGB.} On multi-modal 3D/RGB anomaly detection, we get the highest I-AUROC and outperform M3DM 20.3\% in \textit{Overlap} and Shape-Guided 4.2\% in \textit{Non-Overlap}. For segmentation, we get the best result with AUPRO and outperform Shape-Guided 0.6\% in \textit{Overlap} and Shape-guided 0.4\% in \textit{Non-Overlap}. These results are contributed by our fusion strategy and the high-performance 3D anomaly detection results.
\begin{figure*}[t]
    \centering
    \includegraphics[width=0.9\linewidth]{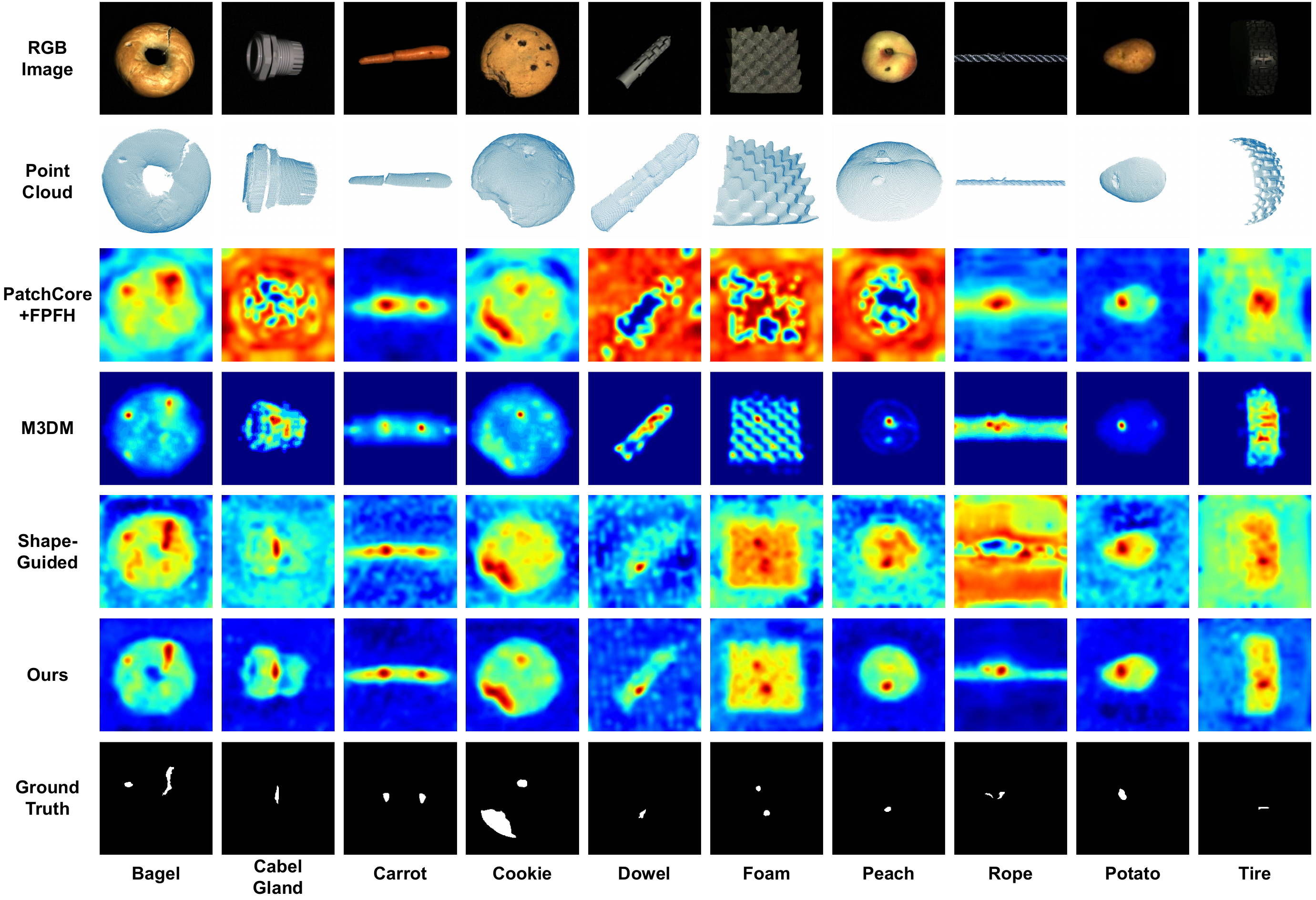}
    \caption{Heatmap of our anomaly segmentation results (multi-modal inputs) under \textit{Overlap} setting. Compared with existing methods, our method remains unaffected by noise and outputs a more accurate segmentation region.}
    \label{fig:overlap}
\end{figure*}

\begin{table*}[t]
\begin{center}
\caption{\textbf{Main ablation study of M3DM-NR.} Stage I\&II indicates removing stage I\&II, $R$ indicates removing Intra-modality Reference Selection, $\mathcal{H}_P$ indicates removing Aligned Multi-scale Point Cloud Extraction and $W$ indicates removing Noise-focused Aggregation. Noise-level refers to the percentage of noise data in the entire training set after denoising in stage I\&II.  We report the mean and standard deviation over 3 random seeds for each measurement. Optimal and sub-optimal results are in \textbf{bold} and {\ul underlined}, respectively.}
\label{tab:ablation}
\begin{tabular}{cccc|ccc|ccc|c}
\toprule
\multirow{2}{*}{Stage I\&II}        & \multirow{2}{*}{$R$}         & \multirow{2}{*}{$\mathcal{H}_P$}   & \multirow{2}{*}{$W$}         & \multicolumn{3}{c|}{{Overlap}}                 & \multicolumn{3}{c|}{{Non-Overlap}}             & \multirow{2}{*}{{Noise-level $\downarrow$}} \\ \cmidrule{5-10}
                             &                              &                              &                              & {I-AUROC $\uparrow$} & {P-AUROC $\uparrow$} & {AUPRO $\uparrow$} & {I-AUROC $\uparrow$} & {P-AUROC $\uparrow$} & {AUPRO $\uparrow$} &                                       \\ \midrule
{\color{red}\ding{55}}       & {\color{red}\ding{55}}       & {\color{red}\ding{55}}       & {\color{red}\ding{55}}       & 66.4{\fontsize{6.3pt}{7pt}\selectfont$\pm$0.4}&     72.9{\fontsize{6.3pt}{7pt}\selectfont$\pm$0.9}&     66.5{\fontsize{6.3pt}{7pt}\selectfont$\pm$3.4}&   87.7{\fontsize{6.3pt}{7pt}\selectfont$\pm$0.5}&     98.7{\fontsize{6.3pt}{7pt}\selectfont$\pm$0.1}&     94.5{\fontsize{6.3pt}{7pt}\selectfont$\pm$0.2}&   9.09{\fontsize{6.3pt}{7pt}\selectfont$\pm$0.00}                        \\
{\color{softgreen}\ding{51}} & {\color{red}\ding{55}}       & {\color{red}\ding{55}}       & {\color{red}\ding{55}}       & 79.7{\fontsize{6.3pt}{7pt}\selectfont$\pm$1.1}&     89.2{\fontsize{6.3pt}{7pt}\selectfont$\pm$1.1}&    84.5{\fontsize{6.3pt}{7pt}\selectfont$\pm$0.6}&   88.6{\fontsize{6.3pt}{7pt}\selectfont$\pm$0.6}&     98.8{\fontsize{6.3pt}{7pt}\selectfont$\pm$0.1}&     94.9{\fontsize{6.3pt}{7pt}\selectfont$\pm$0.3}&   5.13{\fontsize{6.3pt}{7pt}\selectfont$\pm$0.13}                        \\
{\color{softgreen}\ding{51}} & {\color{softgreen}\ding{51}} & {\color{red}\ding{55}}       & {\color{red}\ding{55}}       & 82.6{\fontsize{6.3pt}{7pt}\selectfont$\pm$0.7}&     92.7{\fontsize{6.3pt}{7pt}\selectfont$\pm$0.5}&     87.8{\fontsize{6.3pt}{7pt}\selectfont$\pm$0.3}&   89.2{\fontsize{6.3pt}{7pt}\selectfont$\pm$0.8}&    {\ul 98.7{\fontsize{6.3pt}{7pt}\selectfont$\pm$0.0}}&     94.9{\fontsize{6.3pt}{7pt}\selectfont$\pm$0.1}&   3.87{\fontsize{6.3pt}{7pt}\selectfont$\pm$0.08}                        \\
{\color{softgreen}\ding{51}} & {\color{softgreen}\ding{51}} & {\color{softgreen}\ding{51}} & {\color{red}\ding{55}}       & {\ul 86.2{\fontsize{6.3pt}{7pt}\selectfont$\pm$0.5}}&     {\ul 94.3{\fontsize{6.3pt}{7pt}\selectfont$\pm$0.4}}&     {\ul 90.3{\fontsize{6.3pt}{7pt}\selectfont$\pm$0.5}}&   {\ul 91.3{\fontsize{6.3pt}{7pt}\selectfont$\pm$0.2}}&     \textbf{98.9{\fontsize{6.3pt}{7pt}\selectfont$\pm$0.1}}&     {\ul 95.4{\fontsize{6.3pt}{7pt}\selectfont$\pm$0.0}}&   {\ul 2.79{\fontsize{6.3pt}{7pt}\selectfont$\pm$0.18}}                        \\
{\color{softgreen}\ding{51}} & {\color{softgreen}\ding{51}} & {\color{softgreen}\ding{51}} & {\color{softgreen}\ding{51}} & \textbf{86.6{\fontsize{6.3pt}{7pt}\selectfont$\pm$1.3}}&    \textbf{94.6{\fontsize{6.3pt}{7pt}\selectfont$\pm$0.3}}&    \textbf{90.7{\fontsize{6.3pt}{7pt}\selectfont$\pm$0.2}}&   \textbf{91.9{\fontsize{6.3pt}{7pt}\selectfont$\pm$1.0}}&     \textbf{98.9{\fontsize{6.3pt}{7pt}\selectfont$\pm$0.0}}&     \textbf{95.6{\fontsize{6.3pt}{7pt}\selectfont$\pm$0.1}}&   \textbf{2.73{\fontsize{6.3pt}{7pt}\selectfont$\pm$0.05}}                       \\ \bottomrule
\end{tabular}
\end{center}
\end{table*}
\subsection{Visualization Results}
In this section, we visualize anomaly segmentation results for all categories of MVTec-3D AD datasets under the overlap setting.
As shown in \cref{fig:overlap}, we visualize the heatmap results of our method and PatchCore + FPFH~\cite{3d-ads}, M3DM~\cite{wang2023multimodal} and Shape-Guided~\cite{chu2023shape} with multi-modal inputs. Our method outperforms the previous ones by producing more accurate segmentation maps and exhibiting greater resilience to dataset noise. While the earlier approaches were often confounded by noise samples within the dataset, this is particularly noticeable in the Cable Gland, Dowel, Foam, and Peach results for PatchCore + FPFH, as well as the Foam and Rope results for Shape-Guided. 
More visualization results under the non-overlap setting is shown in \cref{app.vis}.

\subsection{Ablation Study}
We conduct an ablation study on the main components introduced in \cref{sec:method}, namely Stage I \& II two-stage sample-level denoising, intra-modality reference, Aligned Multi-Scale Point Cloud Feature Extraction and Noise-Focused Aggregation. The results are displayed in \cref{tab:ablation}. It was observed that the incremental inclusion of each component led to improvements in I-AUROC, P-AUROC, and AUPRO under both \textit{Overlap} and \textit{Non-Overlap} settings, particularly under the more challenging \textit{Overlap} setting. Besides these metrics, the Noise-level metric also clearly demonstrates that the model's capability for sample-level denoising progressively increased with the addition of each module.

\noindent \textbf{Different Scales.}
We also conduct an ablation study on the feature scales extracted in the Aligned Multi-Scale Point Cloud Feature Extraction, with results presented in \cref{tab:multi-scale}. The model performance varies across different scale configurations. Notably, when incorporating all scales, all performance metrics peaked, demonstrating that multi-scale consideration can enhance model performance. When the small scale is excluded, our model performs nearly as well as the full configuration, indicating that omitting small-scale processing has a relatively minor impact. This could be attributed to small-scale patches often containing too few point cloud points, many of which might be deemed insignificant and discarded during segmentation.
\renewcommand{\arraystretch}{1.3}
\begin{table}[t]
\begin{center}
\caption{\textbf{Ablation Study on Aligned Multi-scale Point Cloud Extraction.} w/o multi-scale represents removing all big. mid and small scales.}
\label{tab:multi-scale}
\setlength{\tabcolsep}{2.0pt}
\begin{tabular}{c|c|clccc}
\toprule
\multicolumn{2}{c}{Methods}                & \begin{tabular}[c]{@{}c@{}}w/o\\ multi-scale\end{tabular} &  \begin{tabular}[c]{@{}c@{}}w/o\\ big-scale\end{tabular} 
&\begin{tabular}[c]{@{}c@{}}w/o\\ mid-scale\end{tabular} & \begin{tabular}[c]{@{}c@{}}w/o\\ small-scale\end{tabular} & Full                 \\ \midrule
\multirow{3}{*}{\rotatebox{90}{Over}}     & I-AUROC $\uparrow$& 82.6{\fontsize{6.3pt}{7pt}\selectfont$\pm$0.7}&                                                  84.6{\fontsize{6.3pt}{7pt}\selectfont$\pm$1.0}
&83.7{\fontsize{6.3pt}{7pt}\selectfont$\pm$1.2}&                                               {\ul 85.3{\fontsize{6.3pt}{7pt}\selectfont$\pm$0.4}}&                                              \textbf{86.6{\fontsize{6.3pt}{7pt}\selectfont$\pm$1.3}}  \\
                          & P-AUROC $\uparrow$& 92.7{\fontsize{6.3pt}{7pt}\selectfont$\pm$0.5}&                                                  94.0{\fontsize{6.3pt}{7pt}\selectfont$\pm$0.3}
&93.6{\fontsize{6.3pt}{7pt}\selectfont$\pm$0.2}&                                               {\ul 94.2{\fontsize{6.3pt}{7pt}\selectfont$\pm$0.4}}&                                              \textbf{94.6{\fontsize{6.3pt}{7pt}\selectfont$\pm$0.3}}  \\
                          & AUPRO $\uparrow$   & 87.8{\fontsize{6.3pt}{7pt}\selectfont$\pm$0.3}&                                                  89.8{\fontsize{6.3pt}{7pt}\selectfont$\pm$0.3}
&89.4{\fontsize{6.3pt}{7pt}\selectfont$\pm$0.4}&                                               {\ul 90.2{\fontsize{6.3pt}{7pt}\selectfont$\pm$0.2}}&                                              \textbf{90.7{\fontsize{6.3pt}{7pt}\selectfont$\pm$0.2}}  \\ \midrule
\multirow{3}{*}{\rotatebox{90}{N-Over}} & I-AUROC $\uparrow$ & 89.2{\fontsize{6.3pt}{7pt}\selectfont$\pm$0.8}&                                                  89.6{\fontsize{6.3pt}{7pt}\selectfont$\pm$0.6}
&89.1{\fontsize{6.3pt}{7pt}\selectfont$\pm$0.9}&                                               {\ul 89.8{\fontsize{6.3pt}{7pt}\selectfont$\pm$0.1}}&                                              \textbf{91.9{\fontsize{6.3pt}{7pt}\selectfont$\pm$1.0}}  \\
                          & P-AUROC $\uparrow$ & 98.7{\fontsize{6.3pt}{7pt}\selectfont$\pm$0.0}&                                                  98.7{\fontsize{6.3pt}{7pt}\selectfont$\pm$0.1}
&{\ul 98.8{\fontsize{6.3pt}{7pt}\selectfont$\pm$0.1}}&                                         {\ul 98.8{\fontsize{6.3pt}{7pt}\selectfont$\pm$0.1}}&                                              \textbf{98.9{\fontsize{6.3pt}{7pt}\selectfont$\pm$0.0}}  \\
                          & AUPRO $\uparrow$  & 94.9{\fontsize{6.3pt}{7pt}\selectfont$\pm$0.1}&                                                  95.0{\fontsize{6.3pt}{7pt}\selectfont$\pm$0.2}
&{\ul 95.4{\fontsize{6.3pt}{7pt}\selectfont$\pm$0.2}}&                                         {\ul 95.4{\fontsize{6.3pt}{7pt}\selectfont$\pm$0.2}}&                                              \textbf{95.6{\fontsize{6.3pt}{7pt}\selectfont$\pm$0.1}}  \\ \midrule
\multicolumn{2}{c}{Noise-level $\downarrow$}     & 3.87{\fontsize{6.3pt}{7pt}\selectfont$\pm$0.08}&                                                 2.77{\fontsize{6.3pt}{7pt}\selectfont$\pm$0.18}&3.18{\fontsize{6.3pt}{7pt}\selectfont$\pm$0.20}&                                              {\ul 2.76{\fontsize{6.3pt}{7pt}\selectfont$\pm$0.07}}&                                             \textbf{2.73{\fontsize{6.3pt}{7pt}\selectfont$\pm$0.05}} \\ \bottomrule
\end{tabular}    
\end{center}
\end{table}
\renewcommand{\arraystretch}{1}
\renewcommand{\arraystretch}{1.2}
\begin{table}[t]
\begin{center}
\caption{\textbf{Exploring Aligned Multi-scale Point Cloud Extraction Setting.} $\sigma$ represents the thresholds for the minimum number of points required in a point cloud patch to be considered meaningful.}
\label{tab:pc_threshold}
\setlength{\tabcolsep}{5.2pt}
\begin{tabular}{c|c|cccc}
\toprule
\multicolumn{2}{c}{\fontsize{10pt}{11pt}\selectfont$\theta$}                              & 128                  & 256               & 512              & 1024        \\ \midrule
\multirow{3}{*}{\rotatebox{90}{Over}}   & I-AUROC $\uparrow$ & \textbf{86.6{\fontsize{6.3pt}{7pt}\selectfont$\pm$1.3}}  & {\ul 86.2{\fontsize{6.3pt}{7pt}\selectfont$\pm$0.6}}  & 85.9{\fontsize{6.3pt}{7pt}\selectfont$\pm$0.4}       & 84.0{\fontsize{6.3pt}{7pt}\selectfont$\pm$0.5}  \\
                                        & P-AUROC $\uparrow$ & \textbf{94.6{\fontsize{6.3pt}{7pt}\selectfont$\pm$0.3}}  & 94.3{\fontsize{6.3pt}{7pt}\selectfont$\pm$0.7}        & {\ul 94.4{\fontsize{6.3pt}{7pt}\selectfont$\pm$0.1}} & 93.4{\fontsize{6.3pt}{7pt}\selectfont$\pm$0.4}  \\
                                        & AUPRO   $\uparrow$ & \textbf{90.7{\fontsize{6.3pt}{7pt}\selectfont$\pm$0.2}}  & {\ul 90.3{\fontsize{6.3pt}{7pt}\selectfont$\pm$0.4}}  & 90.2{\fontsize{6.3pt}{7pt}\selectfont$\pm$0.4}       & 89.0{\fontsize{6.3pt}{7pt}\selectfont$\pm$0.5}  \\ \midrule
\multirow{3}{*}{\rotatebox{90}{N-Over}} & I-AUROC $\uparrow$ & \textbf{91.9{\fontsize{6.3pt}{7pt}\selectfont$\pm$1.0}}  & {\ul 91.4{\fontsize{6.3pt}{7pt}\selectfont$\pm$0.9}}  & 91.0{\fontsize{6.3pt}{7pt}\selectfont$\pm$0.2}       & 89.4{\fontsize{6.3pt}{7pt}\selectfont$\pm$0.5}  \\
                                        & P-AUROC $\uparrow$ & \textbf{98.9{\fontsize{6.3pt}{7pt}\selectfont$\pm$0.0}}  & {\ul 98.9{\fontsize{6.3pt}{7pt}\selectfont$\pm$0.0}}  & {\ul 98.9{\fontsize{6.3pt}{7pt}\selectfont$\pm$0.1}} & 98.7{\fontsize{6.3pt}{7pt}\selectfont$\pm$0.3}  \\
                                        & AUPRO   $\uparrow$ & \textbf{95.5{\fontsize{6.3pt}{7pt}\selectfont$\pm$0.1}}  & {\ul 95.5{\fontsize{6.3pt}{7pt}\selectfont$\pm$0.1}}  & 95.4{\fontsize{6.3pt}{7pt}\selectfont$\pm$0.1}       & 95.0{\fontsize{6.3pt}{7pt}\selectfont$\pm$0.3}  \\ \midrule
\multicolumn{2}{c}{Noise-level $\downarrow$}                   & \textbf{2.73{\fontsize{6.3pt}{7pt}\selectfont$\pm$0.05}} & {\bf 2.73{\fontsize{6.3pt}{7pt}\selectfont$\pm$0.05}} & 2.75{\fontsize{6.3pt}{7pt}\selectfont$\pm$0.13}      & 3.46{\fontsize{6.3pt}{7pt}\selectfont$\pm$0.10} \\ \bottomrule
\end{tabular}
\end{center}
\end{table}
\renewcommand{\arraystretch}{1}
\noindent \textbf{Point Cloud Threshold.}
We also perform an ablation study on the hyper-parameter $\theta$ introduced, representing the thresholds for the minimum number of points required in a point cloud patch to be considered meaningful. The experimental results are shown in \cref{tab:pc_threshold}. Given that the point cloud encoder used in our experiments has a minimum group size of 128, we commence our testing from this threshold. The findings indicate that for most metrics, a threshold of 128 points is the most appropriate, aligning with expectations as a lower threshold would mean considering more patches for computing the anomaly score, potentially leading to better accuracy. Therefore, after balancing the considerations of computational complexity and the accuracy of RGB-3D multi-modal anomaly detection, we opted for a threshold $\theta$ of 128 in this paper.

\noindent \textbf{$\lambda_{I}$ and $\lambda_{P}$.}
\renewcommand{\arraystretch}{1.3}
\begin{table}[t]
\centering
\caption{\textbf{Exploring RGB and Point Cloud Integration Setting.} $\lambda_{rgb}$ and $\lambda_{pc}$ are hyper-parameters controlling the extent to which RGB and point cloud modalities are integrated.}
\label{tab:lambda}
\setlength{\tabcolsep}{2.2pt}
\begin{tabular}{c|c|ccccc}
\toprule
\multicolumn{2}{c}{$\lambda_{rgb} \quad \lambda_{pc}$}  & 1.0 1.3             & 1.0 1.4             & 1.0 1.5              & 1.0 1.6             & 1.0 1.7             \\ \midrule
\multirow{3}{*}{\rotatebox{90}{Over}}   & I-AUROC $\uparrow$ & 86.1{\fontsize{6.3pt}{7pt}\selectfont$\pm$0.7}          & 85.6{\fontsize{6.3pt}{7pt}\selectfont$\pm$0.5}          & \textbf{86.6{\fontsize{6.3pt}{7pt}\selectfont$\pm$1.3}}  & 86.1{\fontsize{6.3pt}{7pt}\selectfont$\pm$1.0}          & 86.1{\fontsize{6.3pt}{7pt}\selectfont$\pm$1.0}          \\
                                        & P-AUROC $\uparrow$ & 94.3{\fontsize{6.3pt}{7pt}\selectfont$\pm$0.7}          & 94.2{\fontsize{6.3pt}{7pt}\selectfont$\pm$0.7}          & \textbf{94.6{\fontsize{6.3pt}{7pt}\selectfont$\pm$0.3}}  & 94.2{\fontsize{6.3pt}{7pt}\selectfont$\pm$0.7}          & 94.2{\fontsize{6.3pt}{7pt}\selectfont$\pm$0.0}          \\
                                        & AUPRO $\uparrow$   & 90.3{\fontsize{6.3pt}{7pt}\selectfont$\pm$0.4}          & 90.2{\fontsize{6.3pt}{7pt}\selectfont$\pm$0.3}          & \textbf{90.7{\fontsize{6.3pt}{7pt}\selectfont$\pm$0.3}}  & 90.3{\fontsize{6.3pt}{7pt}\selectfont$\pm$0.4}          & 90.3{\fontsize{6.3pt}{7pt}\selectfont$\pm$0.3}          \\ \midrule
\multirow{3}{*}{\rotatebox{90}{N-Over}} & I-AUROC $\uparrow$ & 91.3{\fontsize{6.3pt}{7pt}\selectfont$\pm$0.5}          & 90.7{\fontsize{6.3pt}{7pt}\selectfont$\pm$0.8}          & \textbf{91.9{\fontsize{6.3pt}{7pt}\selectfont$\pm$1.0}}  & 91.2{\fontsize{6.3pt}{7pt}\selectfont$\pm$1.1}          & 91.1{\fontsize{6.3pt}{7pt}\selectfont$\pm$0.8}          \\
                                        & P-AUROC $\uparrow$ & \textbf{98.9{\fontsize{6.3pt}{7pt}\selectfont$\pm$0.1}} & \textbf{98.9{\fontsize{6.3pt}{7pt}\selectfont$\pm$0.1}} & \textbf{98.9{\fontsize{6.3pt}{7pt}\selectfont$\pm$0.0}}  & \textbf{98.9{\fontsize{6.3pt}{7pt}\selectfont$\pm$0.0}} & \textbf{98.9{\fontsize{6.3pt}{7pt}\selectfont$\pm$0.1}} \\
                                        & AUPRO $\uparrow$   & 95.4{\fontsize{6.3pt}{7pt}\selectfont$\pm$0.2}          & 95.4{\fontsize{6.3pt}{7pt}\selectfont$\pm$0.1}          & \textbf{95.5{\fontsize{6.3pt}{7pt}\selectfont$\pm$0.1}}  & 95.4{\fontsize{6.3pt}{7pt}\selectfont$\pm$0.2}          & \textbf{95.5{\fontsize{6.3pt}{7pt}\selectfont$\pm$0.2}} \\ \midrule
\multicolumn{2}{c}{Noise-level $\downarrow$}                   & 2.74{\fontsize{6.3pt}{7pt}\selectfont$\pm$0.09}         & 2.75{\fontsize{6.3pt}{7pt}\selectfont$\pm$0.07}         & \textbf{2.71{\fontsize{6.3pt}{7pt}\selectfont$\pm$0.19}} & 2.72{\fontsize{6.3pt}{7pt}\selectfont$\pm$0.04}         & 2.75{\fontsize{6.3pt}{7pt}\selectfont$\pm$0.06}        \\ \bottomrule
\end{tabular}
\end{table}
\renewcommand{\arraystretch}{1.0}
\renewcommand{\arraystretch}{1.3}
\begin{table}[t]
\centering
\caption{\textbf{Exploring the Number of Intra-modal Reference Samples.} Ref Num represents the number of intra-modal reference samples selected.}
\label{tab:ref_num}
\setlength{\tabcolsep}{2.2pt}
\begin{tabular}{c|c|ccccc}
\toprule
\multicolumn{2}{c}{Ref Num}                       & 0           & 1                   & 2                   & 3           & 4                    \\ \midrule
\multirow{3}{*}{\rotatebox{90}{Over}}   & I-AUROC $\uparrow$ & 80.7{\fontsize{6.3pt}{7pt}\selectfont$\pm$0.9}  & 84.8{\fontsize{6.3pt}{7pt}\selectfont$\pm$0.7}          & 85.6{\fontsize{6.3pt}{7pt}\selectfont$\pm$1.5}          & 86.1{\fontsize{6.3pt}{7pt}\selectfont$\pm$0.5}  & \textbf{86.6{\fontsize{6.3pt}{7pt}\selectfont$\pm$1.3}}  \\
                                        & P-AUROC $\uparrow$ & 89.4{\fontsize{6.3pt}{7pt}\selectfont$\pm$1.3}  & 93.5{\fontsize{6.3pt}{7pt}\selectfont$\pm$0.4}          & 93.8{\fontsize{6.3pt}{7pt}\selectfont$\pm$0.3}          & 93.9{\fontsize{6.3pt}{7pt}\selectfont$\pm$0.2}  & \textbf{94.6{\fontsize{6.3pt}{7pt}\selectfont$\pm$0.3}}  \\
                                        & AUPRO $\uparrow$   & 85.5{\fontsize{6.3pt}{7pt}\selectfont$\pm$0.7}  & 89.3{\fontsize{6.3pt}{7pt}\selectfont$\pm$0.1}          & 89.8{\fontsize{6.3pt}{7pt}\selectfont$\pm$0.3}          & 90.0{\fontsize{6.3pt}{7pt}\selectfont$\pm$0.4}  & \textbf{90.7{\fontsize{6.3pt}{7pt}\selectfont$\pm$0.3}}  \\ \midrule
\multirow{3}{*}{\rotatebox{90}{N-Over}} & I-AUROC $\uparrow$ & 88.7{\fontsize{6.3pt}{7pt}\selectfont$\pm$0.9}  & 90.6{\fontsize{6.3pt}{7pt}\selectfont$\pm$0.4}          & 91.0{\fontsize{6.3pt}{7pt}\selectfont$\pm$0.9}          & 91.5{\fontsize{6.3pt}{7pt}\selectfont$\pm$0.4}  & \textbf{91.9{\fontsize{6.3pt}{7pt}\selectfont$\pm$1.0}}  \\
                                        & P-AUROC $\uparrow$ & 98.8{\fontsize{6.3pt}{7pt}\selectfont$\pm$0.1}  & 98.8{\fontsize{6.3pt}{7pt}\selectfont$\pm$0.1}          & \textbf{98.9{\fontsize{6.3pt}{7pt}\selectfont$\pm$0.0}} & 98.8{\fontsize{6.3pt}{7pt}\selectfont$\pm$0.1}  & \textbf{98.9{\fontsize{6.3pt}{7pt}\selectfont$\pm$0.0}}  \\
                                        & AUPRO $\uparrow$   & 94.9{\fontsize{6.3pt}{7pt}\selectfont$\pm$0.4}  & \textbf{95.5{\fontsize{6.3pt}{7pt}\selectfont$\pm$0.2}} & \textbf{95.5{\fontsize{6.3pt}{7pt}\selectfont$\pm$0.1}} & 95.4{\fontsize{6.3pt}{7pt}\selectfont$\pm$0.1}  & \textbf{95.5{\fontsize{6.3pt}{7pt}\selectfont$\pm$0.1}}  \\ \midrule
\multicolumn{2}{c}{Noise-level $\downarrow$}                   & 5.07{\fontsize{6.3pt}{7pt}\selectfont$\pm$0.13} & 3.20{\fontsize{6.3pt}{7pt}\selectfont$\pm$0.04}         & 2.88{\fontsize{6.3pt}{7pt}\selectfont$\pm$0.20}         & 2.82{\fontsize{6.3pt}{7pt}\selectfont$\pm$0.19} & \textbf{2.71{\fontsize{6.3pt}{7pt}\selectfont$\pm$0.19}}
\\ \bottomrule
\end{tabular}
\end{table}
\renewcommand{\arraystretch}{1.0}
To assess the extent to which RGB and Point Cloud modalities should be integrated, we conducted experiments with the hyper-parameters $\lambda_{I}$ and $\lambda_{P}$, which control the level of integration. The results of these experiments are presented in \cref{tab:lambda}. We observed that the model achieves optimal performance across all metrics for both anomaly detection and segmentation with $\lambda_{I}=1.0$ and $\lambda_{P}=1.5$. This indicates that enhancing the integration of the 3D Point Cloud modality can further improve performance. This outcome aligns with findings reported in \cref{sec:regular_anomaly,sec:noisy_anomaly}, where most methods performed better using purely 3D data rather than solely RGB data. This suggests that the 3D Point Cloud data in the MVTec 3D-AD dataset~\cite{mvtec3dad} contains richer information and facilitates more effective anomaly detection compared to RGB data within the same dataset.

\noindent \textbf{Number of Intra-Modal Reference Samples.}
To determine the appropriate number of intra-modal reference samples in Stage I, we conducted an ablation study on the quantity of these samples. The results are shown in \cref{tab:ref_num}. We conclude that increasing the number of intra-modal reference samples enhances the model's performance. This improvement is logical, as more reference samples mean more normal cases for the model to learn from, naturally boosting performance. However, selecting too many intra-modal reference samples can lead to the inclusion of noise samples and increase computational complexity. Therefore, in practical implementation, we opted for 4 intra-modal reference samples, striking a balance between model performance and computational efficiency.

\section{Conclusion}
In this paper, we first delve into the RGB-3D multi-modal noisy anomaly detection problem and have introduced a novel framework, M3DM-NR, to address the challenging task of RGB-3D multi-modal noisy industrial anomaly detection. Our approach systematically tackles the issues of reference selection, denoising, and final anomaly detection and segmentation through a three-stage process. 
In Stage I, we developed the Initial Feature Extraction, Suspected References Selection, and Suspected Anomaly Map Computation modules to filter normal samples and generate suspected anomaly maps, providing a robust foundation for subsequent stages. Stage II, termed Enhanced Multi-modal Denoising, leverages multi-scale feature comparison and weighting methods to refine and denoise the training samples, ensuring cleaner data for model training. Finally, Stage III integrates Point Feature Alignment, Unsupervised Feature Fusion, Noise Discriminative Coreset Selection, and Decision Layer Fusion to achieve precise anomaly detection and segmentation while effectively filtering out noise at the patch level. 
Extensive experiments demonstrate that our M3DM-NR framework significantly outperforms existing state-of-the-art methods in both detection and segmentation precision for 3D-RGB multi-modal noisy anomaly detection. The ablation studies further validate the effectiveness of each component within our framework, highlighting the importance of our systematic and hierarchical approach.

\noindent \textbf{Future Works.} 
Our work not only advances the field of industrial anomaly detection but also sets a new benchmark for handling noisy multi-modal data. Future research can build upon our framework to explore additional modalities and further enhance the robustness and accuracy of anomaly detection systems in practical industrial settings. 
Future work could consider more realistic methods of injecting noise into the training set. Currently, the approach of using anomalous samples from the test set as noise in the training set is rather naive. Future research could explore how noise naturally occurs in normal samples within real industrial production environments and attempt to construct new multi-modal noisy industrial detection datasets.
Additionally, future efforts could look into fine-tuning the CLIP model to better handle the task of multi-modal noisy industrial anomaly detection. The current method employs a training-free approach. The pre-trained CLIP model used in M3DM-NR is trained on a large-scale image dataset containing all categories of images. Subsequent work could consider fine-tuning the CLIP model on specific industrial detection datasets before using it for multi-modal noisy industrial anomaly detection.

\bibliographystyle{IEEEtran}
\bibliography{main}

\clearpage
\renewcommand\thefigure{A\arabic{figure}}
\renewcommand\thetable{A\arabic{table}}  
\renewcommand\theequation{A\arabic{equation}}
\setcounter{equation}{0}
\setcounter{table}{0}
\setcounter{figure}{0}
\appendix

\section*{Overview}
The appendix provides additional sections below to enhance the main manuscript: 
\begin{itemize}
    \item We report the P-AUROC for regular anomaly segmentation on MVTec 3D-AD in \cref{tab:pauroc_ori}.
    \item We report the P-AUROC for noisy anomaly segmentation on MVTec 3D-AD in \cref{tab:p-auroc overlap,tab:p-auroc}.
    \item We show the visualization results of noisy anomaly segmentation under \textit{Non-Overlap} setiing in \cref{fig:non-overlap}.
    \item We report the experiment results on Eycandies~\cite{eyecandies} dataset in \cref{tab:iauroc_overlap_eye,tab:iauroc_eye,tab:aupro_overlap_eye,tab:aupro_eye,tab:pauroc-overlap_eye,tab:pauroc_eye}.
    \item We reprot experiment results when injecting different percentages of noise into the training set in \cref{tab:iauroc_overlap_noise,tab:iauroc_noise,tab:aupro_overlap_noise,tab:aupro_noise,tab:pauroc_overlap_noise,tab:pauroc_noise}.
\end{itemize}
\section*{P-AUROC for regular anomaly segmentation on MVTec 3D-AD}
\label{app:paurco_regular}
\begin{table*}[!htbp]
\footnotesize
\centering
\caption{\textbf{P-AUROC score for regular anomaly segmentation of all categories of MVTec 3D-AD\cite{mvtec3dad} dataset.} Our method maintains the regular anomaly segmentation ability.  The results of baselines are from the ~\cite{mvtec3dad, 3d-ads, benchmarking}. Optimal and sub-optimal results are in \textbf{bold} and {\ul underlined}, respectively.}
\label{tab:pauroc_ori}
\setlength{\tabcolsep}{4.2pt}
\begin{tabular}{cc|cccccccccc|c}
\toprule
         & Method                         & Bagel         & \begin{tabular}[c]{@{}c@{}}Cable\\ Gland\end{tabular} & Carrot        & Cookie        & Dowel         & Foam          & Peach         & Potato        & Rope          & Tire          & Mean       \\ \midrule
\multirow{3}*{\rotatebox{90}{3D}}     & FPFH~\cite{3d-ads}              & \textbf{99.4} & \textbf{96.6}                                        & 99.9          & \textbf{94.6} & \textbf{96.6} & \textbf{92.7} & \textbf{99.6} & 99.9          & \textbf{99.6} & \textbf{99.0} & {\bf 97.8} \\
         & M3DM~\cite{wang2023multimodal}  & {\ul 98.1}    & 94.9                                                 & 99.7          & {\ul 93.2}    & {\ul 95.9}    & {\ul 92.5}    & {\ul 98.9}    & 99.5          & {\ul 99.4}    & {\ul 98.1}    & 97.0       \\
~        & Ours                           & {\ul 98.1}    & {\ul 95.0}                                           & 99.6          & {\ul 93.2}    & {\ul 95.9}    & 92.4          & {\ul 98.9}    & 99.6          & {\ul 99.4}    & {\ul 98.1}    & 97.0       \\ \midrule
\multirow{3}*{\rotatebox{90}{RGB}}    & PatchCore\cite{patchcore}      & 98.3          & 98.4                                                 & 98.0          & 97.4          & 97.2          & 84.9          & 97.6          & 98.3          & 98.7          & 97.7          & 96.7       \\
         & M3DM~\cite{wang2023multimodal} & \textbf{99.2} & \textbf{99.0}                                        & \textbf{99.4} & \textbf{97.7} & {\ul 98.3}    & \textbf{95.5} & \textbf{99.4} & \textbf{99.0} & \textbf{99.5} & {\ul 99.4}    & {\bf 98.7} \\
         & Ours                           & {\ul 99.1}    & \textbf{99.0}                                        & \textbf{99.4} & \textbf{97.7} & \textbf{98.4} & \textbf{95.5} & {\ul 99.3}    & \textbf{99.0} & \textbf{99.5} & \textbf{99.5} & {\bf 98.7} \\ \midrule
\multirow{4}*{\rotatebox{90}{RGB+3D}}  & AST\cite{ast}                  & -             & -                                                    & -             & -             & -             & -             & -             & -             & -             & -             & 97.6       \\
         & PatchCore + FPFH\cite{3d-ads}  & \textbf{99.6} & 99.2                                                 & \textbf{99.7} & \textbf{99.4} & 98.1          & 97.4          & \textbf{99.6} & \textbf{99.8} & 99.4          & 99.5          & {\bf 99.2} \\
         & M3DM~\cite{wang2023multimodal} & 99.5          & \textbf{99.3}                                        & \textbf{99.7} & {\ul 98.5}    & \textbf{98.5} & {\ul 98.4}    & \textbf{99.6} & 99.4          & \textbf{99.7} & \textbf{99.6} & {\bf 99.2} \\
         & Ours                           & \textbf{99.6} & \textbf{99.3}                                        & \textbf{99.7} & 97.9          & \textbf{98.5} & \textbf{98.9} & \textbf{99.6} & {\ul 99.5}    & \textbf{99.7} & \textbf{99.6} & {\bf 99.2} \\ \bottomrule
\end{tabular}
\end{table*}

In the regular anomaly segmentation setting, we compare our method with several 3D-based, RGB-based, and hybrid multi-modal 3D/RGB methods on MVTec-3D. 
\cref{tab:pauroc_ori} shows the segmentation results record with P-AUROC and we can conclude that our M3DM-NR also maintains the regular anomaly segmentation ability. 

\section*{P-AUROC for noisy anomaly segmentation on MVTec 3D-AD}
\label{app.pauroc_noisy}
In the main paper, we report the AUPRO score for anomaly segmentation.
In this section, we report the P-AUROC score under \textit{Overlap} and \textit{Non-Overlap} settings to further verify the segmentation performance of our method, as shown in \cref{tab:p-auroc overlap} and \cref{tab:p-auroc}. 

\textbf{3D.} On pure 3D anomaly segmentation, we get the highest P-AUROC and outperform Shape-Guided~\cite{chu2023shape} 0.8\% in \textit{Overlap} and M3DM~\cite{wang2023multimodal} 0.1\% in \textit{Non-Overlap}. This shows our method has better segmentation performance than the previous method and is more resistant to noise in the training dataset, and with our PFA, the Point Transformer is the better 3D feature extractor for this task.

\textbf{RGB.} Our P-AUROC in RGB domain is the same as SoftPatch~\cite{jiang2022softpatch} in \textit{Overlap} and the same as M3DM in \textit{Non-Overlap}. But our method has a lower standard deviation, which means our method is more robust.

\textbf{3D+RGB.} On 3D + RGB multi-modal anomaly segmentation, we get the best result with AUPRO and outperform Shape-Guided 0.6\% in \textit{Overlap} and PatchCore+FPFH~\cite{3d-ads} 0.1\% in \textit{Non-Overlap}. These results are contributed by our novel 3-stage multi-modal noise-resistant framework.
\begin{table*}[t]
\caption{\textbf{P-AUROC score for anomaly segmentation under \textit{Overlap} setting of all categories of MVTec 3D-AD.} Our method clearly outperforms other methods in 3D, RGB, and 3D + RGB settings, indicating the superior anomaly detection ability of our method. We report the mean and standard deviation over 3 random seeds for each measurement. Optimal and sub-optimal results are in \textbf{bold} and {\ul underlined}, respectively.}
\label{tab:p-auroc overlap}
\setlength{\tabcolsep}{4.7pt}
\begin{tabular}{cc|cccccccccc|c}
\toprule
\textbf{}               &{Method}         & Bagel                & \begin{tabular}[c]{@{}c@{}}Cable\\ Gland\end{tabular}          & Carrot               & Cookie               & Dowel                & Foam                 & Peach                & Potato               & Rope                 & Tire                 & Mean                \\ \midrule
\multirow{6}{*}{\rotatebox{90}{3D}}     &{SIFT}           & 69.8{\fontsize{6.5pt}{7pt}\selectfont$\pm$4.6}          & 80.6{\fontsize{6.5pt}{7pt}\selectfont$\pm$1.4}          & 95.4{\fontsize{6.5pt}{7pt}\selectfont$\pm$0.5}          & 78.2{\fontsize{6.5pt}{7pt}\selectfont$\pm$0.9}          & 70.6{\fontsize{6.5pt}{7pt}\selectfont$\pm$2.0}          & 77.1{\fontsize{6.5pt}{7pt}\selectfont$\pm$1.6}          & 66.6{\fontsize{6.5pt}{7pt}\selectfont$\pm$1.5}          & 76.4{\fontsize{6.5pt}{7pt}\selectfont$\pm$10.0}         & 91.1{\fontsize{6.5pt}{7pt}\selectfont$\pm$0.3}          & 75.9{\fontsize{6.5pt}{7pt}\selectfont$\pm$1.7}          & 78.2{\fontsize{6.5pt}{7pt}\selectfont$\pm$1.6}          \\
                        &{FPFH}           & 84.5{\fontsize{6.5pt}{7pt}\selectfont$\pm$2.7}          & 92.6{\fontsize{6.5pt}{7pt}\selectfont$\pm$0.2}          & 96.5{\fontsize{6.5pt}{7pt}\selectfont$\pm$0.4}          & 85.8{\fontsize{6.5pt}{7pt}\selectfont$\pm$0.6}          & 86.3{\fontsize{6.5pt}{7pt}\selectfont$\pm$2.2}          & 84.5{\fontsize{6.5pt}{7pt}\selectfont$\pm$1.4}          & 87.8{\fontsize{6.5pt}{7pt}\selectfont$\pm$1.2}          & 87.4{\fontsize{6.5pt}{7pt}\selectfont$\pm$2.0}          & 83.3{\fontsize{6.5pt}{7pt}\selectfont$\pm$0.7}          & 91.8{\fontsize{6.5pt}{7pt}\selectfont$\pm$0.7}          & 88.0{\fontsize{6.5pt}{7pt}\selectfont$\pm$0.3}          \\
                        &{AST}            & 89.5{\fontsize{6.5pt}{7pt}\selectfont$\pm$0.6}          & 90.2{\fontsize{6.5pt}{7pt}\selectfont$\pm$0.0}          & 96.9{\fontsize{6.5pt}{7pt}\selectfont$\pm$0.0}          & 85.7{\fontsize{6.5pt}{7pt}\selectfont$\pm$0.6}          & 86.8{\fontsize{6.5pt}{7pt}\selectfont$\pm$0.0}          & 86.4{\fontsize{6.5pt}{7pt}\selectfont$\pm$0.0}          & 93.5{\fontsize{6.5pt}{7pt}\selectfont$\pm$0.0}          & \textbf{97.0{\fontsize{6.5pt}{7pt}\selectfont$\pm$0.6}} & 89.6{\fontsize{6.5pt}{7pt}\selectfont$\pm$0.6}          & 89.9{\fontsize{6.5pt}{7pt}\selectfont$\pm$0.6}          & 90.6{\fontsize{6.5pt}{7pt}\selectfont$\pm$0.2}          \\
                        &{Shape-Guided}   & 93.5{\fontsize{6.5pt}{7pt}\selectfont$\pm$1.7}          & {\ul 94.2{\fontsize{6.5pt}{7pt}\selectfont$\pm$1.5}}    & \textbf{99.4{\fontsize{6.5pt}{7pt}\selectfont$\pm$0.6}} & \textbf{92.4{\fontsize{6.5pt}{7pt}\selectfont$\pm$1.2}} & 88.1{\fontsize{6.5pt}{7pt}\selectfont$\pm$6.5}          & \textbf{91.0{\fontsize{6.5pt}{7pt}\selectfont$\pm$3.1}} & 94.6{\fontsize{6.5pt}{7pt}\selectfont$\pm$0.8}          & 92.5{\fontsize{6.5pt}{7pt}\selectfont$\pm$3.8}          & 97.1{\fontsize{6.5pt}{7pt}\selectfont$\pm$1.9}          & 91.2{\fontsize{6.5pt}{7pt}\selectfont$\pm$1.2}          & {\ul 93.4{\fontsize{6.5pt}{7pt}\selectfont$\pm$0.6}}    \\
                        &{M3DM}           & {\ul 94.3{\fontsize{6.5pt}{7pt}\selectfont$\pm$1.1}}    & 94.2{\fontsize{6.5pt}{7pt}\selectfont$\pm$0.9}          & 98.9{\fontsize{6.5pt}{7pt}\selectfont$\pm$0.2}          & 90.6{\fontsize{6.5pt}{7pt}\selectfont$\pm$0.9}          & {\ul 89.8{\fontsize{6.5pt}{7pt}\selectfont$\pm$6.7}}    & 87.3{\fontsize{6.5pt}{7pt}\selectfont$\pm$2.8}          & {\ul 95.1{\fontsize{6.5pt}{7pt}\selectfont$\pm$1.0}}    & 91.9{\fontsize{6.5pt}{7pt}\selectfont$\pm$5.1}          & {\ul 98.0{\fontsize{6.5pt}{7pt}\selectfont$\pm$0.5}}    & {\ul 92.6{\fontsize{6.5pt}{7pt}\selectfont$\pm$3.8}}    & 93.3{\fontsize{6.5pt}{7pt}\selectfont$\pm$0.9}          \\
                        &{Ours}           & \textbf{96.6{\fontsize{6.5pt}{7pt}\selectfont$\pm$1.7}} & \textbf{94.3{\fontsize{6.5pt}{7pt}\selectfont$\pm$0.3}} & {\ul 99.3{\fontsize{6.5pt}{7pt}\selectfont$\pm$0.3}}    & {\ul 91.8{\fontsize{6.5pt}{7pt}\selectfont$\pm$0.4}}    & \textbf{90.2{\fontsize{6.5pt}{7pt}\selectfont$\pm$4.9}} & {\ul 88.8{\fontsize{6.5pt}{7pt}\selectfont$\pm$1.8}}    & \textbf{95.7{\fontsize{6.5pt}{7pt}\selectfont$\pm$1.2}} & {\ul 92.6{\fontsize{6.5pt}{7pt}\selectfont$\pm$3.2}}    & \textbf{98.7{\fontsize{6.5pt}{7pt}\selectfont$\pm$0.7}} & \textbf{94.3{\fontsize{6.5pt}{7pt}\selectfont$\pm$2.6}} & \textbf{94.2{\fontsize{6.5pt}{7pt}\selectfont$\pm$0.7}} \\ \midrule
\multirow{7}{*}{\rotatebox{90}{RGB}}    &{PaDim}          & {\ul 93.4{\fontsize{6.5pt}{7pt}\selectfont$\pm$0.9}}    & {\ul 93.9{\fontsize{6.5pt}{7pt}\selectfont$\pm$0.9}}    & {\ul 97.3{\fontsize{6.5pt}{7pt}\selectfont$\pm$0.4}}    & {\ul 90.6{\fontsize{6.5pt}{7pt}\selectfont$\pm$1.3}}    & {\ul 93.5{\fontsize{6.5pt}{7pt}\selectfont$\pm$6.1}}    & {\ul 88.4{\fontsize{6.5pt}{7pt}\selectfont$\pm$0.5}}    & 91.8{\fontsize{6.5pt}{7pt}\selectfont$\pm$4.5}          & 89.3{\fontsize{6.5pt}{7pt}\selectfont$\pm$1.2}          & {\ul 98.5{\fontsize{6.5pt}{7pt}\selectfont$\pm$0.2}}    & 93.8{\fontsize{6.5pt}{7pt}\selectfont$\pm$3.8}          & 93.1{\fontsize{6.5pt}{7pt}\selectfont$\pm$0.1}          \\
                        &{PatchCore}      & 75.2{\fontsize{6.5pt}{7pt}\selectfont$\pm$3.2}          & 73.6{\fontsize{6.5pt}{7pt}\selectfont$\pm$6.2}          & 80.0{\fontsize{6.5pt}{7pt}\selectfont$\pm$4.0}          & 80.2{\fontsize{6.5pt}{7pt}\selectfont$\pm$3.4}          & 71.1{\fontsize{6.5pt}{7pt}\selectfont$\pm$5.5}          & 75.4{\fontsize{6.5pt}{7pt}\selectfont$\pm$9.5}          & 68.9{\fontsize{6.5pt}{7pt}\selectfont$\pm$7.8}          & 72.3{\fontsize{6.5pt}{7pt}\selectfont$\pm$9.3}          & 64.9{\fontsize{6.5pt}{7pt}\selectfont$\pm$17.3}         & 75.3{\fontsize{6.5pt}{7pt}\selectfont$\pm$6.8}          & 73.7{\fontsize{6.5pt}{7pt}\selectfont$\pm$1.4}          \\
                        &{AST}            & 67.8{\fontsize{6.5pt}{7pt}\selectfont$\pm$0.0}          & 74.2{\fontsize{6.5pt}{7pt}\selectfont$\pm$0.0}          & 54.2{\fontsize{6.5pt}{7pt}\selectfont$\pm$0.0}          & 65.8{\fontsize{6.5pt}{7pt}\selectfont$\pm$0.6}          & 68.9{\fontsize{6.5pt}{7pt}\selectfont$\pm$0.0}          & 63.4{\fontsize{6.5pt}{7pt}\selectfont$\pm$0.6}          & 57.5{\fontsize{6.5pt}{7pt}\selectfont$\pm$0.6}          & 61.1{\fontsize{6.5pt}{7pt}\selectfont$\pm$0.6}          & 57.2{\fontsize{6.5pt}{7pt}\selectfont$\pm$0.0}          & 69.3{\fontsize{6.5pt}{7pt}\selectfont$\pm$0.6}          & 63.9{\fontsize{6.5pt}{7pt}\selectfont$\pm$0.1}          \\
                        &{Shape-Guided}   & 78.0{\fontsize{6.5pt}{7pt}\selectfont$\pm$3.5}          & 91.2{\fontsize{6.5pt}{7pt}\selectfont$\pm$1.4}          & 93.1{\fontsize{6.5pt}{7pt}\selectfont$\pm$1.1}          & 84.7{\fontsize{6.5pt}{7pt}\selectfont$\pm$0.3}          & 90.1{\fontsize{6.5pt}{7pt}\selectfont$\pm$0.4}          & 73.8{\fontsize{6.5pt}{7pt}\selectfont$\pm$1.6}          & 82.8{\fontsize{6.5pt}{7pt}\selectfont$\pm$1.1}          & 89.3{\fontsize{6.5pt}{7pt}\selectfont$\pm$0.8}          & 88.6{\fontsize{6.5pt}{7pt}\selectfont$\pm$0.2}          & 88.8{\fontsize{6.5pt}{7pt}\selectfont$\pm$0.3}          & 86.0{\fontsize{6.5pt}{7pt}\selectfont$\pm$0.6}          \\
                        &{SoftPatch}      & 90.4{\fontsize{6.5pt}{7pt}\selectfont$\pm$1.7}          & 91.9{\fontsize{6.5pt}{7pt}\selectfont$\pm$4.1}          & 96.9{\fontsize{6.5pt}{7pt}\selectfont$\pm$1.1}          & 87.7{\fontsize{6.5pt}{7pt}\selectfont$\pm$2.2}          & \textbf{94.8{\fontsize{6.5pt}{7pt}\selectfont$\pm$4.6}} & \textbf{96.5{\fontsize{6.5pt}{7pt}\selectfont$\pm$4.9}} & \textbf{94.4{\fontsize{6.5pt}{7pt}\selectfont$\pm$0.5}} & \textbf{90.9{\fontsize{6.5pt}{7pt}\selectfont$\pm$0.7}} & 96.7{\fontsize{6.5pt}{7pt}\selectfont$\pm$1.6}          & \textbf{97.3{\fontsize{6.5pt}{7pt}\selectfont$\pm$0.8}} & \textbf{93.8{\fontsize{6.5pt}{7pt}\selectfont$\pm$0.5}}    \\
                        &{M3DM}           & 68.8{\fontsize{6.5pt}{7pt}\selectfont$\pm$5.0}          & 77.0{\fontsize{6.5pt}{7pt}\selectfont$\pm$1.8}          & 77.2{\fontsize{6.5pt}{7pt}\selectfont$\pm$2.6}          & 77.1{\fontsize{6.5pt}{7pt}\selectfont$\pm$0.4}          & 71.8{\fontsize{6.5pt}{7pt}\selectfont$\pm$2.0}          & 68.9{\fontsize{6.5pt}{7pt}\selectfont$\pm$2.3}          & 65.8{\fontsize{6.5pt}{7pt}\selectfont$\pm$1.7}          & 65.8{\fontsize{6.5pt}{7pt}\selectfont$\pm$3.8}          & 60.5{\fontsize{6.5pt}{7pt}\selectfont$\pm$2.3}          & 75.2{\fontsize{6.5pt}{7pt}\selectfont$\pm$1.4}          & 70.8{\fontsize{6.5pt}{7pt}\selectfont$\pm$1.1}          \\
                        &{Ours}           & \textbf{98.5{\fontsize{6.5pt}{7pt}\selectfont$\pm$0.5}} & \textbf{95.8{\fontsize{6.5pt}{7pt}\selectfont$\pm$1.6}} & \textbf{98.7{\fontsize{6.5pt}{7pt}\selectfont$\pm$0.4}} & \textbf{95.0{\fontsize{6.5pt}{7pt}\selectfont$\pm$1.1}} & 88.5{\fontsize{6.5pt}{7pt}\selectfont$\pm$5.9}          & 85.9{\fontsize{6.5pt}{7pt}\selectfont$\pm$1.7}          & {\ul 93.4{\fontsize{6.5pt}{7pt}\selectfont$\pm$2.6}}    & {\ul 89.5{\fontsize{6.5pt}{7pt}\selectfont$\pm$1.0}}    & \textbf{98.6{\fontsize{6.5pt}{7pt}\selectfont$\pm$0.3}} & {\ul 94.6{\fontsize{6.5pt}{7pt}\selectfont$\pm$0.4}}    & \textbf{93.8{\fontsize{6.5pt}{7pt}\selectfont$\pm$0.7}} \\ \midrule
\multirow{5}{*}{\rotatebox{90}{3D+RGB}} &{PatchCore+FPFH} & 69.1{\fontsize{6.5pt}{7pt}\selectfont$\pm$4.8}          & 77.0{\fontsize{6.5pt}{7pt}\selectfont$\pm$1.8}          & 77.4{\fontsize{6.5pt}{7pt}\selectfont$\pm$2.6}          & 78.4{\fontsize{6.5pt}{7pt}\selectfont$\pm$0.4}          & 71.5{\fontsize{6.5pt}{7pt}\selectfont$\pm$2.1}          & 69.3{\fontsize{6.5pt}{7pt}\selectfont$\pm$1.5}          & 66.0{\fontsize{6.5pt}{7pt}\selectfont$\pm$1.7}          & 65.8{\fontsize{6.5pt}{7pt}\selectfont$\pm$3.8}          & 60.5{\fontsize{6.5pt}{7pt}\selectfont$\pm$2.3}          & 75.2{\fontsize{6.5pt}{7pt}\selectfont$\pm$1.4}          & 71.0{\fontsize{6.5pt}{7pt}\selectfont$\pm$0.9}          \\
                        &{AST}            & 90.7{\fontsize{6.5pt}{7pt}\selectfont$\pm$0.6}          & 94.3{\fontsize{6.5pt}{7pt}\selectfont$\pm$0.6}          & 97.5{\fontsize{6.5pt}{7pt}\selectfont$\pm$0.0}          & 89.4{\fontsize{6.5pt}{7pt}\selectfont$\pm$0.0}          & 90.6{\fontsize{6.5pt}{7pt}\selectfont$\pm$0.6}          & {\ul 89.4{\fontsize{6.5pt}{7pt}\selectfont$\pm$0.0}}    & 93.3{\fontsize{6.5pt}{7pt}\selectfont$\pm$0.6}          & \textbf{96.9{\fontsize{6.5pt}{7pt}\selectfont$\pm$0.6}} & 90.6{\fontsize{6.5pt}{7pt}\selectfont$\pm$0.6}          & 93.6{\fontsize{6.5pt}{7pt}\selectfont$\pm$0.0}          & 92.6{\fontsize{6.5pt}{7pt}\selectfont$\pm$0.2}          \\
                        &{Shape-Guided}   & {\ul 91.0{\fontsize{6.5pt}{7pt}\selectfont$\pm$1.7}}    & {\ul 94.7{\fontsize{6.5pt}{7pt}\selectfont$\pm$0.4}}    & {\ul 98.1{\fontsize{6.5pt}{7pt}\selectfont$\pm$0.2}}    & {\ul 90.9{\fontsize{6.5pt}{7pt}\selectfont$\pm$0.1}}    & \textbf{91.6{\fontsize{6.5pt}{7pt}\selectfont$\pm$5.3}} & \textbf{90.8{\fontsize{6.5pt}{7pt}\selectfont$\pm$1.6}} & \textbf{95.3{\fontsize{6.5pt}{7pt}\selectfont$\pm$0.3}} & {\ul 95.8{\fontsize{6.5pt}{7pt}\selectfont$\pm$4.6}}    & {\ul 96.0{\fontsize{6.5pt}{7pt}\selectfont$\pm$0.3}}    & \textbf{95.5{\fontsize{6.5pt}{7pt}\selectfont$\pm$2.7}} & {\ul 94.0{\fontsize{6.5pt}{7pt}\selectfont$\pm$1.0}}    \\
                        &{M3DM}           & 69.8{\fontsize{6.5pt}{7pt}\selectfont$\pm$4.7}          & 77.0{\fontsize{6.5pt}{7pt}\selectfont$\pm$2.0}          & 77.4{\fontsize{6.5pt}{7pt}\selectfont$\pm$2.6}          & 79.2{\fontsize{6.5pt}{7pt}\selectfont$\pm$0.5}          & 71.9{\fontsize{6.5pt}{7pt}\selectfont$\pm$3.1}          & 74.0{\fontsize{6.5pt}{7pt}\selectfont$\pm$2.4}          & 66.2{\fontsize{6.5pt}{7pt}\selectfont$\pm$1.8}          & 66.2{\fontsize{6.5pt}{7pt}\selectfont$\pm$3.8}          & 61.8{\fontsize{6.5pt}{7pt}\selectfont$\pm$2.5}          & 75.6{\fontsize{6.5pt}{7pt}\selectfont$\pm$1.3}          & 71.9{\fontsize{6.5pt}{7pt}\selectfont$\pm$1.2}          \\
                        &{Ours}           & \textbf{99.1{\fontsize{6.5pt}{7pt}\selectfont$\pm$0.5}} & \textbf{95.8{\fontsize{6.5pt}{7pt}\selectfont$\pm$1.7}} & \textbf{99.0{\fontsize{6.5pt}{7pt}\selectfont$\pm$0.5}} & \textbf{95.8{\fontsize{6.5pt}{7pt}\selectfont$\pm$1.0}} & {\ul 90.7{\fontsize{6.5pt}{7pt}\selectfont$\pm$2.8}}    & 88.1{\fontsize{6.5pt}{7pt}\selectfont$\pm$2.5}          & {\ul 93.8{\fontsize{6.5pt}{7pt}\selectfont$\pm$2.8}}    & 89.8{\fontsize{6.5pt}{7pt}\selectfont$\pm$1.1}          & \textbf{98.8{\fontsize{6.5pt}{7pt}\selectfont$\pm$0.2}} & {\ul 94.9{\fontsize{6.5pt}{7pt}\selectfont$\pm$0.5}}    & \textbf{94.6{\fontsize{6.5pt}{7pt}\selectfont$\pm$0.3}}                        \\ \bottomrule
\end{tabular}
\end{table*}
\begin{table*}[t]
\caption{\textbf{P-AUROC score for anomaly segmentation under \textit{Non-Overlap} setting of all categories of MVTec 3D-AD.} Our method clearly outperforms other methods in 3D, RGB, and 3D + RGB settings, indicating the superior anomaly detection ability of our method. We report the mean and standard deviation over 3 random seeds for each measurement. Optimal and sub-optimal results are in \textbf{bold} and {\ul underlined}, respectively.}
\label{tab:p-auroc}
\setlength{\tabcolsep}{4.7pt}
\begin{tabular}{cc|cccccccccc|c}
\toprule
\textbf{}               &{Method}         & Bagel                & \begin{tabular}[c]{@{}c@{}}Cable\\ Gland\end{tabular}          & Carrot               & Cookie               & Dowel                & Foam                 & Peach                & Potato               & Rope                 & Tire                 & Mean                \\ \midrule
\multirow{6}{*}{\rotatebox{90}{3D}}     &{SIFT}           & 94.0{\fontsize{6.5pt}{7pt}\selectfont$\pm$3.1}          & 94.2{\fontsize{6.5pt}{7pt}\selectfont$\pm$3.0}          & 93.9{\fontsize{6.5pt}{7pt}\selectfont$\pm$4.9}          & {\ul 93.0{\fontsize{6.5pt}{7pt}\selectfont$\pm$1.9}}    & {\ul 95.7{\fontsize{6.5pt}{7pt}\selectfont$\pm$1.3}}    & 92.3{\fontsize{6.5pt}{7pt}\selectfont$\pm$2.9}          & 96.0{\fontsize{6.5pt}{7pt}\selectfont$\pm$2.8}          & 98.1{\fontsize{6.5pt}{7pt}\selectfont$\pm$2.9}          & {\ul 99.2{\fontsize{6.5pt}{7pt}\selectfont$\pm$0.7}}    & {\ul 98.6{\fontsize{6.5pt}{7pt}\selectfont$\pm$0.7}}    & 95.5{\fontsize{6.5pt}{7pt}\selectfont$\pm$0.6}          \\
                        &{FPFH}           & 97.7{\fontsize{6.5pt}{7pt}\selectfont$\pm$0.5}          & 93.8{\fontsize{6.5pt}{7pt}\selectfont$\pm$2.4}          & 95.2{\fontsize{6.5pt}{7pt}\selectfont$\pm$4.5}          & \textbf{94.4{\fontsize{6.5pt}{7pt}\selectfont$\pm$0.4}} & \textbf{96.5{\fontsize{6.5pt}{7pt}\selectfont$\pm$0.5}} & {\ul 92.6{\fontsize{6.5pt}{7pt}\selectfont$\pm$1.4}}    & 96.1{\fontsize{6.5pt}{7pt}\selectfont$\pm$1.1}          & {\ul 99.1{\fontsize{6.5pt}{7pt}\selectfont$\pm$1.2}}    & 98.9{\fontsize{6.5pt}{7pt}\selectfont$\pm$1.2}          & \textbf{99.1{\fontsize{6.5pt}{7pt}\selectfont$\pm$0.1}} & 96.3{\fontsize{6.5pt}{7pt}\selectfont$\pm$0.5}          \\
                        &{AST}            & 96.4{\fontsize{6.5pt}{7pt}\selectfont$\pm$0.6}          & 91.3{\fontsize{6.5pt}{7pt}\selectfont$\pm$0.6}          & 98.3{\fontsize{6.5pt}{7pt}\selectfont$\pm$0.6}          & 91.9{\fontsize{6.5pt}{7pt}\selectfont$\pm$0.6}          & 86.4{\fontsize{6.5pt}{7pt}\selectfont$\pm$0.6}          & \textbf{94.0{\fontsize{6.5pt}{7pt}\selectfont$\pm$0.6}} & \textbf{98.9{\fontsize{6.5pt}{7pt}\selectfont$\pm$0.6}} & \textbf{99.3{\fontsize{6.5pt}{7pt}\selectfont$\pm$0.6}} & 92.9{\fontsize{6.5pt}{7pt}\selectfont$\pm$0.0}          & 93.8{\fontsize{6.5pt}{7pt}\selectfont$\pm$0.0}          & 94.3{\fontsize{6.5pt}{7pt}\selectfont$\pm$0.3}          \\
                        &{Shape-Guided}   & {\ul 98.4{\fontsize{6.5pt}{7pt}\selectfont$\pm$0.5}}    & 94.4{\fontsize{6.5pt}{7pt}\selectfont$\pm$1.5}          & 98.8{\fontsize{6.5pt}{7pt}\selectfont$\pm$1.0}          & 93.0{\fontsize{6.5pt}{7pt}\selectfont$\pm$1.7}          & 95.5{\fontsize{6.5pt}{7pt}\selectfont$\pm$0.6}          & 90.9{\fontsize{6.5pt}{7pt}\selectfont$\pm$4.0}          & {\ul 98.7{\fontsize{6.5pt}{7pt}\selectfont$\pm$1.2}}    & 97.9{\fontsize{6.5pt}{7pt}\selectfont$\pm$2.0}          & 98.0{\fontsize{6.5pt}{7pt}\selectfont$\pm$0.6}          & 97.7{\fontsize{6.5pt}{7pt}\selectfont$\pm$0.1}          & 96.3{\fontsize{6.5pt}{7pt}\selectfont$\pm$0.6}          \\
                        &{M3DM}           & 97.9{\fontsize{6.5pt}{7pt}\selectfont$\pm$0.3}          & \textbf{94.8{\fontsize{6.5pt}{7pt}\selectfont$\pm$0.3}} & \textbf{99.6{\fontsize{6.5pt}{7pt}\selectfont$\pm$0.1}} & 91.9{\fontsize{6.5pt}{7pt}\selectfont$\pm$0.9}          & 94.8{\fontsize{6.5pt}{7pt}\selectfont$\pm$2.0}          & 91.5{\fontsize{6.5pt}{7pt}\selectfont$\pm$3.1}          & 97.5{\fontsize{6.5pt}{7pt}\selectfont$\pm$2.2}          & {\ul 99.1{\fontsize{6.5pt}{7pt}\selectfont$\pm$0.1}}    & \textbf{99.3{\fontsize{6.5pt}{7pt}\selectfont$\pm$0.1}} & 97.5{\fontsize{6.5pt}{7pt}\selectfont$\pm$1.0}          & {\ul 96.4{\fontsize{6.5pt}{7pt}\selectfont$\pm$0.7}}    \\
                        &{Ours}           & \textbf{98.6{\fontsize{6.5pt}{7pt}\selectfont$\pm$0.2}} & {\ul 94.6{\fontsize{6.5pt}{7pt}\selectfont$\pm$0.2}}    & \textbf{99.6{\fontsize{6.5pt}{7pt}\selectfont$\pm$0.1}} & 92.4{\fontsize{6.5pt}{7pt}\selectfont$\pm$0.6}          & 95.4{\fontsize{6.5pt}{7pt}\selectfont$\pm$0.9}          & 90.8{\fontsize{6.5pt}{7pt}\selectfont$\pm$2.9}          & 98.1{\fontsize{6.5pt}{7pt}\selectfont$\pm$1.1}          & 98.2{\fontsize{6.5pt}{7pt}\selectfont$\pm$1.6}          & {\ul 99.2{\fontsize{6.5pt}{7pt}\selectfont$\pm$0.3}}    & 97.7{\fontsize{6.5pt}{7pt}\selectfont$\pm$0.6}          & \textbf{96.5{\fontsize{6.5pt}{7pt}\selectfont$\pm$0.7}} \\ \midrule
\multirow{7}{*}{\rotatebox{90}{RGB}}    &{PaDim}          & 97.5{\fontsize{6.5pt}{7pt}\selectfont$\pm$1.2}          & 96.1{\fontsize{6.5pt}{7pt}\selectfont$\pm$0.9}          & 97.9{\fontsize{6.5pt}{7pt}\selectfont$\pm$0.2}          & 95.1{\fontsize{6.5pt}{7pt}\selectfont$\pm$0.2}          & 97.8{\fontsize{6.5pt}{7pt}\selectfont$\pm$0.4}          & {\ul 99.6{\fontsize{6.5pt}{7pt}\selectfont$\pm$0.3}}    & {\ul 99.1{\fontsize{6.5pt}{7pt}\selectfont$\pm$0.2}}    & {\ul 98.6{\fontsize{6.5pt}{7pt}\selectfont$\pm$0.3}}    & 98.8{\fontsize{6.5pt}{7pt}\selectfont$\pm$0.4}          & 99.2{\fontsize{6.5pt}{7pt}\selectfont$\pm$0.2}          & 98.0{\fontsize{6.5pt}{7pt}\selectfont$\pm$0.2}          \\
                        &{PatchCore}      & 96.0{\fontsize{6.5pt}{7pt}\selectfont$\pm$0.2}          & \textbf{98.9{\fontsize{6.5pt}{7pt}\selectfont$\pm$0.0}} & 98.1{\fontsize{6.5pt}{7pt}\selectfont$\pm$1.9}          & {\ul 96.7{\fontsize{6.5pt}{7pt}\selectfont$\pm$0.4}}    & {\ul 98.9{\fontsize{6.5pt}{7pt}\selectfont$\pm$0.1}}    & \textbf{99.9{\fontsize{6.5pt}{7pt}\selectfont$\pm$0.0}} & 98.1{\fontsize{6.5pt}{7pt}\selectfont$\pm$0.1}          & 96.3{\fontsize{6.5pt}{7pt}\selectfont$\pm$2.3}          & 98.8{\fontsize{6.5pt}{7pt}\selectfont$\pm$0.8}          & {\ul 99.2{\fontsize{6.5pt}{7pt}\selectfont$\pm$0.6}}    & 98.1{\fontsize{6.5pt}{7pt}\selectfont$\pm$0.5}          \\
                        &{AST}            & 88.5{\fontsize{6.5pt}{7pt}\selectfont$\pm$0.6}          & 92.7{\fontsize{6.5pt}{7pt}\selectfont$\pm$0.6}          & 65.8{\fontsize{6.5pt}{7pt}\selectfont$\pm$0.6}          & 79.4{\fontsize{6.5pt}{7pt}\selectfont$\pm$1.0}          & 96.0{\fontsize{6.5pt}{7pt}\selectfont$\pm$0.6}          & 80.6{\fontsize{6.5pt}{7pt}\selectfont$\pm$1.0}          & 84.4{\fontsize{6.5pt}{7pt}\selectfont$\pm$0.6}          & 80.0{\fontsize{6.5pt}{7pt}\selectfont$\pm$0.0}          & 89.1{\fontsize{6.5pt}{7pt}\selectfont$\pm$0.6}          & 85.6{\fontsize{6.5pt}{7pt}\selectfont$\pm$0.6}          & 84.2{\fontsize{6.5pt}{7pt}\selectfont$\pm$0.2}          \\
                        &{Shape-Guided}   & 94.5{\fontsize{6.5pt}{7pt}\selectfont$\pm$0.4}          & 97.2{\fontsize{6.5pt}{7pt}\selectfont$\pm$0.4}          & 98.3{\fontsize{6.5pt}{7pt}\selectfont$\pm$0.2}          & 95.0{\fontsize{6.5pt}{7pt}\selectfont$\pm$0.6}          & 98.1{\fontsize{6.5pt}{7pt}\selectfont$\pm$0.1}          & 87.8{\fontsize{6.5pt}{7pt}\selectfont$\pm$0.8}          & 95.1{\fontsize{6.5pt}{7pt}\selectfont$\pm$0.2}          & 96.1{\fontsize{6.5pt}{7pt}\selectfont$\pm$0.3}          & 97.3{\fontsize{6.5pt}{7pt}\selectfont$\pm$1.0}          & 97.5{\fontsize{6.5pt}{7pt}\selectfont$\pm$0.5}          & 95.7{\fontsize{6.5pt}{7pt}\selectfont$\pm$0.1}          \\
                        &{SoftPatch}      & 96.3{\fontsize{6.5pt}{7pt}\selectfont$\pm$0.5}          & 98.5{\fontsize{6.5pt}{7pt}\selectfont$\pm$0.3}          & {\ul 99.2{\fontsize{6.5pt}{7pt}\selectfont$\pm$0.1}}    & \textbf{96.8{\fontsize{6.5pt}{7pt}\selectfont$\pm$0.4}} & \textbf{98.9{\fontsize{6.5pt}{7pt}\selectfont$\pm$0.1}} & 98.9{\fontsize{6.5pt}{7pt}\selectfont$\pm$1.0}          & 98.3{\fontsize{6.5pt}{7pt}\selectfont$\pm$0.3}          & 97.1{\fontsize{6.5pt}{7pt}\selectfont$\pm$1.3}          & 98.2{\fontsize{6.5pt}{7pt}\selectfont$\pm$0.4}          & 98.5{\fontsize{6.5pt}{7pt}\selectfont$\pm$1.0}          & 98.1{\fontsize{6.5pt}{7pt}\selectfont$\pm$0.1}          \\
                        &{M3DM}           & {\ul 98.8{\fontsize{6.5pt}{7pt}\selectfont$\pm$0.3}}    & {\ul 98.9{\fontsize{6.5pt}{7pt}\selectfont$\pm$0.6}}    & 99.0{\fontsize{6.5pt}{7pt}\selectfont$\pm$0.6}          & 96.6{\fontsize{6.5pt}{7pt}\selectfont$\pm$0.3}          & 98.4{\fontsize{6.5pt}{7pt}\selectfont$\pm$0.4}          & 93.9{\fontsize{6.5pt}{7pt}\selectfont$\pm$0.8}          & \textbf{99.1{\fontsize{6.5pt}{7pt}\selectfont$\pm$0.1}} & \textbf{98.7{\fontsize{6.5pt}{7pt}\selectfont$\pm$0.3}} & \textbf{99.5{\fontsize{6.5pt}{7pt}\selectfont$\pm$0.1}} & \textbf{99.4{\fontsize{6.5pt}{7pt}\selectfont$\pm$0.1}} & \textbf{98.2{\fontsize{6.5pt}{7pt}\selectfont$\pm$0.2}} \\
                        &{Ours}           & \textbf{99.0{\fontsize{6.5pt}{7pt}\selectfont$\pm$0.2}} & {\ul 98.9{\fontsize{6.5pt}{7pt}\selectfont$\pm$0.2}}    & \textbf{99.2{\fontsize{6.5pt}{7pt}\selectfont$\pm$0.1}} & 96.4{\fontsize{6.5pt}{7pt}\selectfont$\pm$0.3}          & 97.7{\fontsize{6.5pt}{7pt}\selectfont$\pm$0.8}          & 94.6{\fontsize{6.5pt}{7pt}\selectfont$\pm$0.4}          & 98.9{\fontsize{6.5pt}{7pt}\selectfont$\pm$0.1}          & 98.4{\fontsize{6.5pt}{7pt}\selectfont$\pm$0.5}          & {\ul 99.4{\fontsize{6.5pt}{7pt}\selectfont$\pm$0.2}}    & 98.9{\fontsize{6.5pt}{7pt}\selectfont$\pm$0.1}          & \textbf{98.2{\fontsize{6.5pt}{7pt}\selectfont$\pm$0.0}} \\ \midrule
\multirow{5}{*}{\rotatebox{90}{3D+RGB}} &{PatchCore+FPFH} & \textbf{99.4{\fontsize{6.5pt}{7pt}\selectfont$\pm$0.1}} & 98.8{\fontsize{6.5pt}{7pt}\selectfont$\pm$0.5}          & 99.3{\fontsize{6.5pt}{7pt}\selectfont$\pm$0.6}          & \textbf{98.1{\fontsize{6.5pt}{7pt}\selectfont$\pm$1.6}} & 98.1{\fontsize{6.5pt}{7pt}\selectfont$\pm$0.5}          & {\ul 97.5{\fontsize{6.5pt}{7pt}\selectfont$\pm$0.2}}    & {\ul 99.3{\fontsize{6.5pt}{7pt}\selectfont$\pm$0.1}}    & 98.6{\fontsize{6.5pt}{7pt}\selectfont$\pm$0.1}          & 99.5{\fontsize{6.5pt}{7pt}\selectfont$\pm$0.1}          & 99.1{\fontsize{6.5pt}{7pt}\selectfont$\pm$0.6}          & {\ul 98.8{\fontsize{6.5pt}{7pt}\selectfont$\pm$0.1}}    \\
                        &{AST}            & 97.4{\fontsize{6.5pt}{7pt}\selectfont$\pm$0.6}          & 97.1{\fontsize{6.5pt}{7pt}\selectfont$\pm$0.6}          & {\ul 99.5{\fontsize{6.5pt}{7pt}\selectfont$\pm$0.6}}    & 94.0{\fontsize{6.5pt}{7pt}\selectfont$\pm$0.0}          & 91.3{\fontsize{6.5pt}{7pt}\selectfont$\pm$0.6}          & 97.1{\fontsize{6.5pt}{7pt}\selectfont$\pm$0.6}          & 98.7{\fontsize{6.5pt}{7pt}\selectfont$\pm$0.0}          & 98.7{\fontsize{6.5pt}{7pt}\selectfont$\pm$0.6}          & 93.2{\fontsize{6.5pt}{7pt}\selectfont$\pm$0.6}          & 96.9{\fontsize{6.5pt}{7pt}\selectfont$\pm$0.0}          & 96.4{\fontsize{6.5pt}{7pt}\selectfont$\pm$0.1}          \\
                        &{Shape-Guided}   & 97.6{\fontsize{6.5pt}{7pt}\selectfont$\pm$0.1}          & 98.2{\fontsize{6.5pt}{7pt}\selectfont$\pm$0.3}          & 99.5{\fontsize{6.5pt}{7pt}\selectfont$\pm$0.1}          & 97.0{\fontsize{6.5pt}{7pt}\selectfont$\pm$0.3}          & \textbf{98.9{\fontsize{6.5pt}{7pt}\selectfont$\pm$0.1}} & 97.2{\fontsize{6.5pt}{7pt}\selectfont$\pm$0.2}          & 98.6{\fontsize{6.5pt}{7pt}\selectfont$\pm$0.1}          & {\ul 99.1{\fontsize{6.5pt}{7pt}\selectfont$\pm$1.0}}    & 98.9{\fontsize{6.5pt}{7pt}\selectfont$\pm$0.5}          & \textbf{99.6{\fontsize{6.5pt}{7pt}\selectfont$\pm$0.2}} & 98.5{\fontsize{6.5pt}{7pt}\selectfont$\pm$0.2}          \\
                        &{M3DM}           & 98.9{\fontsize{6.5pt}{7pt}\selectfont$\pm$0.2}          & \textbf{99.1{\fontsize{6.5pt}{7pt}\selectfont$\pm$0.1}} & 99.3{\fontsize{6.5pt}{7pt}\selectfont$\pm$0.6}          & 96.8{\fontsize{6.5pt}{7pt}\selectfont$\pm$0.3}          & 97.5{\fontsize{6.5pt}{7pt}\selectfont$\pm$0.9}          & 96.0{\fontsize{6.5pt}{7pt}\selectfont$\pm$0.3}          & 99.2{\fontsize{6.5pt}{7pt}\selectfont$\pm$0.1}          & 99.0{\fontsize{6.5pt}{7pt}\selectfont$\pm$0.3}          & \textbf{99.7{\fontsize{6.5pt}{7pt}\selectfont$\pm$0.1}} & {\ul 99.3{\fontsize{6.5pt}{7pt}\selectfont$\pm$0.1}}    & 98.5{\fontsize{6.5pt}{7pt}\selectfont$\pm$0.1}          \\
                        &{Ours}           & \textbf{99.4{\fontsize{6.5pt}{7pt}\selectfont$\pm$0.1}} & {\ul 99.0{\fontsize{6.5pt}{7pt}\selectfont$\pm$0.1}}    & \textbf{99.5{\fontsize{6.5pt}{7pt}\selectfont$\pm$0.1}} & {\ul 97.2{\fontsize{6.5pt}{7pt}\selectfont$\pm$0.2}}    & {\ul 98.2{\fontsize{6.5pt}{7pt}\selectfont$\pm$0.4}}    & \textbf{98.1{\fontsize{6.5pt}{7pt}\selectfont$\pm$0.4}} & \textbf{99.3{\fontsize{6.5pt}{7pt}\selectfont$\pm$0.1}} & \textbf{99.2{\fontsize{6.5pt}{7pt}\selectfont$\pm$0.0}} & {\ul 99.6{\fontsize{6.5pt}{7pt}\selectfont$\pm$0.1}}    & {\ul 99.2{\fontsize{6.5pt}{7pt}\selectfont$\pm$0.1}}    & \textbf{98.9{\fontsize{6.5pt}{7pt}\selectfont$\pm$0.0}}                        \\ \bottomrule
\end{tabular}
\end{table*}

\section*{Visualization results of \textit{Non-Overlap} setiing}
\label{app.vis}
\begin{figure*}[t]
    \centering
    \includegraphics[width=1.0\linewidth]{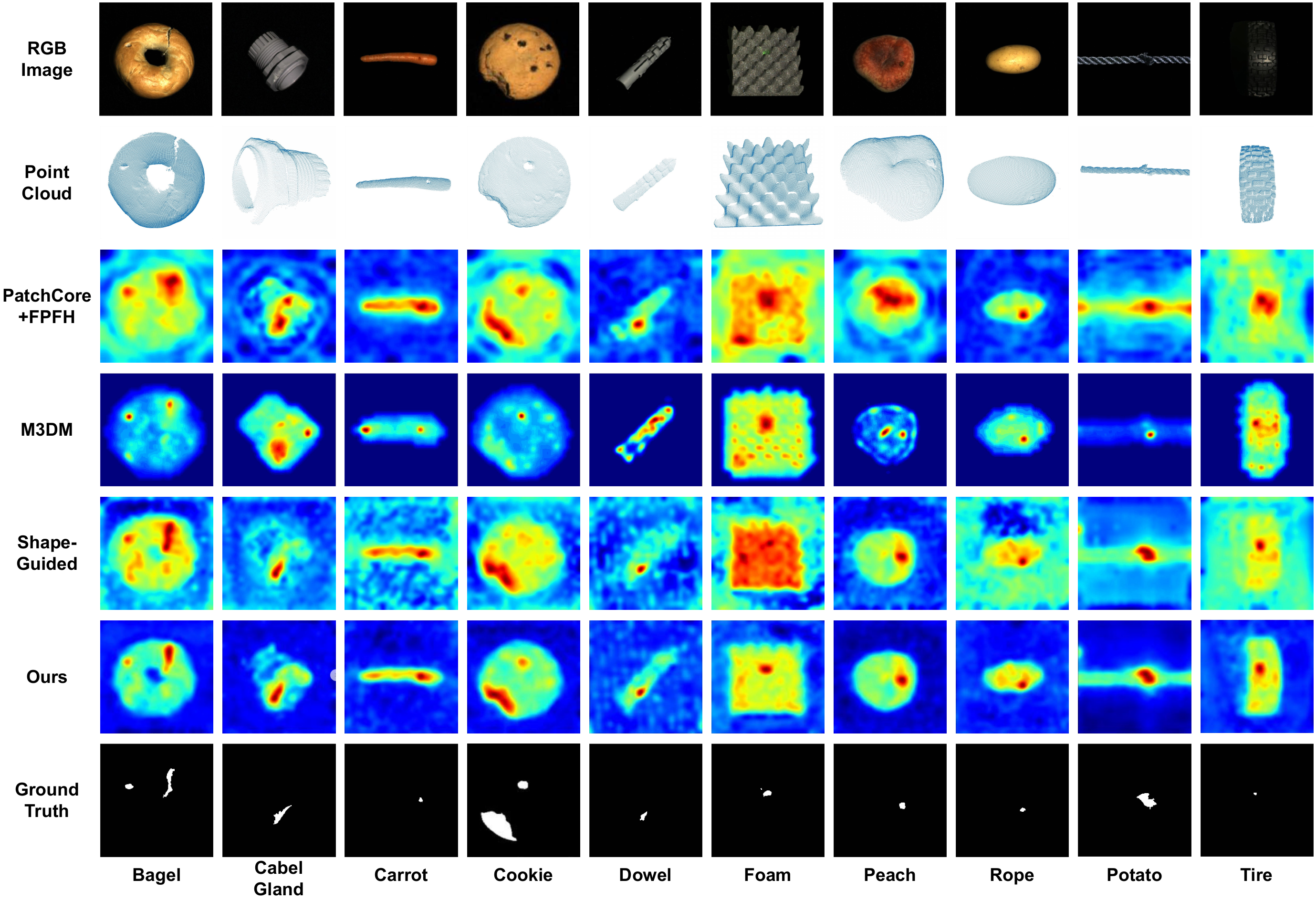}
    \caption{Heatmap of our anomaly segmentation results (multi-modal inputs) under \textit{Non-Overlap} setting. Compared with existing methods, our method remains unaffected by noise and outputs a more accurate segmentation region.}
    \label{fig:non-overlap}
\end{figure*}

In this section, we visualize anomaly segmentation results for all categories of MVTec-3D AD datasets under \textit{Non-Overlap} setting.
As shown in \cref{fig:non-overlap}, we visualize the heatmap results of our method and PatchCore + FPFH~\cite{3d-ads}, M3DM~\cite{wang2023multimodal} and Shape-Guided~\cite{chu2023shape} with multi-modal inputs. Compared with previous methods, our method gets better segmentation maps.

\section*{Eyecandies}
\label{app.eyecan}
We have noticed that recently a new dataset Eyecandies~\cite{eyecandies} provides multimodel information of 10 categories of candies, and each category contains 1000 samples for training, 50 labeled samples for public testing and 400 unlabeled samples for private testing. 
The source dataset provides 6 RGB images, which are in different light conditions, a depth map, and a normal map of each sample.
In this section, we convert the Eyecandies dataset to the format supported by M3DM-NR.
In detail, we use the environment light image as our input RGB data, and for 3D data, we first convert the depth image to point clouds with internal parameters, then we remove the background points with point coordinates.
For computation efficiency, we use only less than 400 samples from each category for training.
Because the public test dataset only contains 25 normal and 25 anomalous samples, which doesn't meet 10\% of the size of training dataset, we implement the \textit{Overlap} and \textit{Non-Overlap} setting differently. For \textit{Overlap} setting, we only conduct experiments of 5\% noise by selecting 400 images from training dataset and 20 images from public test dataset as the whole noisy training dataset. For \textit{Non-Overlap} setting, as the private test dataset contains 200 normal samples and 200 anomalous samples mixed together, we random select 80 samples from the private test dataset and regard it as 40 normal samples and 40 anomalous samples. These 80 samples, along with 320 normal samples selected from the training dataset, make up of the whole noisy training dataset. We report the mean and standard deviation over 3 random seeds for each measurement.

As illustrated in \cref{tab:iauroc_overlap_eye,tab:iauroc_eye,tab:aupro_overlap_eye,tab:aupro_eye,tab:pauroc-overlap_eye,tab:pauroc_eye}, we report the best I-AUCROC, AUPRO and P-AUCROC scores. under both \textit{Overlap} and \textit{Non-Overlap} settings.
\begin{table*}[t]
\begin{center}
\caption{\textbf{I-AUROC score for anomaly detection under \textit{Overlap} setting of all categories in Eyecandies~\cite{eyecandies}.} Our method clearly outperforms other methods in 3D + RGB settings, indicating the superior anomaly detection ability of our method. We report the mean and standard deviation over 3 random seeds for each measurement. Optimal and sub-optimal results are in \textbf{bold} and {\ul underlined}, respectively.}
\label{tab:iauroc_overlap_eye}
\setlength{\tabcolsep}{3.5pt}
\begin{tabular}{cc|cccccccccc|c}
\toprule
\textbf{}               & Method         & \begin{tabular}[c]{@{}c@{}}Candy\\ Cane\end{tabular} & \begin{tabular}[c]{@{}c@{}}Chocolate\\ Cookie\end{tabular} & \begin{tabular}[c]{@{}c@{}}Chocolate\\ Praline\end{tabular} & Confetto            & \begin{tabular}[c]{@{}c@{}}Gummy\\ Bear\end{tabular} & \begin{tabular}[c]{@{}c@{}}Hazelnut\\ Truffle\end{tabular} & \begin{tabular}[c]{@{}c@{}}Licorice\\ Sandwich\end{tabular} & Lollipop            & \begin{tabular}[c]{@{}c@{}}Marsh-\\ mallow\end{tabular}         & \begin{tabular}[c]{@{}c@{}}Peppermint\\ Candy\end{tabular} & Mean                \\ \midrule
\multirow{5}{*}{\rotatebox{90}{3D+RGB}} & PatchCore+FPFH & 11.4{\fontsize{6.5pt}{7pt}\selectfont$\pm$2.8}          & 19.2{\fontsize{6.5pt}{7pt}\selectfont$\pm$3.6}          & 20.9{\fontsize{6.5pt}{7pt}\selectfont$\pm$1.6}          & 19.7{\fontsize{6.5pt}{7pt}\selectfont$\pm$0.9}          & 25.1{\fontsize{6.5pt}{7pt}\selectfont$\pm$5.9}          & {\ul 20.8{\fontsize{6.5pt}{7pt}\selectfont$\pm$4.7}}    & 17.6{\fontsize{6.5pt}{7pt}\selectfont$\pm$1.2}          & 24.5{\fontsize{6.5pt}{7pt}\selectfont$\pm$3.6}          & 24.8{\fontsize{6.5pt}{7pt}\selectfont$\pm$1.4}          & 19.1{\fontsize{6.5pt}{7pt}\selectfont$\pm$1.3}          & 20.3{\fontsize{6.5pt}{7pt}\selectfont$\pm$0.3}          \\
                        & AST            & 8.0{\fontsize{6.5pt}{7pt}\selectfont$\pm$0.6}           & 13.8{\fontsize{6.5pt}{7pt}\selectfont$\pm$0.6}          & 6.7{\fontsize{6.5pt}{7pt}\selectfont$\pm$0.6}           & 10.9{\fontsize{6.5pt}{7pt}\selectfont$\pm$0.6}          & 16.7{\fontsize{6.5pt}{7pt}\selectfont$\pm$0.6}          & 10.9{\fontsize{6.5pt}{7pt}\selectfont$\pm$0.6}          & 18.4{\fontsize{6.5pt}{7pt}\selectfont$\pm$0.6}          & 24.0{\fontsize{6.5pt}{7pt}\selectfont$\pm$1.0}          & 9.4{\fontsize{6.5pt}{7pt}\selectfont$\pm$0.0}           & 13.7{\fontsize{6.5pt}{7pt}\selectfont$\pm$0.0}          & 13.4{\fontsize{6.5pt}{7pt}\selectfont$\pm$0.2}          \\
                        & Shape-Guided   & 9.1{\fontsize{6.5pt}{7pt}\selectfont$\pm$4.5}           & 18.5{\fontsize{6.5pt}{7pt}\selectfont$\pm$1.0}          & 15.3{\fontsize{6.5pt}{7pt}\selectfont$\pm$2.5}          & 24.7{\fontsize{6.5pt}{7pt}\selectfont$\pm$2.2}          & 15.5{\fontsize{6.5pt}{7pt}\selectfont$\pm$3.0}          & 11.8{\fontsize{6.5pt}{7pt}\selectfont$\pm$2.4}          & 15.8{\fontsize{6.5pt}{7pt}\selectfont$\pm$0.6}          & 25.7{\fontsize{6.5pt}{7pt}\selectfont$\pm$1.2}          & 25.9{\fontsize{6.5pt}{7pt}\selectfont$\pm$1.3}          & 23.6{\fontsize{6.5pt}{7pt}\selectfont$\pm$3.1}          & 18.6{\fontsize{6.5pt}{7pt}\selectfont$\pm$0.8}          \\
                        & M3DM           & {\ul 17.0{\fontsize{6.5pt}{7pt}\selectfont$\pm$3.6}}    & {\ul 30.5{\fontsize{6.5pt}{7pt}\selectfont$\pm$4.2}}    & {\ul 39.6{\fontsize{6.5pt}{7pt}\selectfont$\pm$2.7}}    & {\ul 41.9{\fontsize{6.5pt}{7pt}\selectfont$\pm$1.6}}    & {\ul 39.4{\fontsize{6.5pt}{7pt}\selectfont$\pm$3.4}}    & 20.7{\fontsize{6.5pt}{7pt}\selectfont$\pm$3.8}          & {\ul 28.2{\fontsize{6.5pt}{7pt}\selectfont$\pm$2.3}}    & {\ul 33.1{\fontsize{6.5pt}{7pt}\selectfont$\pm$3.4}}    & {\ul 54.6{\fontsize{6.5pt}{7pt}\selectfont$\pm$0.4}}    & {\ul 50.9{\fontsize{6.5pt}{7pt}\selectfont$\pm$0.9}}    & {\ul 35.6{\fontsize{6.5pt}{7pt}\selectfont$\pm$0.9}}    \\
                        & Ours           & \textbf{33.5{\fontsize{6.5pt}{7pt}\selectfont$\pm$3.4}} & \textbf{74.9{\fontsize{6.5pt}{7pt}\selectfont$\pm$4.5}} & \textbf{76.9{\fontsize{6.5pt}{7pt}\selectfont$\pm$5.5}} & \textbf{89.3{\fontsize{6.5pt}{7pt}\selectfont$\pm$3.0}} & \textbf{55.8{\fontsize{6.5pt}{7pt}\selectfont$\pm$6.1}} & \textbf{48.0{\fontsize{6.5pt}{7pt}\selectfont$\pm$5.7}} & \textbf{79.4{\fontsize{6.5pt}{7pt}\selectfont$\pm$5.2}} & \textbf{65.0{\fontsize{6.5pt}{7pt}\selectfont$\pm$4.9}} & \textbf{98.9{\fontsize{6.5pt}{7pt}\selectfont$\pm$1.0}} & \textbf{70.5{\fontsize{6.5pt}{7pt}\selectfont$\pm$2.4}} & \textbf{69.2{\fontsize{6.5pt}{7pt}\selectfont$\pm$1.9}}
                        \\ \bottomrule
\end{tabular}
\end{center}
\end{table*}

\begin{table*}[t]
\begin{center}
\caption{\textbf{I-AUROC score for anomaly detection under \textit{Non-Overlap} setting of all categories in Eyecandies~\cite{eyecandies}.} Our method clearly outperforms other methods in 3D + RGB settings, indicating the superior anomaly detection ability of our method. We report the mean and standard deviation over 3 random seeds for each measurement. Optimal and sub-optimal results are in \textbf{bold} and {\ul underlined}, respectively.}
\label{tab:iauroc_eye}
\setlength{\tabcolsep}{3.5pt}
\begin{tabular}{cc|cccccccccc|c}
\toprule
\textbf{}               & Method         & \begin{tabular}[c]{@{}c@{}}Candy\\ Cane\end{tabular} & \begin{tabular}[c]{@{}c@{}}Chocolate\\ Cookie\end{tabular} & \begin{tabular}[c]{@{}c@{}}Chocolate\\ Praline\end{tabular} & Confetto            & \begin{tabular}[c]{@{}c@{}}Gummy\\ Bear\end{tabular} & \begin{tabular}[c]{@{}c@{}}Hazelnut\\ Truffle\end{tabular} & \begin{tabular}[c]{@{}c@{}}Licorice\\ Sandwich\end{tabular} & Lollipop            & \begin{tabular}[c]{@{}c@{}}Marsh-\\ mallow\end{tabular}         & \begin{tabular}[c]{@{}c@{}}Peppermint\\ Candy\end{tabular} & Mean                \\ \midrule
\multirow{5}{*}{\rotatebox{90}{3D+RGB}} & PatchCore+FPFH & \textbf{55.4{\fontsize{6.5pt}{7pt}\selectfont$\pm$0.8}} & 86.4{\fontsize{6.5pt}{7pt}\selectfont$\pm$2.3}          & 72.2{\fontsize{6.5pt}{7pt}\selectfont$\pm$2.1}          & 94.3{\fontsize{6.5pt}{7pt}\selectfont$\pm$1.9}          & 71.5{\fontsize{6.5pt}{7pt}\selectfont$\pm$3.5}          & 49.2{\fontsize{6.5pt}{7pt}\selectfont$\pm$5.3}          & 80.9{\fontsize{6.5pt}{7pt}\selectfont$\pm$1.0}          & 82.0{\fontsize{6.5pt}{7pt}\selectfont$\pm$1.2}          & 99.1{\fontsize{6.5pt}{7pt}\selectfont$\pm$0.8}          & 85.8{\fontsize{6.5pt}{7pt}\selectfont$\pm$4.7}          & 77.7{\fontsize{6.5pt}{7pt}\selectfont$\pm$0.6}          \\
                        & AST            & 47.7{\fontsize{6.5pt}{7pt}\selectfont$\pm$0.6}          & {\ul 93.4{\fontsize{6.5pt}{7pt}\selectfont$\pm$1.0}}    & 78.3{\fontsize{6.5pt}{7pt}\selectfont$\pm$0.6}          & 93.9{\fontsize{6.5pt}{7pt}\selectfont$\pm$0.0}          & 74.7{\fontsize{6.5pt}{7pt}\selectfont$\pm$0.6}          & \textbf{66.2{\fontsize{6.5pt}{7pt}\selectfont$\pm$1.0}} & 83.1{\fontsize{6.5pt}{7pt}\selectfont$\pm$0.6}          & \textbf{87.3{\fontsize{6.5pt}{7pt}\selectfont$\pm$0.0}} & 99.4{\fontsize{6.5pt}{7pt}\selectfont$\pm$0.6}          & 92.9{\fontsize{6.5pt}{7pt}\selectfont$\pm$0.6}          & 81.7{\fontsize{6.5pt}{7pt}\selectfont$\pm$0.2}          \\
                        & Shape-Guided   & 49.4{\fontsize{6.5pt}{7pt}\selectfont$\pm$0.5}          & \textbf{94.8{\fontsize{6.5pt}{7pt}\selectfont$\pm$1.3}} & 77.5{\fontsize{6.5pt}{7pt}\selectfont$\pm$2.2}          & 93.9{\fontsize{6.5pt}{7pt}\selectfont$\pm$1.1}          & 74.8{\fontsize{6.5pt}{7pt}\selectfont$\pm$0.9}          & {\ul 64.9{\fontsize{6.5pt}{7pt}\selectfont$\pm$2.0}}    & {\ul 83.3{\fontsize{6.5pt}{7pt}\selectfont$\pm$0.4}}    & {\ul 86.0{\fontsize{6.5pt}{7pt}\selectfont$\pm$1.6}}    & {\ul 99.6{\fontsize{6.5pt}{7pt}\selectfont$\pm$0.1}}    & 92.6{\fontsize{6.5pt}{7pt}\selectfont$\pm$1.3}          & 81.7{\fontsize{6.5pt}{7pt}\selectfont$\pm$0.7}          \\
                        & M3DM           & 53.9{\fontsize{6.5pt}{7pt}\selectfont$\pm$5.0}          & 90.1{\fontsize{6.5pt}{7pt}\selectfont$\pm$0.6}          & \textbf{89.4{\fontsize{6.5pt}{7pt}\selectfont$\pm$0.8}} & \textbf{98.4{\fontsize{6.5pt}{7pt}\selectfont$\pm$0.4}} & {\ul 81.5{\fontsize{6.5pt}{7pt}\selectfont$\pm$1.0}}    & 52.3{\fontsize{6.5pt}{7pt}\selectfont$\pm$1.8}          & 78.4{\fontsize{6.5pt}{7pt}\selectfont$\pm$1.1}          & 83.3{\fontsize{6.5pt}{7pt}\selectfont$\pm$1.7}          & 99.5{\fontsize{6.5pt}{7pt}\selectfont$\pm$0.2}          & \textbf{99.4{\fontsize{6.5pt}{7pt}\selectfont$\pm$0.2}} & {\ul 82.6{\fontsize{6.5pt}{7pt}\selectfont$\pm$0.5}}    \\
                        & Ours           & {\ul 54.5{\fontsize{6.5pt}{7pt}\selectfont$\pm$7.7}}    & 85.6{\fontsize{6.5pt}{7pt}\selectfont$\pm$0.5}          & {\ul 88.9{\fontsize{6.5pt}{7pt}\selectfont$\pm$2.1}}    & {\ul 97.2{\fontsize{6.5pt}{7pt}\selectfont$\pm$0.7}}    & \textbf{82.2{\fontsize{6.5pt}{7pt}\selectfont$\pm$6.1}} & 54.3{\fontsize{6.5pt}{7pt}\selectfont$\pm$2.5}          & \textbf{86.8{\fontsize{6.5pt}{7pt}\selectfont$\pm$0.2}} & 85.6{\fontsize{6.5pt}{7pt}\selectfont$\pm$1.2}          & \textbf{99.8{\fontsize{6.5pt}{7pt}\selectfont$\pm$0.1}} & {\ul 98.6{\fontsize{6.5pt}{7pt}\selectfont$\pm$0.7}}    & \textbf{83.3{\fontsize{6.5pt}{7pt}\selectfont$\pm$0.6}}
                        \\ \bottomrule
\end{tabular}
\end{center}
\end{table*}

\begin{table*}[t]
\begin{center}
\caption{\textbf{AUPRO score for anomaly segmentation under \textit{Overlap} setting of all categories in Eyecandies~\cite{eyecandies}.} Our method clearly outperforms other methods in 3D + RGB settings, indicating the superior anomaly detection ability of our method. We report the mean and standard deviation over 3 random seeds for each measurement. Optimal and sub-optimal results are in \textbf{bold} and {\ul underlined}, respectively.}
\label{tab:aupro_overlap_eye}
\setlength{\tabcolsep}{3.5pt}
\begin{tabular}{cc|cccccccccc|c}
\toprule
\textbf{}               & Method         & \begin{tabular}[c]{@{}c@{}}Candy\\ Cane\end{tabular} & \begin{tabular}[c]{@{}c@{}}Chocolate\\ Cookie\end{tabular} & \begin{tabular}[c]{@{}c@{}}Chocolate\\ Praline\end{tabular} & Confetto            & \begin{tabular}[c]{@{}c@{}}Gummy\\ Bear\end{tabular} & \begin{tabular}[c]{@{}c@{}}Hazelnut\\ Truffle\end{tabular} & \begin{tabular}[c]{@{}c@{}}Licorice\\ Sandwich\end{tabular} & Lollipop            & \begin{tabular}[c]{@{}c@{}}Marsh-\\ mallow\end{tabular}         & \begin{tabular}[c]{@{}c@{}}Peppermint\\ Candy\end{tabular} & Mean                \\ \midrule
\multirow{4}{*}{\rotatebox{90}{3D+RGB}} & PatchCore+FPFH & 16.7{\fontsize{6.5pt}{7pt}\selectfont$\pm$1.9}          & 20.5{\fontsize{6.5pt}{7pt}\selectfont$\pm$2.1}          & 15.6{\fontsize{6.5pt}{7pt}\selectfont$\pm$1.4}          & 18.7{\fontsize{6.5pt}{7pt}\selectfont$\pm$3.2}          & 22.2{\fontsize{6.5pt}{7pt}\selectfont$\pm$4.8}          & 18.3{\fontsize{6.5pt}{7pt}\selectfont$\pm$2.6}          & 17.3{\fontsize{6.5pt}{7pt}\selectfont$\pm$1.9}          & 25.8{\fontsize{6.5pt}{7pt}\selectfont$\pm$6.2}          & 19.0{\fontsize{6.5pt}{7pt}\selectfont$\pm$1.1}          & 19.6{\fontsize{6.5pt}{7pt}\selectfont$\pm$0.5}          & 19.4{\fontsize{6.5pt}{7pt}\selectfont$\pm$0.6 }         \\
                         & Shape-Guided   & \textbf{65.6{\fontsize{6.5pt}{7pt}\selectfont$\pm$0.6}} & {\ul 44.1{\fontsize{6.5pt}{7pt}\selectfont$\pm$0.9}}    & {\ul 21.1{\fontsize{6.5pt}{7pt}\selectfont$\pm$0.9}}    & {\ul 57.8{\fontsize{6.5pt}{7pt}\selectfont$\pm$4.2}}    & {\ul 52.8{\fontsize{6.5pt}{7pt}\selectfont$\pm$2.2}}    & 20.7{\fontsize{6.5pt}{7pt}\selectfont$\pm$1.7}          & {\ul 34.3{\fontsize{6.5pt}{7pt}\selectfont$\pm$2.0}}    & \textbf{84.0{\fontsize{6.5pt}{7pt}\selectfont$\pm$3.2}} & {\ul 59.1{\fontsize{6.5pt}{7pt}\selectfont$\pm$3.0}}    & \textbf{57.6{\fontsize{6.5pt}{7pt}\selectfont$\pm$2.2}} & {\ul 49.7{\fontsize{6.5pt}{7pt}\selectfont$\pm$1.1}}    \\
                         & M3DM           & 21.7{\fontsize{6.5pt}{7pt}\selectfont$\pm$3.2}          & 21.0{\fontsize{6.5pt}{7pt}\selectfont$\pm$2.3}          & 18.3{\fontsize{6.5pt}{7pt}\selectfont$\pm$0.2}          & 18.8{\fontsize{6.5pt}{7pt}\selectfont$\pm$3.2}          & 23.3{\fontsize{6.5pt}{7pt}\selectfont$\pm$5.1}          & {\ul 21.5{\fontsize{6.5pt}{7pt}\selectfont$\pm$2.1}}    & 17.6{\fontsize{6.5pt}{7pt}\selectfont$\pm$2.3}          & 26.7{\fontsize{6.5pt}{7pt}\selectfont$\pm$4.7}          & 19.1{\fontsize{6.5pt}{7pt}\selectfont$\pm$1.2}          & 20.2{\fontsize{6.5pt}{7pt}\selectfont$\pm$0.0}          & 20.8{\fontsize{6.5pt}{7pt}\selectfont$\pm$0.7 }         \\
                         & Ours           & {\ul 50.5{\fontsize{6.5pt}{7pt}\selectfont$\pm$2.5}}    & \textbf{82.1{\fontsize{6.5pt}{7pt}\selectfont$\pm$2.9}} & \textbf{66.8{\fontsize{6.5pt}{7pt}\selectfont$\pm$2.2}} & \textbf{89.7{\fontsize{6.5pt}{7pt}\selectfont$\pm$2.7}} & \textbf{60.7{\fontsize{6.5pt}{7pt}\selectfont$\pm$4.0}} & \textbf{59.3{\fontsize{6.5pt}{7pt}\selectfont$\pm$2.2}} & \textbf{80.8{\fontsize{6.5pt}{7pt}\selectfont$\pm$1.6}} & {\ul 70.3{\fontsize{6.5pt}{7pt}\selectfont$\pm$2.9}}    & \textbf{94.1{\fontsize{6.5pt}{7pt}\selectfont$\pm$2.6}} & {\ul 55.9{\fontsize{6.5pt}{7pt}\selectfont$\pm$3.4}}    & \textbf{71.0{\fontsize{6.5pt}{7pt}\selectfont$\pm$0.8}}
                        \\ \bottomrule
\end{tabular}
\end{center}
\end{table*}

\begin{table*}[t]
\begin{center}
\caption{\textbf{AUPRO score for anomaly segmentation under \textit{Non-Overlap} setting of all categories in Eyecandies~\cite{eyecandies}.} Our method clearly outperforms other methods in 3D + RGB settings, indicating the superior anomaly detection ability of our method. We report the mean and standard deviation over 3 random seeds for each measurement. Optimal and sub-optimal results are in \textbf{bold} and {\ul underlined}, respectively.}
\label{tab:aupro_eye}
\setlength{\tabcolsep}{3.5pt}
\begin{tabular}{cc|cccccccccc|c}
\toprule
\textbf{}               & Method         & \begin{tabular}[c]{@{}c@{}}Candy\\ Cane\end{tabular} & \begin{tabular}[c]{@{}c@{}}Chocolate\\ Cookie\end{tabular} & \begin{tabular}[c]{@{}c@{}}Chocolate\\ Praline\end{tabular} & Confetto            & \begin{tabular}[c]{@{}c@{}}Gummy\\ Bear\end{tabular} & \begin{tabular}[c]{@{}c@{}}Hazelnut\\ Truffle\end{tabular} & \begin{tabular}[c]{@{}c@{}}Licorice\\ Sandwich\end{tabular} & Lollipop            & \begin{tabular}[c]{@{}c@{}}Marsh-\\ mallow\end{tabular}         & \begin{tabular}[c]{@{}c@{}}Peppermint\\ Candy\end{tabular} & Mean                \\ \midrule
\multirow{4}{*}{\rotatebox{90}{3D+RGB}} & PatchCore+FPFH & 83.5{\fontsize{6.5pt}{7pt}\selectfont$\pm$1.5}          & 89.9{\fontsize{6.5pt}{7pt}\selectfont$\pm$0.7}          & 67.0{\fontsize{6.5pt}{7pt}\selectfont$\pm$0.8}       & {\ul 96.4{\fontsize{6.5pt}{7pt}\selectfont$\pm$0.0}}    & 81.9{\fontsize{6.5pt}{7pt}\selectfont$\pm$0.8}       & 51.6{\fontsize{6.5pt}{7pt}\selectfont$\pm$1.2}       & 86.7{\fontsize{6.5pt}{7pt}\selectfont$\pm$0.6}       & 89.9{\fontsize{6.5pt}{7pt}\selectfont$\pm$0.3}          & 94.6{\fontsize{6.5pt}{7pt}\selectfont$\pm$0.6}          & 88.6{\fontsize{6.5pt}{7pt}\selectfont$\pm$0.7}          & 83.0{\fontsize{6.5pt}{7pt}\selectfont$\pm$0.3}          \\
                         & Shape-Guided   & 84.9{\fontsize{6.5pt}{7pt}\selectfont$\pm$0.5}          & {\ul 91.0{\fontsize{6.5pt}{7pt}\selectfont$\pm$0.1}}    & 69.8{\fontsize{6.5pt}{7pt}\selectfont$\pm$0.4}       & 95.5{\fontsize{6.5pt}{7pt}\selectfont$\pm$0.3}          & 84.6{\fontsize{6.5pt}{7pt}\selectfont$\pm$0.7}       & 61.1{\fontsize{6.5pt}{7pt}\selectfont$\pm$0.9}       & {\ul 90.5{\fontsize{6.5pt}{7pt}\selectfont$\pm$0.8}} & \textbf{95.1{\fontsize{6.5pt}{7pt}\selectfont$\pm$0.2}} & {\ul 96.4{\fontsize{6.5pt}{7pt}\selectfont$\pm$0.2}}    & 93.8{\fontsize{6.5pt}{7pt}\selectfont$\pm$0.3}          & 86.3{\fontsize{6.5pt}{7pt}\selectfont$\pm$0.2}          \\
                         & M3DM           & {\ul 88.0{\fontsize{6.5pt}{7pt}\selectfont$\pm$1.1}}    & 90.4{\fontsize{6.5pt}{7pt}\selectfont$\pm$1.2}          & {\ul 80.6{\fontsize{6.5pt}{7pt}\selectfont$\pm$0.2}} & 96.1{\fontsize{6.5pt}{7pt}\selectfont$\pm$3.6}          & {\ul 87.4{\fontsize{6.5pt}{7pt}\selectfont$\pm$1.2}} & {\ul 65.7{\fontsize{6.5pt}{7pt}\selectfont$\pm$1.3}} & 86.4{\fontsize{6.5pt}{7pt}\selectfont$\pm$1.4}       & 91.2{\fontsize{6.5pt}{7pt}\selectfont$\pm$0.2}          & 96.2{\fontsize{6.5pt}{7pt}\selectfont$\pm$0.6}          & \textbf{96.2{\fontsize{6.5pt}{7pt}\selectfont$\pm$0.8}} & {\ul 87.8{\fontsize{6.5pt}{7pt}\selectfont$\pm$0.3}}    \\
                         & Ours           & \textbf{89.8{\fontsize{6.5pt}{7pt}\selectfont$\pm$0.6}} & \textbf{91.6{\fontsize{6.5pt}{7pt}\selectfont$\pm$0.3}} & 77.6{\fontsize{6.5pt}{7pt}\selectfont$\pm$1.8}       & \textbf{98.1{\fontsize{6.5pt}{7pt}\selectfont$\pm$0.1}} & 86.6{\fontsize{6.5pt}{7pt}\selectfont$\pm$2.0}       & 65.2{\fontsize{6.5pt}{7pt}\selectfont$\pm$1.1}       & 85.8{\fontsize{6.5pt}{7pt}\selectfont$\pm$1.4}       & 90.8{\fontsize{6.5pt}{7pt}\selectfont$\pm$0.6}          & \textbf{96.9{\fontsize{6.5pt}{7pt}\selectfont$\pm$0.3}} & {\ul 96.1{\fontsize{6.5pt}{7pt}\selectfont$\pm$0.8}}    & \textbf{87.8{\fontsize{6.5pt}{7pt}\selectfont$\pm$0.2}}
                        \\ \bottomrule
\end{tabular}
\end{center}
\end{table*}

\begin{table*}[t]
\begin{center}
\caption{\textbf{P-AUROC score for anomaly segmentation under \textit{Overlap} setting of all categories in Eyecandies~\cite{eyecandies}.} Our method clearly outperforms other methods in 3D + RGB settings, indicating the superior anomaly detection ability of our method. We report the mean and standard deviation over 3 random seeds for each measurement. Optimal and sub-optimal results are in \textbf{bold} and {\ul underlined}, respectively.}
\label{tab:pauroc-overlap_eye}
\setlength{\tabcolsep}{3.5pt}
\begin{tabular}{cc|cccccccccc|c}
\toprule
\textbf{}               & Method         & \begin{tabular}[c]{@{}c@{}}Candy\\ Cane\end{tabular} & \begin{tabular}[c]{@{}c@{}}Chocolate\\ Cookie\end{tabular} & \begin{tabular}[c]{@{}c@{}}Chocolate\\ Praline\end{tabular} & Confetto            & \begin{tabular}[c]{@{}c@{}}Gummy\\ Bear\end{tabular} & \begin{tabular}[c]{@{}c@{}}Hazelnut\\ Truffle\end{tabular} & \begin{tabular}[c]{@{}c@{}}Licorice\\ Sandwich\end{tabular} & Lollipop            & \begin{tabular}[c]{@{}c@{}}Marsh-\\ mallow\end{tabular}         & \begin{tabular}[c]{@{}c@{}}Peppermint\\ Candy\end{tabular} & Mean                \\ \midrule
\multirow{5}{*}{\rotatebox{90}{3D+RGB}} & PatchCore+FPFH & 21.7{\fontsize{6.5pt}{7pt}\selectfont$\pm$2.3}          & 21.4{\fontsize{6.5pt}{7pt}\selectfont$\pm$2.5}          & 28.9{\fontsize{6.5pt}{7pt}\selectfont$\pm$3.0}          & 25.0{\fontsize{6.5pt}{7pt}\selectfont$\pm$2.0}          & 34.6{\fontsize{6.5pt}{7pt}\selectfont$\pm$5.5}          & 35.5{\fontsize{6.5pt}{7pt}\selectfont$\pm$3.7}          & 20.6{\fontsize{6.5pt}{7pt}\selectfont$\pm$2.0}          & 25.6{\fontsize{6.5pt}{7pt}\selectfont$\pm$21.0}         & 22.3{\fontsize{6.5pt}{7pt}\selectfont$\pm$3.3}          & 26.8{\fontsize{6.5pt}{7pt}\selectfont$\pm$7.9}          & 26.2{\fontsize{6.5pt}{7pt}\selectfont$\pm$1.1}          \\
                        & AST            & 48.3{\fontsize{6.5pt}{7pt}\selectfont$\pm$0.6}          & 49.3{\fontsize{6.5pt}{7pt}\selectfont$\pm$0.6}          & 48.3{\fontsize{6.5pt}{7pt}\selectfont$\pm$0.6}          & 48.6{\fontsize{6.5pt}{7pt}\selectfont$\pm$0.6}          & \textbf{78.1{\fontsize{6.5pt}{7pt}\selectfont$\pm$1.0}} & 49.0{\fontsize{6.5pt}{7pt}\selectfont$\pm$1.0}          & 76.1{\fontsize{6.5pt}{7pt}\selectfont$\pm$1.0}          & 48.7{\fontsize{6.5pt}{7pt}\selectfont$\pm$1.0}          & 77.0{\fontsize{6.5pt}{7pt}\selectfont$\pm$0.6}          & 49.0{\fontsize{6.5pt}{7pt}\selectfont$\pm$0.0}          & 57.2{\fontsize{6.5pt}{7pt}\selectfont$\pm$0.5 }         \\
                        & Shape-Guided   & \textbf{89.7{\fontsize{6.5pt}{7pt}\selectfont$\pm$0.4}} & {\ul 82.4{\fontsize{6.5pt}{7pt}\selectfont$\pm$0.8}}    & {\ul 71.6{\fontsize{6.5pt}{7pt}\selectfont$\pm$1.2}}    & {\ul 86.0{\fontsize{6.5pt}{7pt}\selectfont$\pm$1.5}}    & {\ul 78.1{\fontsize{6.5pt}{7pt}\selectfont$\pm$1.5}}    & {\ul 67.6{\fontsize{6.5pt}{7pt}\selectfont$\pm$2.4}}    & {\ul 78.4{\fontsize{6.5pt}{7pt}\selectfont$\pm$0.7}}    & \textbf{94.1{\fontsize{6.5pt}{7pt}\selectfont$\pm$2.0}} & {\ul 81.0{\fontsize{6.5pt}{7pt}\selectfont$\pm$0.6}}    & \textbf{65.5{\fontsize{6.5pt}{7pt}\selectfont$\pm$3.1}} & {\ul 79.5{\fontsize{6.5pt}{7pt}\selectfont$\pm$1.0}}    \\
                        & M3DM           & 37.5{\fontsize{6.5pt}{7pt}\selectfont$\pm$2.6}          & 24.2{\fontsize{6.5pt}{7pt}\selectfont$\pm$1.8}          & 30.2{\fontsize{6.5pt}{7pt}\selectfont$\pm$3.9}          & 22.7{\fontsize{6.5pt}{7pt}\selectfont$\pm$2.1}          & 34.8{\fontsize{6.5pt}{7pt}\selectfont$\pm$4.9}          & 39.7{\fontsize{6.5pt}{7pt}\selectfont$\pm$3.0}          & 21.6{\fontsize{6.5pt}{7pt}\selectfont$\pm$2.6}          & 26.5{\fontsize{6.5pt}{7pt}\selectfont$\pm$21.1}         & 19.6{\fontsize{6.5pt}{7pt}\selectfont$\pm$3.6}          & 19.0{\fontsize{6.5pt}{7pt}\selectfont$\pm$1.3}          & 27.6{\fontsize{6.5pt}{7pt}\selectfont$\pm$1.2 }         \\
                        & Ours           & {\ul 57.0{\fontsize{6.5pt}{7pt}\selectfont$\pm$2.5}}    & \textbf{87.4{\fontsize{6.5pt}{7pt}\selectfont$\pm$6.0}} & \textbf{78.0{\fontsize{6.5pt}{7pt}\selectfont$\pm$2.4}} & \textbf{91.6{\fontsize{6.5pt}{7pt}\selectfont$\pm$3.9}} & 70.7{\fontsize{6.5pt}{7pt}\selectfont$\pm$3.3}          & \textbf{82.0{\fontsize{6.5pt}{7pt}\selectfont$\pm$4.0}} & \textbf{90.2{\fontsize{6.5pt}{7pt}\selectfont$\pm$2.3}} & {\ul 81.8{\fontsize{6.5pt}{7pt}\selectfont$\pm$6.4}}    & \textbf{98.5{\fontsize{6.5pt}{7pt}\selectfont$\pm$1.2}} & {\ul 60.3{\fontsize{6.5pt}{7pt}\selectfont$\pm$8.3}}    & \textbf{79.8{\fontsize{6.5pt}{7pt}\selectfont$\pm$0.7}}
                        \\ \bottomrule
\end{tabular}
\end{center}
\end{table*}

\begin{table*}[t]
\begin{center}
\caption{\textbf{P-AUROC score for anomaly segmentation under \textit{Non-Overlap} setting of all categories in Eyecandies~\cite{eyecandies}.} Our method clearly outperforms other methods in 3D + RGB settings, indicating the superior anomaly detection ability of our method. We report the mean and standard deviation over 3 random seeds for each measurement. Optimal and sub-optimal results are in \textbf{bold} and {\ul underlined}, respectively.}
\label{tab:pauroc_eye}
\setlength{\tabcolsep}{3.5pt}
\begin{tabular}{cc|cccccccccc|c}
\toprule
\textbf{}               & Method         & \begin{tabular}[c]{@{}c@{}}Candy\\ Cane\end{tabular} & \begin{tabular}[c]{@{}c@{}}Chocolate\\ Cookie\end{tabular} & \begin{tabular}[c]{@{}c@{}}Chocolate\\ Praline\end{tabular} & Confetto            & \begin{tabular}[c]{@{}c@{}}Gummy\\ Bear\end{tabular} & \begin{tabular}[c]{@{}c@{}}Hazelnut\\ Truffle\end{tabular} & \begin{tabular}[c]{@{}c@{}}Licorice\\ Sandwich\end{tabular} & Lollipop            & \begin{tabular}[c]{@{}c@{}}Marsh-\\ mallow\end{tabular}         & \begin{tabular}[c]{@{}c@{}}Peppermint\\ Candy\end{tabular} & Mean                \\ \midrule
\multirow{5}{*}{\rotatebox{90}{3D+RGB}} & PatchCore+FPFH & 95.7{\fontsize{6.5pt}{7pt}\selectfont$\pm$0.2}          & 97.4{\fontsize{6.5pt}{7pt}\selectfont$\pm$0.1}          & 91.7{\fontsize{6.5pt}{7pt}\selectfont$\pm$0.3}          & 99.4{\fontsize{6.5pt}{7pt}\selectfont$\pm$0.0}          & 92.9{\fontsize{6.5pt}{7pt}\selectfont$\pm$0.2}          & 87.4{\fontsize{6.5pt}{7pt}\selectfont$\pm$0.5}          & 96.9{\fontsize{6.5pt}{7pt}\selectfont$\pm$0.2}          & 98.1{\fontsize{6.5pt}{7pt}\selectfont$\pm$0.2}          & 99.2{\fontsize{6.5pt}{7pt}\selectfont$\pm$0.1}          & 97.3{\fontsize{6.5pt}{7pt}\selectfont$\pm$0.2}          & 95.6{\fontsize{6.5pt}{7pt}\selectfont$\pm$0.1 }         \\
                        & AST            & 95.1{\fontsize{6.5pt}{7pt}\selectfont$\pm$0.6}          & 98.3{\fontsize{6.5pt}{7pt}\selectfont$\pm$1.0}          & 91.4{\fontsize{6.5pt}{7pt}\selectfont$\pm$0.6}          & 99.3{\fontsize{6.5pt}{7pt}\selectfont$\pm$0.6}          & 92.0{\fontsize{6.5pt}{7pt}\selectfont$\pm$0.6}          & 88.2{\fontsize{6.5pt}{7pt}\selectfont$\pm$0.6}          & 96.0{\fontsize{6.5pt}{7pt}\selectfont$\pm$0.6}          & 95.9{\fontsize{6.5pt}{7pt}\selectfont$\pm$0.6}          & 98.8{\fontsize{6.5pt}{7pt}\selectfont$\pm$0.6}          & 97.0{\fontsize{6.5pt}{7pt}\selectfont$\pm$0.6}          & 95.2{\fontsize{6.5pt}{7pt}\selectfont$\pm$0.2}          \\
                        & Shape-Guided   & 95.8{\fontsize{6.5pt}{7pt}\selectfont$\pm$0.1}          & 98.3{\fontsize{6.5pt}{7pt}\selectfont$\pm$0.0}          & 92.7{\fontsize{6.5pt}{7pt}\selectfont$\pm$0.0}          & 99.0{\fontsize{6.5pt}{7pt}\selectfont$\pm$0.1}          & 91.9{\fontsize{6.5pt}{7pt}\selectfont$\pm$0.3}          & 89.0{\fontsize{6.5pt}{7pt}\selectfont$\pm$0.2}          & \textbf{97.9{\fontsize{6.5pt}{7pt}\selectfont$\pm$0.2}} & 98.5{\fontsize{6.5pt}{7pt}\selectfont$\pm$0.1}          & 99.5{\fontsize{6.5pt}{7pt}\selectfont$\pm$0.1}          & 98.4{\fontsize{6.5pt}{7pt}\selectfont$\pm$0.1}          & 96.1{\fontsize{6.5pt}{7pt}\selectfont$\pm$0.1}          \\
                        & M3DM           & {\ul 96.4{\fontsize{6.5pt}{7pt}\selectfont$\pm$0.3}}    & {\ul 98.3{\fontsize{6.5pt}{7pt}\selectfont$\pm$0.3}}    & {\ul 95.2{\fontsize{6.5pt}{7pt}\selectfont$\pm$1.9}}    & \textbf{99.8{\fontsize{6.5pt}{7pt}\selectfont$\pm$0.0}} & \textbf{97.5{\fontsize{6.5pt}{7pt}\selectfont$\pm$0.3}} & \textbf{93.3{\fontsize{6.5pt}{7pt}\selectfont$\pm$0.2}} & 95.5{\fontsize{6.5pt}{7pt}\selectfont$\pm$3.1}          & \textbf{98.9{\fontsize{6.5pt}{7pt}\selectfont$\pm$0.0}} & {\ul 99.6{\fontsize{6.5pt}{7pt}\selectfont$\pm$0.1}}    & \textbf{99.4{\fontsize{6.5pt}{7pt}\selectfont$\pm$0.1}} & {\ul 97.4{\fontsize{6.5pt}{7pt}\selectfont$\pm$0.5}}    \\
                        & Ours           & \textbf{96.9{\fontsize{6.5pt}{7pt}\selectfont$\pm$0.3}} & \textbf{98.4{\fontsize{6.5pt}{7pt}\selectfont$\pm$0.0}} & \textbf{95.5{\fontsize{6.5pt}{7pt}\selectfont$\pm$0.7}} & {\ul 99.8{\fontsize{6.5pt}{7pt}\selectfont$\pm$0.1}}    & {\ul 96.7{\fontsize{6.5pt}{7pt}\selectfont$\pm$0.5}}    & {\ul 92.8{\fontsize{6.5pt}{7pt}\selectfont$\pm$0.7}}    & {\ul 97.1{\fontsize{6.5pt}{7pt}\selectfont$\pm$0.2}}    & {\ul 98.7{\fontsize{6.5pt}{7pt}\selectfont$\pm$0.1}}    & \textbf{99.7{\fontsize{6.5pt}{7pt}\selectfont$\pm$0.0}} & {\ul 99.3{\fontsize{6.5pt}{7pt}\selectfont$\pm$0.3}}    & \textbf{97.5{\fontsize{6.5pt}{7pt}\selectfont$\pm$0.0}}
                        \\ \bottomrule
\end{tabular}
\end{center}
\end{table*}

\section*{Experiments on different noise level}
\label{app.different_noise}
To further validate the robustness of our method against noise in the training dataset, we conducted experiments by injecting different percentages of noise into the training set. Specifically, we performed experiments with 20\% and 30\% noise data injected into the training dataset. The results of these experiments are presented in the \cref{tab:iauroc_overlap_noise,tab:iauroc_noise,tab:aupro_overlap_noise,tab:aupro_noise,tab:pauroc_overlap_noise,tab:pauroc_noise} below.
Comparing the results of injecting 10\% noise, 20\% noise and 30\% noise, we can conclude that our method is much more robust to noise in the training dataset than previous methods.
\begin{table*}[t]
\begin{center}
\caption{\textbf{I-AUROC score for anomaly detection under \textit{Overlap} setting of all categories in MVTec 3D-AD.} We inject 20\% and 30\% noise into the training dataset. Our method outperforms other methods, indicating the superior anomaly detection ability of our method. We report the mean and standard deviation over 3 random seeds for each measurement. Optimal and sub-optimal results are in \textbf{bold} and {\ul underlined}, respectively.}
\label{tab:iauroc_noise}
\setlength{\tabcolsep}{4.7pt}
\begin{tabular}{cc|cccccccccc|c}
\toprule
\textbf{}               & Method         & Bagel               & \begin{tabular}[c]{@{}c@{}}Cable\\ Gland\end{tabular}         & Carrot              & Cookie              & Dowel               & Foam                & Peach               & Potato              & Rope                & Tire       & Mean                \\ \midrule
\multirow{5}{*}{\rotatebox{90}{Noise 20\%}} & PatchCore+FPFH & 42.0{\fontsize{6.5pt}{7pt}\selectfont$\pm$1.6}          & 40.8{\fontsize{6.5pt}{7pt}\selectfont$\pm$2.8}          & 49.5{\fontsize{6.5pt}{7pt}\selectfont$\pm$0.3}          & 53.0{\fontsize{6.5pt}{7pt}\selectfont$\pm$0.6}          & 44.1{\fontsize{6.5pt}{7pt}\selectfont$\pm$1.3}          & 28.2{\fontsize{6.5pt}{7pt}\selectfont$\pm$2.1}          & 27.3{\fontsize{6.5pt}{7pt}\selectfont$\pm$1.2}          & 25.9{\fontsize{6.5pt}{7pt}\selectfont$\pm$1.5}          & 13.2{\fontsize{6.5pt}{7pt}\selectfont$\pm$1.3}          & 45.1{\fontsize{6.5pt}{7pt}\selectfont$\pm$2.0}          & 36.9{\fontsize{6.5pt}{7pt}\selectfont$\pm$0.3}          \\
                        & AST            & 37.3{\fontsize{6.5pt}{7pt}\selectfont$\pm$1.0}          & 44.8{\fontsize{6.5pt}{7pt}\selectfont$\pm$0.6}          & 50.3{\fontsize{6.5pt}{7pt}\selectfont$\pm$0.6}          & {\ul 59.5{\fontsize{6.5pt}{7pt}\selectfont$\pm$0.0}}    & 43.2{\fontsize{6.5pt}{7pt}\selectfont$\pm$0.6}          & 33.2{\fontsize{6.5pt}{7pt}\selectfont$\pm$0.6}          & 29.4{\fontsize{6.5pt}{7pt}\selectfont$\pm$1.0}          & {\ul 31.5{\fontsize{6.5pt}{7pt}\selectfont$\pm$1.0}}    & 12.4{\fontsize{6.5pt}{7pt}\selectfont$\pm$0.6}          & 38.1{\fontsize{6.5pt}{7pt}\selectfont$\pm$0.6}          & 38.0{\fontsize{6.5pt}{7pt}\selectfont$\pm$0.1}          \\
                        & Shape-Guided   & 42.3{\fontsize{6.5pt}{7pt}\selectfont$\pm$1.1}          & 45.1{\fontsize{6.5pt}{7pt}\selectfont$\pm$1.6}          & {\ul 53.2{\fontsize{6.5pt}{7pt}\selectfont$\pm$0.3}}    & 50.6{\fontsize{6.5pt}{7pt}\selectfont$\pm$0.5}          & 44.6{\fontsize{6.5pt}{7pt}\selectfont$\pm$1.3}          & 32.8{\fontsize{6.5pt}{7pt}\selectfont$\pm$0.7}          & 29.4{\fontsize{6.5pt}{7pt}\selectfont$\pm$0.1}          & 30.1{\fontsize{6.5pt}{7pt}\selectfont$\pm$0.5}          & 14.0{\fontsize{6.5pt}{7pt}\selectfont$\pm$0.7}          & 45.9{\fontsize{6.5pt}{7pt}\selectfont$\pm$1.3}          & 38.8{\fontsize{6.5pt}{7pt}\selectfont$\pm$0.3}          \\
                        & M3DM           & {\ul 45.0{\fontsize{6.5pt}{7pt}\selectfont$\pm$1.1}}    & {\ul 47.3{\fontsize{6.5pt}{7pt}\selectfont$\pm$1.0}}    & 47.6{\fontsize{6.5pt}{7pt}\selectfont$\pm$1.0}          & 56.8{\fontsize{6.5pt}{7pt}\selectfont$\pm$1.9}          & {\ul 51.4{\fontsize{6.5pt}{7pt}\selectfont$\pm$1.0}}    & {\ul 41.3{\fontsize{6.5pt}{7pt}\selectfont$\pm$0.5}}    & {\ul 32.7{\fontsize{6.5pt}{7pt}\selectfont$\pm$0.7}}    & 27.9{\fontsize{6.5pt}{7pt}\selectfont$\pm$1.5}          & {\ul 25.5{\fontsize{6.5pt}{7pt}\selectfont$\pm$1.4}}    & {\ul 53.8{\fontsize{6.5pt}{7pt}\selectfont$\pm$1.2}}    & {\ul 42.9{\fontsize{6.5pt}{7pt}\selectfont$\pm$0.5}}    \\
                        & Ours           & \textbf{92.8{\fontsize{6.5pt}{7pt}\selectfont$\pm$1.5}} & \textbf{76.4{\fontsize{6.5pt}{7pt}\selectfont$\pm$1.8}} & \textbf{93.0{\fontsize{6.5pt}{7pt}\selectfont$\pm$0.5}} & \textbf{85.7{\fontsize{6.5pt}{7pt}\selectfont$\pm$0.9}} & \textbf{82.4{\fontsize{6.5pt}{7pt}\selectfont$\pm$0.7}} & \textbf{71.4{\fontsize{6.5pt}{7pt}\selectfont$\pm$5.2}} & \textbf{67.7{\fontsize{6.5pt}{7pt}\selectfont$\pm$5.0}} & \textbf{60.2{\fontsize{6.5pt}{7pt}\selectfont$\pm$2.9}} & \textbf{90.2{\fontsize{6.5pt}{7pt}\selectfont$\pm$1.5}} & \textbf{73.3{\fontsize{6.5pt}{7pt}\selectfont$\pm$2.3}} & \textbf{79.3{\fontsize{6.5pt}{7pt}\selectfont$\pm$1.0}} \\ \midrule
\multirow{5}{*}{\rotatebox{90}{Noise 30\%}} & PatchCore+FPFH & 18.6{\fontsize{6.5pt}{7pt}\selectfont$\pm$1.5}          & 22.2{\fontsize{6.5pt}{7pt}\selectfont$\pm$1.8}          & 30.8{\fontsize{6.5pt}{7pt}\selectfont$\pm$0.8}          & 39.7{\fontsize{6.5pt}{7pt}\selectfont$\pm$3.4}          & 18.2{\fontsize{6.5pt}{7pt}\selectfont$\pm$1.2}          & 13.4{\fontsize{6.5pt}{7pt}\selectfont$\pm$2.0}          & 4.2{\fontsize{6.5pt}{7pt}\selectfont$\pm$0.4}           & 4.1{\fontsize{6.5pt}{7pt}\selectfont$\pm$0.4}           & 7.0{\fontsize{6.5pt}{7pt}\selectfont$\pm$0.3}           & 24.9{\fontsize{6.5pt}{7pt}\selectfont$\pm$1.3}          & 18.3{\fontsize{6.5pt}{7pt}\selectfont$\pm$0.7}           \\
                        & AST            & 14.6{\fontsize{6.5pt}{7pt}\selectfont$\pm$0.6}          & 21.4{\fontsize{6.5pt}{7pt}\selectfont$\pm$1.0}          & 28.7{\fontsize{6.5pt}{7pt}\selectfont$\pm$0.6}          & 38.4{\fontsize{6.5pt}{7pt}\selectfont$\pm$0.0}          & 16.4{\fontsize{6.5pt}{7pt}\selectfont$\pm$0.0}          & 9.3{\fontsize{6.5pt}{7pt}\selectfont$\pm$1.0}           & 4.3{\fontsize{6.5pt}{7pt}\selectfont$\pm$0.6}           & 5.6{\fontsize{6.5pt}{7pt}\selectfont$\pm$0.6}           & 6.8{\fontsize{6.5pt}{7pt}\selectfont$\pm$0.0}           & 20.2{\fontsize{6.5pt}{7pt}\selectfont$\pm$1.0}          & 16.6{\fontsize{6.5pt}{7pt}\selectfont$\pm$0.1}           \\
                        & Shape-Guided   & 15.7{\fontsize{6.5pt}{7pt}\selectfont$\pm$0.6}          & 22.3{\fontsize{6.5pt}{7pt}\selectfont$\pm$1.2}          & {\ul 32.8{\fontsize{6.5pt}{7pt}\selectfont$\pm$1.0}}    & 31.3{\fontsize{6.5pt}{7pt}\selectfont$\pm$0.2}          & 18.3{\fontsize{6.5pt}{7pt}\selectfont$\pm$0.3}          & 9.7{\fontsize{6.5pt}{7pt}\selectfont$\pm$0.9}           & 4.2{\fontsize{6.5pt}{7pt}\selectfont$\pm$0.1}           & 4.7{\fontsize{6.5pt}{7pt}\selectfont$\pm$0.8}           & 7.2{\fontsize{6.5pt}{7pt}\selectfont$\pm$0.1}           & 24.7{\fontsize{6.5pt}{7pt}\selectfont$\pm$1.5}          & 17.1{\fontsize{6.5pt}{7pt}\selectfont$\pm$0.3}           \\
                        & M3DM           & {\ul 30.4{\fontsize{6.5pt}{7pt}\selectfont$\pm$1.6}}    & {\ul 27.4{\fontsize{6.5pt}{7pt}\selectfont$\pm$1.9}}    & 32.5{\fontsize{6.5pt}{7pt}\selectfont$\pm$0.8}          & {\ul 40.7{\fontsize{6.5pt}{7pt}\selectfont$\pm$1.4}}    & {\ul 36.7{\fontsize{6.5pt}{7pt}\selectfont$\pm$2.4}}    & {\ul 25.5{\fontsize{6.5pt}{7pt}\selectfont$\pm$3.1}}    & {\ul 16.0{\fontsize{6.5pt}{7pt}\selectfont$\pm$1.4}}    & {\ul 12.2{\fontsize{6.5pt}{7pt}\selectfont$\pm$1.2}}    & {\ul 19.9{\fontsize{6.5pt}{7pt}\selectfont$\pm$2.0}}    & {\ul 37.9{\fontsize{6.5pt}{7pt}\selectfont$\pm$1.3}}    & {\ul 27.9{\fontsize{6.5pt}{7pt}\selectfont$\pm$0.8}}     \\
                        & Ours           & \textbf{89.7{\fontsize{6.5pt}{7pt}\selectfont$\pm$1.3}} & \textbf{69.1{\fontsize{6.5pt}{7pt}\selectfont$\pm$1.8}} & \textbf{93.7{\fontsize{6.5pt}{7pt}\selectfont$\pm$0.7}} & \textbf{83.7{\fontsize{6.5pt}{7pt}\selectfont$\pm$2.0}} & \textbf{78.8{\fontsize{6.5pt}{7pt}\selectfont$\pm$2.1}} & \textbf{69.9{\fontsize{6.5pt}{7pt}\selectfont$\pm$4.9}} & \textbf{67.1{\fontsize{6.5pt}{7pt}\selectfont$\pm$3.4}} & \textbf{55.3{\fontsize{6.5pt}{7pt}\selectfont$\pm$2.0}} & \textbf{90.5{\fontsize{6.5pt}{7pt}\selectfont$\pm$0.9}} & \textbf{70.0{\fontsize{6.5pt}{7pt}\selectfont$\pm$2.1}} & \textbf{76.8{\fontsize{6.5pt}{7pt}\selectfont$\pm$0.6}}
                        \\ \bottomrule
\end{tabular}
\end{center}
\end{table*}

\begin{table*}[t]
\begin{center}
\caption{\textbf{I-AUROC score for anomaly detection under \textit{Non-Overlap} setting of all categories in MVTec 3D-AD.} We inject 20\% and 30\% noise into the training dataset. Our method outperforms other methods, indicating the superior anomaly detection ability of our method. We report the mean and standard deviation over 3 random seeds for each measurement. Optimal and sub-optimal results are in \textbf{bold} and {\ul underlined}, respectively.}
\label{tab:iauroc_overlap_noise}
\setlength{\tabcolsep}{4.7pt}
\begin{tabular}{cc|cccccccccc|c}
\toprule
\textbf{}               & Method         & Bagel               & \begin{tabular}[c]{@{}c@{}}Cable\\ Gland\end{tabular}         & Carrot              & Cookie              & Dowel               & Foam                & Peach               & Potato              & Rope                & Tire       & Mean                \\ \midrule
\multirow{5}{*}{\rotatebox{90}{Noise 20\%}} & PatchCore+FPFH & 84.0{\fontsize{6.5pt}{7pt}\selectfont$\pm$1.4}          & 84.0{\fontsize{6.5pt}{7pt}\selectfont$\pm$0.8}          & 87.5{\fontsize{6.5pt}{7pt}\selectfont$\pm$0.1}          & 79.5{\fontsize{6.5pt}{7pt}\selectfont$\pm$2.5}          & 93.0{\fontsize{6.5pt}{7pt}\selectfont$\pm$0.5}          & 56.9{\fontsize{6.5pt}{7pt}\selectfont$\pm$3.6}          & 82.6{\fontsize{6.5pt}{7pt}\selectfont$\pm$3.7}          & 73.0{\fontsize{6.5pt}{7pt}\selectfont$\pm$4.8}          & 90.3{\fontsize{6.5pt}{7pt}\selectfont$\pm$8.1}          & {\ul 84.8{\fontsize{6.5pt}{7pt}\selectfont$\pm$3.3}}    & 81.6{\fontsize{6.5pt}{7pt}\selectfont$\pm$0.1}          \\
                        & AST            & 82.1{\fontsize{6.5pt}{7pt}\selectfont$\pm$1.0}          & \textbf{91.6{\fontsize{6.5pt}{7pt}\selectfont$\pm$0.6}} & {\ul 87.6{\fontsize{6.5pt}{7pt}\selectfont$\pm$0.6}}    & \textbf{92.8{\fontsize{6.5pt}{7pt}\selectfont$\pm$1.0}} & 93.4{\fontsize{6.5pt}{7pt}\selectfont$\pm$0.6}          & {\ul 79.7{\fontsize{6.5pt}{7pt}\selectfont$\pm$1.0}}    & \textbf{91.1{\fontsize{6.5pt}{7pt}\selectfont$\pm$0.6}} & \textbf{90.1{\fontsize{6.5pt}{7pt}\selectfont$\pm$0.6}} & 88.1{\fontsize{6.5pt}{7pt}\selectfont$\pm$0.6}          & 72.3{\fontsize{6.5pt}{7pt}\selectfont$\pm$0.6}          & {\ul 86.9{\fontsize{6.5pt}{7pt}\selectfont$\pm$0.3}}    \\
                        & Shape-Guided   & 82.8{\fontsize{6.5pt}{7pt}\selectfont$\pm$2.3}          & 81.8{\fontsize{6.5pt}{7pt}\selectfont$\pm$2.9}          & 86.6{\fontsize{6.5pt}{7pt}\selectfont$\pm$0.5}          & 79.0{\fontsize{6.5pt}{7pt}\selectfont$\pm$0.9}          & 86.2{\fontsize{6.5pt}{7pt}\selectfont$\pm$1.3}          & 69.1{\fontsize{6.5pt}{7pt}\selectfont$\pm$1.5}          & 74.1{\fontsize{6.5pt}{7pt}\selectfont$\pm$0.3}          & 72.8{\fontsize{6.5pt}{7pt}\selectfont$\pm$1.2}          & 60.3{\fontsize{6.5pt}{7pt}\selectfont$\pm$3.0}          & 79.8{\fontsize{6.5pt}{7pt}\selectfont$\pm$2.3}          & 77.3{\fontsize{6.5pt}{7pt}\selectfont$\pm$0.7}          \\
                        & M3DM           & {\ul 92.6{\fontsize{6.5pt}{7pt}\selectfont$\pm$3.4}}    & 76.8{\fontsize{6.5pt}{7pt}\selectfont$\pm$2.1}          & 82.6{\fontsize{6.5pt}{7pt}\selectfont$\pm$1.2}          & 82.4{\fontsize{6.5pt}{7pt}\selectfont$\pm$3.1}          & \textbf{95.2{\fontsize{6.5pt}{7pt}\selectfont$\pm$0.8}} & 75.3{\fontsize{6.5pt}{7pt}\selectfont$\pm$0.6}          & 83.0{\fontsize{6.5pt}{7pt}\selectfont$\pm$4.1}          & 74.1{\fontsize{6.5pt}{7pt}\selectfont$\pm$4.2}          & {\ul 98.0{\fontsize{6.5pt}{7pt}\selectfont$\pm$2.4}}    & 84.3{\fontsize{6.5pt}{7pt}\selectfont$\pm$2.1}          & 84.4{\fontsize{6.5pt}{7pt}\selectfont$\pm$1.0}          \\
                        & Ours           & \textbf{97.4{\fontsize{6.5pt}{7pt}\selectfont$\pm$0.3}} & {\ul 85.0{\fontsize{6.5pt}{7pt}\selectfont$\pm$4.2}}    & \textbf{95.1{\fontsize{6.5pt}{7pt}\selectfont$\pm$0.3}} & {\ul 90.6{\fontsize{6.5pt}{7pt}\selectfont$\pm$0.9}}    & {\ul 94.0{\fontsize{6.5pt}{7pt}\selectfont$\pm$1.9}}    & \textbf{88.1{\fontsize{6.5pt}{7pt}\selectfont$\pm$1.9}} & {\ul 87.4{\fontsize{6.5pt}{7pt}\selectfont$\pm$1.4}}    & {\ul 79.8{\fontsize{6.5pt}{7pt}\selectfont$\pm$2.4}}    & \textbf{98.1{\fontsize{6.5pt}{7pt}\selectfont$\pm$1.0}} & \textbf{85.5{\fontsize{6.5pt}{7pt}\selectfont$\pm$0.9}} & \textbf{90.1{\fontsize{6.5pt}{7pt}\selectfont$\pm$0.7}} \\ \midrule
\multirow{5}{*}{\rotatebox{90}{Noise 30\%}} & PatchCore+FPFH & 78.2{\fontsize{6.5pt}{7pt}\selectfont$\pm$2.3}          & 81.5{\fontsize{6.5pt}{7pt}\selectfont$\pm$2.9}          & {\ul 86.5{\fontsize{6.5pt}{7pt}\selectfont$\pm$2.4}}    & 80.7{\fontsize{6.5pt}{7pt}\selectfont$\pm$3.6}          & \textbf{95.4{\fontsize{6.5pt}{7pt}\selectfont$\pm$2.7}} & 62.0{\fontsize{6.5pt}{7pt}\selectfont$\pm$5.8}          & 74.1{\fontsize{6.5pt}{7pt}\selectfont$\pm$3.6}          & 74.6{\fontsize{6.5pt}{7pt}\selectfont$\pm$6.8}          & {\ul 96.7{\fontsize{6.5pt}{7pt}\selectfont$\pm$3.2}}    & \textbf{88.5{\fontsize{6.5pt}{7pt}\selectfont$\pm$4.5}} & 81.8{\fontsize{6.5pt}{7pt}\selectfont$\pm$1.5}           \\
                        & AST            & 73.4{\fontsize{6.5pt}{7pt}\selectfont$\pm$0.6}          & \textbf{88.8{\fontsize{6.5pt}{7pt}\selectfont$\pm$0.6}} & 81.8{\fontsize{6.5pt}{7pt}\selectfont$\pm$0.6}          & \textbf{96.6{\fontsize{6.5pt}{7pt}\selectfont$\pm$0.6}} & {\ul 94.4{\fontsize{6.5pt}{7pt}\selectfont$\pm$1.0}}    & 74.0{\fontsize{6.5pt}{7pt}\selectfont$\pm$0.0}          & \textbf{96.6{\fontsize{6.5pt}{7pt}\selectfont$\pm$0.6}} & \textbf{94.4{\fontsize{6.5pt}{7pt}\selectfont$\pm$1.0}} & 73.7{\fontsize{6.5pt}{7pt}\selectfont$\pm$0.6}          & 85.3{\fontsize{6.5pt}{7pt}\selectfont$\pm$0.6}               & {\ul 85.7{\fontsize{6.5pt}{7pt}\selectfont$\pm$0.8}}     \\
                        & Shape-Guided   & 60.2{\fontsize{6.5pt}{7pt}\selectfont$\pm$2.2}          & 69.2{\fontsize{6.5pt}{7pt}\selectfont$\pm$3.7}          & 77.3{\fontsize{6.5pt}{7pt}\selectfont$\pm$2.3}          & 68.5{\fontsize{6.5pt}{7pt}\selectfont$\pm$0.5}          & 65.9{\fontsize{6.5pt}{7pt}\selectfont$\pm$1.1}          & 45.5{\fontsize{6.5pt}{7pt}\selectfont$\pm$4.1}          & 28.0{\fontsize{6.5pt}{7pt}\selectfont$\pm$0.4}          & 31.0{\fontsize{6.5pt}{7pt}\selectfont$\pm$5.1}          & 41.4{\fontsize{6.5pt}{7pt}\selectfont$\pm$0.5}          & 69.2{\fontsize{6.5pt}{7pt}\selectfont$\pm$4.0}          & 55.6{\fontsize{6.5pt}{7pt}\selectfont$\pm$0.9}           \\
                        & M3DM           & {\ul 90.6{\fontsize{6.5pt}{7pt}\selectfont$\pm$3.5}}    & {\ul 85.7{\fontsize{6.5pt}{7pt}\selectfont$\pm$7.6}}    & 78.5{\fontsize{6.5pt}{7pt}\selectfont$\pm$2.1}          & 82.4{\fontsize{6.5pt}{7pt}\selectfont$\pm$0.9}          & 93.2{\fontsize{6.5pt}{7pt}\selectfont$\pm$0.9}          & \textbf{84.8{\fontsize{6.5pt}{7pt}\selectfont$\pm$3.8}} & 87.2{\fontsize{6.5pt}{7pt}\selectfont$\pm$2.3}          & 71.5{\fontsize{6.5pt}{7pt}\selectfont$\pm$21.3}         & 95.8{\fontsize{6.5pt}{7pt}\selectfont$\pm$4.1}          & {\ul 85.1{\fontsize{6.5pt}{7pt}\selectfont$\pm$5.3}}    & 85.5{\fontsize{6.5pt}{7pt}\selectfont$\pm$3.1}           \\
                        & Ours           & \textbf{97.9{\fontsize{6.5pt}{7pt}\selectfont$\pm$0.9}} & 80.7{\fontsize{6.5pt}{7pt}\selectfont$\pm$6.2}          & \textbf{95.6{\fontsize{6.5pt}{7pt}\selectfont$\pm$1.0}} & {\ul 89.7{\fontsize{6.5pt}{7pt}\selectfont$\pm$1.4}}    & 94.1{\fontsize{6.5pt}{7pt}\selectfont$\pm$2.0}          & {\ul 83.8{\fontsize{6.5pt}{7pt}\selectfont$\pm$1.5}}    & {\ul 90.2{\fontsize{6.5pt}{7pt}\selectfont$\pm$3.5}}    & {\ul 78.5{\fontsize{6.5pt}{7pt}\selectfont$\pm$4.7}}    & \textbf{98.6{\fontsize{6.5pt}{7pt}\selectfont$\pm$1.0}} & 83.8{\fontsize{6.5pt}{7pt}\selectfont$\pm$6.8}          & \textbf{89.3{\fontsize{6.5pt}{7pt}\selectfont$\pm$0.9}}
                        \\ \bottomrule
\end{tabular}
\end{center}
\end{table*}

\begin{table*}[t]
\begin{center}
\caption{\textbf{AUPRO score for anomaly segmentation under \textit{Overlap} setting of all categories in MVTec 3D-AD.} We inject 20\% and 30\% noise into the training dataset. Our method outperforms other methods, indicating the superior anomaly detection ability of our method. We report the mean and standard deviation over 3 random seeds for each measurement. Optimal and sub-optimal results are in \textbf{bold} and {\ul underlined}, respectively.}
\label{tab:aupro_overlap_noise}
\setlength{\tabcolsep}{4.7pt}
\begin{tabular}{cc|cccccccccc|c}
\toprule
\textbf{}               & Method         & Bagel               & \begin{tabular}[c]{@{}c@{}}Cable\\ Gland\end{tabular}         & Carrot              & Cookie              & Dowel               & Foam                & Peach               & Potato              & Rope                & Tire       & Mean                \\ \midrule
\multirow{4}{*}{\rotatebox{90}{Noise 20\%}} & PatchCore+FPFH & 46.3{\fontsize{6.5pt}{7pt}\selectfont$\pm$1.6}          & 48.9{\fontsize{6.5pt}{7pt}\selectfont$\pm$1.1}          & 56.0{\fontsize{6.5pt}{7pt}\selectfont$\pm$0.3}          & 58.5{\fontsize{6.5pt}{7pt}\selectfont$\pm$1.3}          & 43.3{\fontsize{6.5pt}{7pt}\selectfont$\pm$1.0}          & 35.3{\fontsize{6.5pt}{7pt}\selectfont$\pm$1.4}          & 31.9{\fontsize{6.5pt}{7pt}\selectfont$\pm$0.5}          & 35.3{\fontsize{6.5pt}{7pt}\selectfont$\pm$2.8}          & 13.7{\fontsize{6.5pt}{7pt}\selectfont$\pm$0.7}          & 48.2{\fontsize{6.5pt}{7pt}\selectfont$\pm$0.6}          & 41.7{\fontsize{6.5pt}{7pt}\selectfont$\pm$0.2}          \\
                        & Shape-Guided   & {\ul 68.5{\fontsize{6.5pt}{7pt}\selectfont$\pm$0.6}}    & {\ul 69.2{\fontsize{6.5pt}{7pt}\selectfont$\pm$1.3}}    & {\ul 90.0{\fontsize{6.5pt}{7pt}\selectfont$\pm$1.2}}    & {\ul 64.4{\fontsize{6.5pt}{7pt}\selectfont$\pm$1.3}}    & \textbf{85.1{\fontsize{6.5pt}{7pt}\selectfont$\pm$0.9}} & {\ul 60.5{\fontsize{6.5pt}{7pt}\selectfont$\pm$1.6}}    & \textbf{82.7{\fontsize{6.5pt}{7pt}\selectfont$\pm$1.1}} & \textbf{92.4{\fontsize{6.5pt}{7pt}\selectfont$\pm$0.5}} & {\ul 82.4{\fontsize{6.5pt}{7pt}\selectfont$\pm$0.4}}    & \textbf{90.4{\fontsize{6.5pt}{7pt}\selectfont$\pm$1.0}} & {\ul 78.6{\fontsize{6.5pt}{7pt}\selectfont$\pm$0.1}}    \\
                        & M3DM           & 45.7{\fontsize{6.5pt}{7pt}\selectfont$\pm$1.1}          & 48.8{\fontsize{6.5pt}{7pt}\selectfont$\pm$1.4}          & 55.9{\fontsize{6.5pt}{7pt}\selectfont$\pm$0.4}          & 56.1{\fontsize{6.5pt}{7pt}\selectfont$\pm$2.3}          & 43.0{\fontsize{6.5pt}{7pt}\selectfont$\pm$0.7}          & 36.3{\fontsize{6.5pt}{7pt}\selectfont$\pm$1.3}          & 32.3{\fontsize{6.5pt}{7pt}\selectfont$\pm$0.2}          & 35.7{\fontsize{6.5pt}{7pt}\selectfont$\pm$2.9}          & 13.7{\fontsize{6.5pt}{7pt}\selectfont$\pm$0.8}          & 48.2{\fontsize{6.5pt}{7pt}\selectfont$\pm$0.6}          & 41.6{\fontsize{6.5pt}{7pt}\selectfont$\pm$0.3}          \\
                        & Ours           & \textbf{93.0{\fontsize{6.5pt}{7pt}\selectfont$\pm$0.8}} & \textbf{85.5{\fontsize{6.5pt}{7pt}\selectfont$\pm$1.6}} & \textbf{95.2{\fontsize{6.5pt}{7pt}\selectfont$\pm$0.6}} & \textbf{86.3{\fontsize{6.5pt}{7pt}\selectfont$\pm$0.5}} & {\ul 78.3{\fontsize{6.5pt}{7pt}\selectfont$\pm$2.1}}    & \textbf{76.8{\fontsize{6.5pt}{7pt}\selectfont$\pm$2.7}} & {\ul 76.0{\fontsize{6.5pt}{7pt}\selectfont$\pm$5.0}}    & {\ul 74.6{\fontsize{6.5pt}{7pt}\selectfont$\pm$3.1}}    & \textbf{90.3{\fontsize{6.5pt}{7pt}\selectfont$\pm$0.6}} & {\ul 81.3{\fontsize{6.5pt}{7pt}\selectfont$\pm$2.9}}    & \textbf{83.7{\fontsize{6.5pt}{7pt}\selectfont$\pm$0.5}} \\ \midrule
\multirow{4}{*}{\rotatebox{90}{Noise 30\%}} & PatchCore+FPFH & 18.1{\fontsize{6.5pt}{7pt}\selectfont$\pm$1.0}          & 23.6{\fontsize{6.5pt}{7pt}\selectfont$\pm$1.3}          & 35.2{\fontsize{6.5pt}{7pt}\selectfont$\pm$0.6}          & 38.3{\fontsize{6.5pt}{7pt}\selectfont$\pm$0.9}          & 17.2{\fontsize{6.5pt}{7pt}\selectfont$\pm$0.1}          & 11.7{\fontsize{6.5pt}{7pt}\selectfont$\pm$2.7}          & 5.3{\fontsize{6.5pt}{7pt}\selectfont$\pm$1.3}           & 6.2{\fontsize{6.5pt}{7pt}\selectfont$\pm$1.0}           & 7.0{\fontsize{6.5pt}{7pt}\selectfont$\pm$0.8}           & 25.0{\fontsize{6.5pt}{7pt}\selectfont$\pm$0.1}          & 18.8{\fontsize{6.5pt}{7pt}\selectfont$\pm$0.3}           \\
                        & Shape-Guided   & {\ul 70.9{\fontsize{6.5pt}{7pt}\selectfont$\pm$0.3}}    & {\ul 64.9{\fontsize{6.5pt}{7pt}\selectfont$\pm$1.9}}    & {\ul 89.1{\fontsize{6.5pt}{7pt}\selectfont$\pm$0.3}}    & {\ul 55.3{\fontsize{6.5pt}{7pt}\selectfont$\pm$1.4}}    & \textbf{83.2{\fontsize{6.5pt}{7pt}\selectfont$\pm$0.1}} & {\ul 56.6{\fontsize{6.5pt}{7pt}\selectfont$\pm$2.2}}    & \textbf{85.6{\fontsize{6.5pt}{7pt}\selectfont$\pm$0.5}} & \textbf{93.7{\fontsize{6.5pt}{7pt}\selectfont$\pm$0.3}} & {\ul 82.6{\fontsize{6.5pt}{7pt}\selectfont$\pm$0.4}}    & \textbf{89.7{\fontsize{6.5pt}{7pt}\selectfont$\pm$1.3}} & {\ul 77.2{\fontsize{6.5pt}{7pt}\selectfont$\pm$0.1}}   \\
                        & M3DM           & 18.7{\fontsize{6.5pt}{7pt}\selectfont$\pm$1.0}          & 24.0{\fontsize{6.5pt}{7pt}\selectfont$\pm$1.0}          & 35.3{\fontsize{6.5pt}{7pt}\selectfont$\pm$0.6}          & 39.2{\fontsize{6.5pt}{7pt}\selectfont$\pm$0.6}          & 17.7{\fontsize{6.5pt}{7pt}\selectfont$\pm$0.2}          & 18.2{\fontsize{6.5pt}{7pt}\selectfont$\pm$1.7}          & 5.7{\fontsize{6.5pt}{7pt}\selectfont$\pm$1.4}           & 7.1{\fontsize{6.5pt}{7pt}\selectfont$\pm$0.7}           & 7.6{\fontsize{6.5pt}{7pt}\selectfont$\pm$0.6}           & 25.1{\fontsize{6.5pt}{7pt}\selectfont$\pm$0.2}          & 19.9{\fontsize{6.5pt}{7pt}\selectfont$\pm$0.2}           \\
                        & Ours           & \textbf{90.7{\fontsize{6.5pt}{7pt}\selectfont$\pm$1.2}} & \textbf{81.5{\fontsize{6.5pt}{7pt}\selectfont$\pm$1.4}} & \textbf{94.8{\fontsize{6.5pt}{7pt}\selectfont$\pm$0.3}} & \textbf{84.5{\fontsize{6.5pt}{7pt}\selectfont$\pm$1.5}} & {\ul 75.4{\fontsize{6.5pt}{7pt}\selectfont$\pm$2.0}}    & \textbf{76.5{\fontsize{6.5pt}{7pt}\selectfont$\pm$3.4}} & {\ul 75.2{\fontsize{6.5pt}{7pt}\selectfont$\pm$1.8}}    & {\ul 71.4{\fontsize{6.5pt}{7pt}\selectfont$\pm$1.8}}    & \textbf{90.4{\fontsize{6.5pt}{7pt}\selectfont$\pm$0.6}} & {\ul 80.6{\fontsize{6.5pt}{7pt}\selectfont$\pm$2.8}}    & \textbf{82.1{\fontsize{6.5pt}{7pt}\selectfont$\pm$0.4}}
                        \\ \bottomrule
\end{tabular}
\end{center}
\end{table*}

\begin{table*}[t]
\begin{center}
\caption{\textbf{AUPRO score for anomaly segmentation under \textit{Non-Overlap} setting of all categories in MVTec 3D-AD.} We inject 20\% and 30\% noise into the training dataset. Our method outperforms other methods, indicating the superior anomaly detection ability of our method. We report the mean and standard deviation over 3 random seeds for each measurement. Optimal and sub-optimal results are in \textbf{bold} and {\ul underlined}, respectively.}
\label{tab:aupro_noise}
\setlength{\tabcolsep}{4.7pt}
\begin{tabular}{cc|cccccccccc|c}
\toprule
\textbf{}               & Method         & Bagel               & \begin{tabular}[c]{@{}c@{}}Cable\\ Gland\end{tabular}         & Carrot              & Cookie              & Dowel               & Foam                & Peach               & Potato              & Rope                & Tire       & Mean                \\ \midrule
\multirow{4}{*}{\rotatebox{90}{Noise 20\%}} & PatchCore+FPFH & \textbf{97.0{\fontsize{6.5pt}{7pt}\selectfont$\pm$0.2}} & \textbf{96.8{\fontsize{6.5pt}{7pt}\selectfont$\pm$0.6}} & {\ul 96.8{\fontsize{6.5pt}{7pt}\selectfont$\pm$0.0}}    & \textbf{94.8{\fontsize{6.5pt}{7pt}\selectfont$\pm$1.6}} & 91.6{\fontsize{6.5pt}{7pt}\selectfont$\pm$0.9}          & {\ul 89.7{\fontsize{6.5pt}{7pt}\selectfont$\pm$0.4}}    & 96.6{\fontsize{6.5pt}{7pt}\selectfont$\pm$0.5}          & {\ul 95.6{\fontsize{6.5pt}{7pt}\selectfont$\pm$0.2}}    & 96.6{\fontsize{6.5pt}{7pt}\selectfont$\pm$1.6}          & 95.5{\fontsize{6.5pt}{7pt}\selectfont$\pm$1.1}          & {\ul 95.1{\fontsize{6.5pt}{7pt}\selectfont$\pm$0.3}}    \\
                        & Shape-Guided   & 91.6{\fontsize{6.5pt}{7pt}\selectfont$\pm$1.2}          & 89.4{\fontsize{6.5pt}{7pt}\selectfont$\pm$0.7}          & 96.0{\fontsize{6.5pt}{7pt}\selectfont$\pm$0.5}          & 88.2{\fontsize{6.5pt}{7pt}\selectfont$\pm$0.8}          & \textbf{93.1{\fontsize{6.5pt}{7pt}\selectfont$\pm$0.8}} & 84.9{\fontsize{6.5pt}{7pt}\selectfont$\pm$6.0}          & 90.1{\fontsize{6.5pt}{7pt}\selectfont$\pm$1.5}          & 95.1{\fontsize{6.5pt}{7pt}\selectfont$\pm$1.0}          & 84.4{\fontsize{6.5pt}{7pt}\selectfont$\pm$4.7}          & 96.0{\fontsize{6.5pt}{7pt}\selectfont$\pm$1.2}          & 90.9{\fontsize{6.5pt}{7pt}\selectfont$\pm$1.2}          \\
                        & M3DM           & 93.8{\fontsize{6.5pt}{7pt}\selectfont$\pm$1.3}          & {\ul 95.6{\fontsize{6.5pt}{7pt}\selectfont$\pm$0.8}}    & 96.5{\fontsize{6.5pt}{7pt}\selectfont$\pm$0.1}          & 88.1{\fontsize{6.5pt}{7pt}\selectfont$\pm$1.3}          & {\ul 92.6{\fontsize{6.5pt}{7pt}\selectfont$\pm$2.1}}    & 80.0{\fontsize{6.5pt}{7pt}\selectfont$\pm$0.9}          & {\ul 97.1{\fontsize{6.5pt}{7pt}\selectfont$\pm$0.2}}    & 95.3{\fontsize{6.5pt}{7pt}\selectfont$\pm$0.7}          & \textbf{97.9{\fontsize{6.5pt}{7pt}\selectfont$\pm$0.5}} & \textbf{97.0{\fontsize{6.5pt}{7pt}\selectfont$\pm$0.5}} & 93.4{\fontsize{6.5pt}{7pt}\selectfont$\pm$0.3}          \\
                        & Ours           & {\ul 96.5{\fontsize{6.5pt}{7pt}\selectfont$\pm$0.6}}    & 95.6{\fontsize{6.5pt}{7pt}\selectfont$\pm$0.2}          & \textbf{97.7{\fontsize{6.5pt}{7pt}\selectfont$\pm$0.1}} & {\ul 92.2{\fontsize{6.5pt}{7pt}\selectfont$\pm$0.5}}    & 92.6{\fontsize{6.5pt}{7pt}\selectfont$\pm$1.7}          & \textbf{90.1{\fontsize{6.5pt}{7pt}\selectfont$\pm$0.7}} & \textbf{97.3{\fontsize{6.5pt}{7pt}\selectfont$\pm$0.1}} & \textbf{96.0{\fontsize{6.5pt}{7pt}\selectfont$\pm$0.2}} & {\ul 97.6{\fontsize{6.5pt}{7pt}\selectfont$\pm$1.0}}    & {\ul 96.6{\fontsize{6.5pt}{7pt}\selectfont$\pm$0.7}}    & \textbf{95.2{\fontsize{6.5pt}{7pt}\selectfont$\pm$0.3}} \\ \midrule
\multirow{4}{*}{\rotatebox{90}{Noise 30\%}} & PatchCore+FPFH & {\ul \textbf{96.6{\fontsize{6.5pt}{7pt}\selectfont$\pm$0.9}}} & {\ul 96.3{\fontsize{6.5pt}{7pt}\selectfont$\pm$1.9}}    & {\ul 96.8{\fontsize{6.5pt}{7pt}\selectfont$\pm$1.0}}    & \textbf{94.6{\fontsize{6.5pt}{7pt}\selectfont$\pm$0.9}} & {\ul 93.1{\fontsize{6.5pt}{7pt}\selectfont$\pm$1.2}}    & {\ul 87.9{\fontsize{6.5pt}{7pt}\selectfont$\pm$4.0}}    & 97.0{\fontsize{6.5pt}{7pt}\selectfont$\pm$0.6}          & 92.3{\fontsize{6.5pt}{7pt}\selectfont$\pm$7.5}          & 97.5{\fontsize{6.5pt}{7pt}\selectfont$\pm$1.4}          & \textbf{97.6{\fontsize{6.5pt}{7pt}\selectfont$\pm$0.3}} & {\ul 95.0{\fontsize{6.5pt}{7pt}\selectfont$\pm$0.3}}     \\
                        & Shape-Guided   & 73.7{\fontsize{6.5pt}{7pt}\selectfont$\pm$3.3}                & 79.4{\fontsize{6.5pt}{7pt}\selectfont$\pm$1.9}          & 93.6{\fontsize{6.5pt}{7pt}\selectfont$\pm$0.3}          & 82.4{\fontsize{6.5pt}{7pt}\selectfont$\pm$2.1}          & 88.4{\fontsize{6.5pt}{7pt}\selectfont$\pm$2.6}          & 69.3{\fontsize{6.5pt}{7pt}\selectfont$\pm$0.2}          & 72.6{\fontsize{6.5pt}{7pt}\selectfont$\pm$3.4}          & 88.7{\fontsize{6.5pt}{7pt}\selectfont$\pm$3.3}          & 81.0{\fontsize{6.5pt}{7pt}\selectfont$\pm$5.7}          & 93.7{\fontsize{6.5pt}{7pt}\selectfont$\pm$1.9}          & 82.3{\fontsize{6.5pt}{7pt}\selectfont$\pm$0.7}           \\
                        & M3DM           & 94.3{\fontsize{6.5pt}{7pt}\selectfont$\pm$2.7}                & \textbf{97.2{\fontsize{6.5pt}{7pt}\selectfont$\pm$0.7}} & 96.4{\fontsize{6.5pt}{7pt}\selectfont$\pm$0.9}          & 87.5{\fontsize{6.5pt}{7pt}\selectfont$\pm$0.4}          & 92.5{\fontsize{6.5pt}{7pt}\selectfont$\pm$1.6}          & 83.6{\fontsize{6.5pt}{7pt}\selectfont$\pm$6.5}          & {\ul 97.4{\fontsize{6.5pt}{7pt}\selectfont$\pm$0.1}}    & {\ul 93.3{\fontsize{6.5pt}{7pt}\selectfont$\pm$5.5}}    & \textbf{97.6{\fontsize{6.5pt}{7pt}\selectfont$\pm$1.2}} & {\ul 96.9{\fontsize{6.5pt}{7pt}\selectfont$\pm$1.1}}    & 93.7{\fontsize{6.5pt}{7pt}\selectfont$\pm$0.6}           \\
                        & Ours           & {\ul \textbf{96.6{\fontsize{6.5pt}{7pt}\selectfont$\pm$0.7}}} & 95.0{\fontsize{6.5pt}{7pt}\selectfont$\pm$0.4}          & \textbf{97.7{\fontsize{6.5pt}{7pt}\selectfont$\pm$0.1}} & {\ul 92.3{\fontsize{6.5pt}{7pt}\selectfont$\pm$0.8}}    & \textbf{93.9{\fontsize{6.5pt}{7pt}\selectfont$\pm$0.5}} & \textbf{89.5{\fontsize{6.5pt}{7pt}\selectfont$\pm$2.6}} & \textbf{97.7{\fontsize{6.5pt}{7pt}\selectfont$\pm$0.4}} & \textbf{95.4{\fontsize{6.5pt}{7pt}\selectfont$\pm$0.4}} & {\ul 97.5{\fontsize{6.5pt}{7pt}\selectfont$\pm$1.3}}    & 96.1{\fontsize{6.5pt}{7pt}\selectfont$\pm$1.2}          & \textbf{95.2{\fontsize{6.5pt}{7pt}\selectfont$\pm$0.2}}
                        \\ \bottomrule
\end{tabular}
\end{center}
\end{table*}

\begin{table*}[t]
\begin{center}
\caption{\textbf{P-AUROC score for anomaly segmentation under \textit{Overlap} setting of all categories in MVTec 3D-AD.} We inject 20\% and 30\% noise into the training dataset. Our method outperforms other methods, indicating the superior anomaly detection ability of our method. We report the mean and standard deviation over 3 random seeds for each measurement. Optimal and sub-optimal results are in \textbf{bold} and {\ul underlined}, respectively.}
\label{tab:pauroc_overlap_noise}
\setlength{\tabcolsep}{4.7pt}
\begin{tabular}{cc|cccccccccc|c}
\toprule
\textbf{}               & Method         & Bagel               & \begin{tabular}[c]{@{}c@{}}Cable\\ Gland\end{tabular}         & Carrot              & Cookie              & Dowel               & Foam                & Peach               & Potato              & Rope                & Tire       & Mean                \\ \midrule
\multirow{5}{*}{\rotatebox{90}{Noise 20\%}} & PatchCore+FPFH & 50.2{\fontsize{6.5pt}{7pt}\selectfont$\pm$2.1}          & 52.4{\fontsize{6.5pt}{7pt}\selectfont$\pm$2.6}          & 55.8{\fontsize{6.5pt}{7pt}\selectfont$\pm$3.0}          & 62.3{\fontsize{6.5pt}{7pt}\selectfont$\pm$2.0}          & 46.3{\fontsize{6.5pt}{7pt}\selectfont$\pm$0.6}          & 39.8{\fontsize{6.5pt}{7pt}\selectfont$\pm$1.0}          & 32.9{\fontsize{6.5pt}{7pt}\selectfont$\pm$2.1}          & 36.3{\fontsize{6.5pt}{7pt}\selectfont$\pm$2.1}          & 18.3{\fontsize{6.5pt}{7pt}\selectfont$\pm$6.3}          & 49.4{\fontsize{6.5pt}{7pt}\selectfont$\pm$0.8} & 44.4{\fontsize{6.5pt}{7pt}\selectfont$\pm$0.8}          \\
                        & AST            & 83.1{\fontsize{6.5pt}{7pt}\selectfont$\pm$0.0}          & \textbf{91.9{\fontsize{6.5pt}{7pt}\selectfont$\pm$0.6}} & 95.8{\fontsize{6.5pt}{7pt}\selectfont$\pm$1.0}          & 83.6{\fontsize{6.5pt}{7pt}\selectfont$\pm$1.0}          & \textbf{89.1{\fontsize{6.5pt}{7pt}\selectfont$\pm$0.6}} & {\ul 84.6{\fontsize{6.5pt}{7pt}\selectfont$\pm$0.6}}    & \textbf{88.8{\fontsize{6.5pt}{7pt}\selectfont$\pm$0.6}} & \textbf{88.0{\fontsize{6.5pt}{7pt}\selectfont$\pm$0.6}} & 89.0{\fontsize{6.5pt}{7pt}\selectfont$\pm$0.6}          & 88.9{\fontsize{6.5pt}{7pt}\selectfont$\pm$1.0}      & 88.8{\fontsize{6.5pt}{7pt}\selectfont$\pm$0.2}          \\
                        & Shape-Guided   & {\ul 89.9{\fontsize{6.5pt}{7pt}\selectfont$\pm$0.3}}    & 91.0{\fontsize{6.5pt}{7pt}\selectfont$\pm$0.5}          & \textbf{97.0{\fontsize{6.5pt}{7pt}\selectfont$\pm$0.1}} & {\ul 86.5{\fontsize{6.5pt}{7pt}\selectfont$\pm$0.2}}    & 85.1{\fontsize{6.5pt}{7pt}\selectfont$\pm$0.5}          & \textbf{86.4{\fontsize{6.5pt}{7pt}\selectfont$\pm$0.6}} & {\ul 84.9{\fontsize{6.5pt}{7pt}\selectfont$\pm$0.3}}    & {\ul 81.3{\fontsize{6.5pt}{7pt}\selectfont$\pm$5.8}}    & {\ul 94.7{\fontsize{6.5pt}{7pt}\selectfont$\pm$0.4}}    & 91.1{\fontsize{6.5pt}{7pt}\selectfont$\pm$5.6} & {\ul 88.8{\fontsize{6.5pt}{7pt}\selectfont$\pm$0.5}}    \\
                        & M3DM           & 52.4{\fontsize{6.5pt}{7pt}\selectfont$\pm$2.3}          & 53.0{\fontsize{6.5pt}{7pt}\selectfont$\pm$3.1}          & 56.1{\fontsize{6.5pt}{7pt}\selectfont$\pm$2.7}          & 65.8{\fontsize{6.5pt}{7pt}\selectfont$\pm$1.9}          & 47.3{\fontsize{6.5pt}{7pt}\selectfont$\pm$1.3}          & 51.3{\fontsize{6.5pt}{7pt}\selectfont$\pm$0.7}          & 34.4{\fontsize{6.5pt}{7pt}\selectfont$\pm$2.3}          & 37.0{\fontsize{6.5pt}{7pt}\selectfont$\pm$2.1}          & 18.7{\fontsize{6.5pt}{7pt}\selectfont$\pm$6.0}          & 50.7{\fontsize{6.5pt}{7pt}\selectfont$\pm$0.4} & 46.7{\fontsize{6.5pt}{7pt}\selectfont$\pm$0.8}          \\
                        & Ours           & \textbf{97.8{\fontsize{6.5pt}{7pt}\selectfont$\pm$0.4}} & {\ul 91.4{\fontsize{6.5pt}{7pt}\selectfont$\pm$2.1}}    & {\ul 96.4{\fontsize{6.5pt}{7pt}\selectfont$\pm$0.6}}    & \textbf{94.3{\fontsize{6.5pt}{7pt}\selectfont$\pm$0.1}} & {\ul 85.5{\fontsize{6.5pt}{7pt}\selectfont$\pm$2.4}}    & 80.8{\fontsize{6.5pt}{7pt}\selectfont$\pm$2.8}          & 81.4{\fontsize{6.5pt}{7pt}\selectfont$\pm$2.1}          & 78.4{\fontsize{6.5pt}{7pt}\selectfont$\pm$3.0}          & \textbf{97.8{\fontsize{6.5pt}{7pt}\selectfont$\pm$0.4}} & 86.7{\fontsize{6.5pt}{7pt}\selectfont$\pm$1.9} & \textbf{89.0{\fontsize{6.5pt}{7pt}\selectfont$\pm$0.4}} \\ \midrule
\multirow{5}{*}{\rotatebox{90}{Noise 30\%}} & PatchCore+FPFH & 24.0{\fontsize{6.5pt}{7pt}\selectfont$\pm$4.5}          & 26.8{\fontsize{6.5pt}{7pt}\selectfont$\pm$1.3}          & 34.5{\fontsize{6.5pt}{7pt}\selectfont$\pm$2.8}          & 40.6{\fontsize{6.5pt}{7pt}\selectfont$\pm$3.4}          & 21.3{\fontsize{6.5pt}{7pt}\selectfont$\pm$2.6}          & 17.4{\fontsize{6.5pt}{7pt}\selectfont$\pm$0.9}          & 8.8{\fontsize{6.5pt}{7pt}\selectfont$\pm$1.6}           & 8.0{\fontsize{6.5pt}{7pt}\selectfont$\pm$2.1}           & 8.2{\fontsize{6.5pt}{7pt}\selectfont$\pm$3.2}           & 25.8{\fontsize{6.5pt}{7pt}\selectfont$\pm$2.7}          & 21.6{\fontsize{6.5pt}{7pt}\selectfont$\pm$1.4}           \\
                        & AST            & 15.3{\fontsize{6.5pt}{7pt}\selectfont$\pm$0.0}          & 21.4{\fontsize{6.5pt}{7pt}\selectfont$\pm$0.0}          & 29.3{\fontsize{6.5pt}{7pt}\selectfont$\pm$0.6}          & 37.8{\fontsize{6.5pt}{7pt}\selectfont$\pm$0.6}          & 16.4{\fontsize{6.5pt}{7pt}\selectfont$\pm$1.0}          & 8.9{\fontsize{6.5pt}{7pt}\selectfont$\pm$0.6}           & 3.6{\fontsize{6.5pt}{7pt}\selectfont$\pm$1.0}           & 5.3{\fontsize{6.5pt}{7pt}\selectfont$\pm$0.0}           & 6.8{\fontsize{6.5pt}{7pt}\selectfont$\pm$0.0}           & 19.9{\fontsize{6.5pt}{7pt}\selectfont$\pm$0.6}          & 16.5{\fontsize{6.5pt}{7pt}\selectfont$\pm$0.1}           \\
                        & Shape-Guided   & {\ul 90.7{\fontsize{6.5pt}{7pt}\selectfont$\pm$0.7}}    & \textbf{89.4{\fontsize{6.5pt}{7pt}\selectfont$\pm$0.5}} & \textbf{96.6{\fontsize{6.5pt}{7pt}\selectfont$\pm$0.3}} & {\ul 83.0{\fontsize{6.5pt}{7pt}\selectfont$\pm$1.0}}    & {\ul 80.9{\fontsize{6.5pt}{7pt}\selectfont$\pm$5.8}}    & \textbf{80.8{\fontsize{6.5pt}{7pt}\selectfont$\pm$4.8}} & \textbf{90.0{\fontsize{6.5pt}{7pt}\selectfont$\pm$5.1}} & \textbf{81.6{\fontsize{6.5pt}{7pt}\selectfont$\pm$5.7}} & {\ul 94.5{\fontsize{6.5pt}{7pt}\selectfont$\pm$0.2}}    & \textbf{87.8{\fontsize{6.5pt}{7pt}\selectfont$\pm$0.5}} & {\ul 87.5{\fontsize{6.5pt}{7pt}\selectfont$\pm$0.9}}     \\
                        & M3DM           & 26.3{\fontsize{6.5pt}{7pt}\selectfont$\pm$4.5}          & 27.3{\fontsize{6.5pt}{7pt}\selectfont$\pm$1.7}          & 35.0{\fontsize{6.5pt}{7pt}\selectfont$\pm$2.3}          & 48.3{\fontsize{6.5pt}{7pt}\selectfont$\pm$5.0}          & 22.6{\fontsize{6.5pt}{7pt}\selectfont$\pm$3.1}          & 35.4{\fontsize{6.5pt}{7pt}\selectfont$\pm$1.6}          & 9.9{\fontsize{6.5pt}{7pt}\selectfont$\pm$1.4}           & 9.0{\fontsize{6.5pt}{7pt}\selectfont$\pm$1.9}           & 7.9{\fontsize{6.5pt}{7pt}\selectfont$\pm$3.5}           & 28.4{\fontsize{6.5pt}{7pt}\selectfont$\pm$2.4}          & 25.0{\fontsize{6.5pt}{7pt}\selectfont$\pm$1.4}           \\
                        & Ours           & \textbf{96.6{\fontsize{6.5pt}{7pt}\selectfont$\pm$0.5}} & {\ul 89.3{\fontsize{6.5pt}{7pt}\selectfont$\pm$1.5}}    & {\ul 96.5{\fontsize{6.5pt}{7pt}\selectfont$\pm$0.3}}    & \textbf{92.7{\fontsize{6.5pt}{7pt}\selectfont$\pm$1.5}} & \textbf{81.9{\fontsize{6.5pt}{7pt}\selectfont$\pm$2.8}} & {\ul 79.9{\fontsize{6.5pt}{7pt}\selectfont$\pm$2.7}}    & {\ul 81.9{\fontsize{6.5pt}{7pt}\selectfont$\pm$0.7}}    & {\ul 74.8{\fontsize{6.5pt}{7pt}\selectfont$\pm$1.3}}    & \textbf{97.8{\fontsize{6.5pt}{7pt}\selectfont$\pm$0.4}} & {\ul 86.8{\fontsize{6.5pt}{7pt}\selectfont$\pm$1.9}}    & \textbf{87.8{\fontsize{6.5pt}{7pt}\selectfont$\pm$0.4}}
                        \\ \bottomrule
\end{tabular}
\end{center}
\end{table*}

\begin{table*}[t]
\begin{center}
\caption{\textbf{P-AUROC score for anomaly segmentation under \textit{Non-Overlap} setting of all categories in MVTec 3D-AD.} We inject 20\% and 30\% noise into the training dataset. Our method outperforms other methods, indicating the superior anomaly detection ability of our method. We report the mean and standard deviation over 3 random seeds for each measurement. Optimal and sub-optimal results are in \textbf{bold} and {\ul underlined}, respectively.}
\label{tab:pauroc_noise}
\setlength{\tabcolsep}{4.7pt}
\begin{tabular}{cc|cccccccccc|c}
\toprule
\textbf{}               & Method         & Bagel               & \begin{tabular}[c]{@{}c@{}}Cable\\ Gland\end{tabular}         & Carrot              & Cookie              & Dowel               & Foam                & Peach               & Potato              & Rope                & Tire       & Mean                \\ \midrule
\multirow{5}{*}{\rotatebox{90}{Noise 20\%}} & PatchCore+FPFH & \textbf{99.5{\fontsize{6.5pt}{7pt}\selectfont$\pm$0.0}} & \textbf{99.1{\fontsize{6.5pt}{7pt}\selectfont$\pm$0.3}} & 98.2{\fontsize{6.5pt}{7pt}\selectfont$\pm$0.1}          & \textbf{98.7{\fontsize{6.5pt}{7pt}\selectfont$\pm$0.1}} & \textbf{98.0{\fontsize{6.5pt}{7pt}\selectfont$\pm$0.8}} & 97.6{\fontsize{6.5pt}{7pt}\selectfont$\pm$0.1}          & 99.3{\fontsize{6.5pt}{7pt}\selectfont$\pm$0.1}          & 98.0{\fontsize{6.5pt}{7pt}\selectfont$\pm$0.3}          & 99.7{\fontsize{6.5pt}{7pt}\selectfont$\pm$0.2}          & {\ul 99.5{\fontsize{6.5pt}{7pt}\selectfont$\pm$0.1}}    & {\ul 98.8{\fontsize{6.5pt}{7pt}\selectfont$\pm$0.0}}    \\
                        & AST            & 97.9{\fontsize{6.5pt}{7pt}\selectfont$\pm$0.6}          & 97.8{\fontsize{6.5pt}{7pt}\selectfont$\pm$0.0}          & {\ul 99.4{\fontsize{6.5pt}{7pt}\selectfont$\pm$1.0}}    & 92.2{\fontsize{6.5pt}{7pt}\selectfont$\pm$0.6}          & 93.1{\fontsize{6.5pt}{7pt}\selectfont$\pm$0.0}          & \textbf{99.2{\fontsize{6.5pt}{7pt}\selectfont$\pm$0.6}} & \textbf{99.5{\fontsize{6.5pt}{7pt}\selectfont$\pm$1.0}} & \textbf{99.8{\fontsize{6.5pt}{7pt}\selectfont$\pm$0.6}} & 97.5{\fontsize{6.5pt}{7pt}\selectfont$\pm$0.0}          & 98.6{\fontsize{6.5pt}{7pt}\selectfont$\pm$1.0}          & 97.5{\fontsize{6.5pt}{7pt}\selectfont$\pm$0.1}          \\
                        & Shape-Guided   & 97.5{\fontsize{6.5pt}{7pt}\selectfont$\pm$1.7}          & 97.5{\fontsize{6.5pt}{7pt}\selectfont$\pm$0.5}          & 98.9{\fontsize{6.5pt}{7pt}\selectfont$\pm$0.3}          & 95.2{\fontsize{6.5pt}{7pt}\selectfont$\pm$0.3}          & 97.8{\fontsize{6.5pt}{7pt}\selectfont$\pm$0.4}          & 93.3{\fontsize{6.5pt}{7pt}\selectfont$\pm$3.3}          & 97.1{\fontsize{6.5pt}{7pt}\selectfont$\pm$0.3}          & 98.9{\fontsize{6.5pt}{7pt}\selectfont$\pm$0.2}          & 95.6{\fontsize{6.5pt}{7pt}\selectfont$\pm$1.4}          & 99.2{\fontsize{6.5pt}{7pt}\selectfont$\pm$0.7}          & 97.1{\fontsize{6.5pt}{7pt}\selectfont$\pm$0.4}          \\
                        & M3DM           & 99.0{\fontsize{6.5pt}{7pt}\selectfont$\pm$0.2}          & 98.8{\fontsize{6.5pt}{7pt}\selectfont$\pm$0.3}          & 98.1{\fontsize{6.5pt}{7pt}\selectfont$\pm$0.1}          & 96.8{\fontsize{6.5pt}{7pt}\selectfont$\pm$0.1}          & 97.8{\fontsize{6.5pt}{7pt}\selectfont$\pm$1.0}          & 95.4{\fontsize{6.5pt}{7pt}\selectfont$\pm$0.1}          & 99.4{\fontsize{6.5pt}{7pt}\selectfont$\pm$0.1}          & 99.0{\fontsize{6.5pt}{7pt}\selectfont$\pm$0.2}          & \textbf{99.8{\fontsize{6.5pt}{7pt}\selectfont$\pm$0.1}} & \textbf{99.5{\fontsize{6.5pt}{7pt}\selectfont$\pm$0.2}} & 98.4{\fontsize{6.5pt}{7pt}\selectfont$\pm$0.0}          \\
                        & Ours           & \textbf{99.5{\fontsize{6.5pt}{7pt}\selectfont$\pm$0.1}} & {\ul 99.0{\fontsize{6.5pt}{7pt}\selectfont$\pm$0.1}}    & \textbf{99.6{\fontsize{6.5pt}{7pt}\selectfont$\pm$0.1}} & {\ul 97.3{\fontsize{6.5pt}{7pt}\selectfont$\pm$0.3}}    & {\ul 97.9{\fontsize{6.5pt}{7pt}\selectfont$\pm$0.6}}    & {\ul 97.7{\fontsize{6.5pt}{7pt}\selectfont$\pm$0.2}}    & {\ul 99.5{\fontsize{6.5pt}{7pt}\selectfont$\pm$0.1}}    & {\ul 99.1{\fontsize{6.5pt}{7pt}\selectfont$\pm$0.1}}    & {\ul 99.8{\fontsize{6.5pt}{7pt}\selectfont$\pm$0.2}}    & 99.3{\fontsize{6.5pt}{7pt}\selectfont$\pm$0.2}          & \textbf{98.9{\fontsize{6.5pt}{7pt}\selectfont$\pm$0.1}} \\ \midrule
\multirow{5}{*}{\rotatebox{90}{Noise 30\%}} & PatchCore+FPFH & \textbf{99.6{\fontsize{6.5pt}{7pt}\selectfont$\pm$0.1}} & \textbf{99.3{\fontsize{6.5pt}{7pt}\selectfont$\pm$0.1}} & 98.2{\fontsize{6.5pt}{7pt}\selectfont$\pm$1.3}          & \textbf{98.5{\fontsize{6.5pt}{7pt}\selectfont$\pm$0.2}} & {\ul 98.1{\fontsize{6.5pt}{7pt}\selectfont$\pm$1.3}}    & \textbf{98.3{\fontsize{6.5pt}{7pt}\selectfont$\pm$0.5}} & 99.5{\fontsize{6.5pt}{7pt}\selectfont$\pm$0.2}          & 95.3{\fontsize{6.5pt}{7pt}\selectfont$\pm$7.8}           & 99.6{\fontsize{6.5pt}{7pt}\selectfont$\pm$0.6}          & \textbf{99.6{\fontsize{6.5pt}{7pt}\selectfont$\pm$0.1}} & {\ul 98.6{\fontsize{6.5pt}{7pt}\selectfont$\pm$0.7}}    \\
                        & AST            & 91.0{\fontsize{6.5pt}{7pt}\selectfont$\pm$1.0}          & 96.3{\fontsize{6.5pt}{7pt}\selectfont$\pm$0.6}          & {\ul 99.1{\fontsize{6.5pt}{7pt}\selectfont$\pm$1.0}}    & 92.4{\fontsize{6.5pt}{7pt}\selectfont$\pm$0.6}          & 95.6{\fontsize{6.5pt}{7pt}\selectfont$\pm$0.0}          & {\ul 97.4{\fontsize{6.5pt}{7pt}\selectfont$\pm$0.6}}    & \textbf{99.7{\fontsize{6.5pt}{7pt}\selectfont$\pm$0.6}} & \textbf{100.2{\fontsize{6.5pt}{7pt}\selectfont$\pm$0.6}} & 97.7{\fontsize{6.5pt}{7pt}\selectfont$\pm$0.6}          & 98.6{\fontsize{6.5pt}{7pt}\selectfont$\pm$1.0}          & 96.8{\fontsize{6.5pt}{7pt}\selectfont$\pm$0.2}           \\
                        & Shape-Guided   & 93.3{\fontsize{6.5pt}{7pt}\selectfont$\pm$1.3}          & 93.8{\fontsize{6.5pt}{7pt}\selectfont$\pm$1.0}          & 98.2{\fontsize{6.5pt}{7pt}\selectfont$\pm$0.3}          & 90.5{\fontsize{6.5pt}{7pt}\selectfont$\pm$1.2}          & 97.0{\fontsize{6.5pt}{7pt}\selectfont$\pm$1.1}          & 85.6{\fontsize{6.5pt}{7pt}\selectfont$\pm$2.8}          & 92.7{\fontsize{6.5pt}{7pt}\selectfont$\pm$0.0}          & 96.7{\fontsize{6.5pt}{7pt}\selectfont$\pm$0.5}           & 92.8{\fontsize{6.5pt}{7pt}\selectfont$\pm$1.9}          & 98.6{\fontsize{6.5pt}{7pt}\selectfont$\pm$0.7}          & 93.9{\fontsize{6.5pt}{7pt}\selectfont$\pm$0.2}           \\
                        & M3DM           & 99.1{\fontsize{6.5pt}{7pt}\selectfont$\pm$0.2}          & {\ul 99.3{\fontsize{6.5pt}{7pt}\selectfont$\pm$0.4}}    & 98.0{\fontsize{6.5pt}{7pt}\selectfont$\pm$1.4}          & 96.0{\fontsize{6.5pt}{7pt}\selectfont$\pm$0.8}          & 97.6{\fontsize{6.5pt}{7pt}\selectfont$\pm$1.1}          & 96.6{\fontsize{6.5pt}{7pt}\selectfont$\pm$1.9}          & {\ul 99.6{\fontsize{6.5pt}{7pt}\selectfont$\pm$0.1}}    & 98.2{\fontsize{6.5pt}{7pt}\selectfont$\pm$2.0}           & \textbf{99.7{\fontsize{6.5pt}{7pt}\selectfont$\pm$0.5}} & {\ul 99.5{\fontsize{6.5pt}{7pt}\selectfont$\pm$0.3}}    & 98.4{\fontsize{6.5pt}{7pt}\selectfont$\pm$0.2}            \\
                        & Ours           & {\ul 99.5{\fontsize{6.5pt}{7pt}\selectfont$\pm$0.2}}    & 98.7{\fontsize{6.5pt}{7pt}\selectfont$\pm$0.3}          & \textbf{99.6{\fontsize{6.5pt}{7pt}\selectfont$\pm$0.1}} & {\ul 97.1{\fontsize{6.5pt}{7pt}\selectfont$\pm$0.6}}    & \textbf{98.5{\fontsize{6.5pt}{7pt}\selectfont$\pm$0.3}} & 97.4{\fontsize{6.5pt}{7pt}\selectfont$\pm$1.3}          & {\ul 99.6{\fontsize{6.5pt}{7pt}\selectfont$\pm$0.1}}    & {\ul 98.8{\fontsize{6.5pt}{7pt}\selectfont$\pm$0.3}}     & {\ul 99.6{\fontsize{6.5pt}{7pt}\selectfont$\pm$0.6}}    & 99.2{\fontsize{6.5pt}{7pt}\selectfont$\pm$0.4}          & \textbf{98.8{\fontsize{6.5pt}{7pt}\selectfont$\pm$0.0}}
                        \\ \bottomrule
\end{tabular}
\end{center}
\end{table*}

\vfill

\end{document}